\documentclass[]{fairmeta}

\usepackage{amsmath,amsfonts,bm}

\def\eqref#1{equation~\ref{#1}}

\def\1{\bm{1}}

\DeclareMathAlphabet{\mathsfit}{\encodingdefault}{\sfdefault}{m}{sl}
\SetMathAlphabet{\mathsfit}{bold}{\encodingdefault}{\sfdefault}{bx}{n}

\usepackage{makecell}
\usepackage{hyperref}
\usepackage{url}
\usepackage{graphicx} 
\usepackage{enumitem}
\setlist[itemize]{itemsep=0pt}
\usepackage{amsmath}
\usepackage{amssymb}
\usepackage{mathtools}
\usepackage{amsthm}
\usepackage{bbm}
\usepackage{graphicx,xcolor,placeins,caption}
\usepackage{comment}
\usepackage{booktabs}
\usepackage{multirow}
\usepackage{tabularx}
\usepackage{array}
\usepackage{forest}
\usepackage[most]{tcolorbox}
\usepackage{xcolor}              
\definecolor{promptbg}{gray}{0.95}
\usepackage{bbm}
\usepackage{subcaption}
\usepackage{ulem}
\usepackage[table]{xcolor}
\definecolor{lightred}{RGB}{255, 128, 128}
\definecolor{lightgreen}{RGB}{144, 238, 144}

\theoremstyle{plain}
\newtheorem{theorem}{Theorem}[section]

\theoremstyle{definition}
\newtheorem{definition}[theorem]{Definition}

\theoremstyle{remark}

\newtcolorbox{promptbox}{
  colback=promptbg,              
  colframe=gray!50,              
  boxrule=0.5pt,                 
  arc=1mm,                       
  left=2mm, right=2mm, top=1mm, bottom=1mm,
  fontupper=\ttfamily,           
}

\title{Task-Dependent Evaluation of LLM Output Homogenization: A Taxonomy-Guided Framework}

\author[1,2,\dagger]{Shomik Jain}
\author[1,*]{Jack Lanchantin}
\author[1,*]{Maximilian Nickel}
\author[1,*]{Candace Ross}
\author[1,*]{Karen Ullrich}
\author[2,*]{\\Ashia Wilson}
\author[1]{Jamelle Watson-Daniels}

\affiliation[1]{FAIR at Meta}
\affiliation[2]{Massachusetts Institute of Technology}

\contribution[\dagger]{Work done during an internship at Meta}
\contribution[*]{Middle authors listed alphabetically}

\abstract{
Large language models often generate homogeneous outputs, but whether this is problematic depends on the specific task. For objective math tasks, responses may vary in terms of problem-solving strategy but should maintain the same verifiable answer. Whereas, for creative writing tasks, we often expect variation in key narrative components (e.g. plot, setting, etc.) beyond mere vocabulary diversity. Prior work on homogenization rarely conceptualizes diversity in a task-dependent way. We address this gap with four contributions: (1) a task taxonomy with distinct notions of functional diversity -- whether a user would perceive two responses as meaningfully different for a given task; (2) a small user study validating that the taxonomy aligns with human perception of functional diversity; (3) a task-dependent sampling technique that increases diversity only where homogenization is undesired; (4) evidence challenging the perceived diversity-quality trade-off, showing it may stem from mis-conceptualizing both diversity and quality in a task-agnostic way.
}

\date{April 21, 2026}
\correspondence{Jamelle Watson-Daniels at \email{watsondaniels@meta.com}}

\begin{document}

\maketitle

\section{Introduction}

\begin{figure}[t!]
    \centering
    \includegraphics[width=\linewidth]{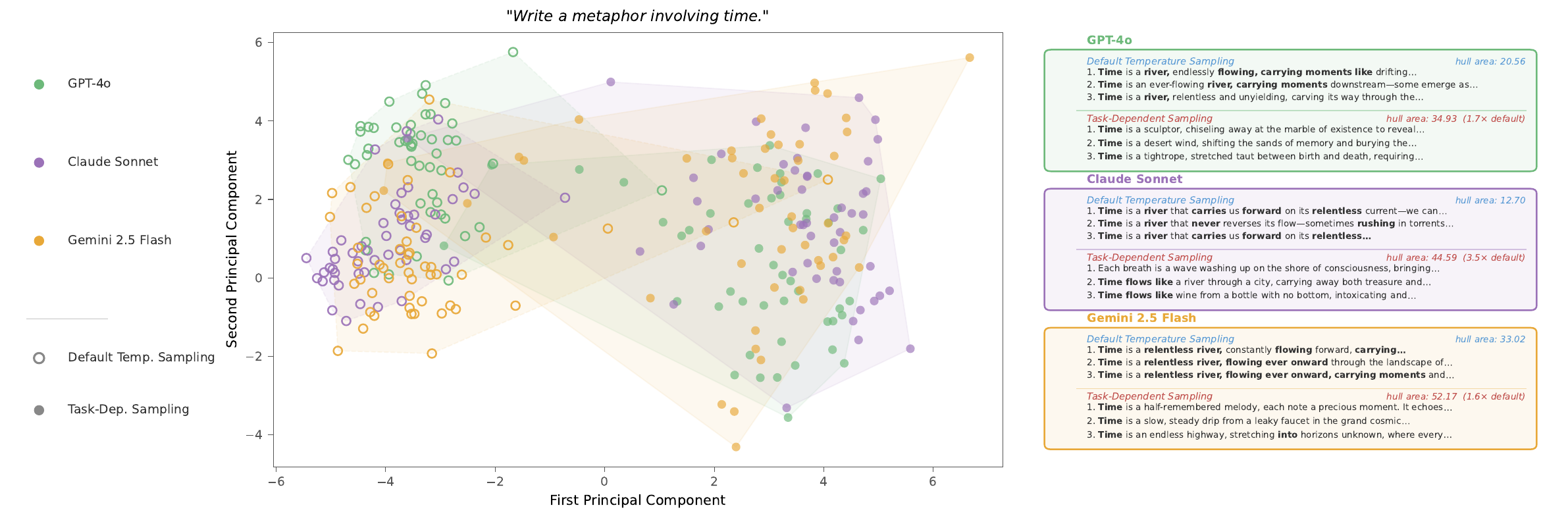}
    \caption{\textbf{Meaningful diversity goes beyond lexical variation; temperature-sampling yields homogeneous outputs for open-ended creative prompts whereas our task-dependent sampling elicits functionally different responses.} As an illustration, we plot a PCA projection of response embeddings for \textit{``Write a metaphor involving time''} (\textsc{Infinity-Chat}). Open circles: default temperature sampling; filled circles: task-dependent sampling. Convex hulls show the spread of model outputs; task-dependent sampling increases hull area by 1.7$\times$ (GPT-4o), 3.5$\times$ (Claude-4-Sonnet), and 1.6$\times$ (Gemini 2.5 Flash). Response excerpts show the qualitative shift: models default to river metaphors, while task-dependent sampling produces distinct conceptual framings beyond surface-level lexical variation.}
    \label{fig:metaphor_time}
\end{figure}

Large language models (LLMs) often generate homogeneous outputs, but whether this is problematic depends on the specific task. Suppose a user asks for a joke and a model always responds with a ``knock-knock'' joke; such homogenization undermines the model's creative utility. By contrast, for tasks with verifiable solutions such as solving a math problem, consistency is desirable, although variation in the explanation or problem-solving approach may be helpful. Our central claim is that the implications of homogenization are task-dependent, and therefore its evaluation and mitigation should be as well (Figure~\ref{fig:metaphor_time}).

Our task-dependent perspective can directly improve large-scale homogenization benchmarks such as \textsc{Novelty-Bench}~\citep{zhang2025noveltybench} and \textsc{Infinity-Chat}~\citep{jiang2025artificial}. These datasets are restricted by design to open-ended queries where diversity is expected. Moreover, they rely on general embedding models or semantic classifiers that apply a single notion of diversity to all tasks. We contend that both the \textit{desirability} of diversity and the \textit{type} of diversity depend on where the task falls on a verifiability spectrum (c.f. Table~\ref{tab:task_categories}). 

Existing approaches to reducing output homogenization also rarely take task dependence into account. Several recent works propose methods that promote diversity in the alignment process~\citep{lanchantin2025diverse, li2025jointly} or when sampling outputs at inference time~\citep{wang2025multilingual, zhang2025cultivating, zhang2025noveltybench}. For example, \citet{zhang2025cultivating} promote ``diverse values'' when sampling multiple outputs. Without a task-dependent approach, such methods may (1) fail to encourage diversity that is meaningful for a task, and/or (2) reduce homogenization in tasks where it is desired. 

Consider the stakes: if a model wrongly conceptualizes functional diversity for a task that mimics an encyclopedia inquiry, the model could misrepresent historical events in an attempt to naively reduce homogenization. Conversely, if a model wrongly conceptualizes functional diversity for a creative writing task, the model might repeat the same story arc no matter how many times a user asks the model to tell a story.

In this work, we introduce a task-dependent framework to evaluate and mitigate output homogenization in LLMs. Our contributions are as follows.

\begin{enumerate}
\item We present a \textit{task taxonomy} to operationalize \textit{task-dependent functional diversity} (\S~\ref{subsec:framework_taxonomy}); each task category represents a distinct notion of output diversity studied in the literature (Table~\ref{tab:task_categories}). Taken together, the categories clarify where response variation is meaningful and also where homogenization is desired for model performance.

\item We validate that the taxonomy aligns with \textit{human judgments of functional diversity} via a small user study ($n=22$). Annotators label functional differences rather than overall preference (to decouple diversity from quality). Taxonomy-guided LLM-judges correlate most strongly with human annotations ($r_s=0.93$), outperforming embedding-based ($r_s=0.81$) and vocabulary-based ($r_s=0.67$) metrics, supporting task-dependent functional diversity as a promising improvement for future homogenization evaluations.

\item We propose a \textit{task-dependent sampling technique} that improves on previous prompt-based methods to sample diverse responses~\citep{zhang2025cultivating, zhang2025noveltybench}. Our approach increases functional diversity for task categories where homogenization is undesired, while preserving homogenization where it is desired (Figure~\ref{fig:fun_div_gpt}). Our technique also generalizes to \textsc{Infinity-Chat}~\cite{jiang2025artificial}, substantially increasing functional diversity on the recent homogenization benchmark (Appx.~\ref{apx:infinity-chat-res}).

\item We challenge the \textit{perceived existence of a diversity-quality trade-off} (a common narrative in the literature) by considering our task-dependent functional diversity with a task-dependent quality measure~\citep{linwildbench, wei2025rocketeval}. Figure~\ref{fig:dq_tradeoff} shows that the diversity-quality tradeoff may simply be the result of mis-conceptualizing both diversity and quality. Our evaluation framework corrects both.    
\end{enumerate}

\section{Background}
\subsection{Homogenization Evaluations}\label{subsec:background_evals}

LLMs exhibit output homogenization across many tasks, including creative writing~\citep{moon2024homogenizing, wu2025generative}, political discussions~\citep{durmus2023towards, santurkar2023whose}, math problem-solving~\citep{slocumdiverse}, and question-answering~\citep{zhang2024forcing}. Because these evaluations are typically domain-specific, what counts as problematic\footnote{The pluralistic alignment literature~\citep{sorensen2024position, zhang2025cultivating} highlights representation harms from homogenization, particularly when users seek advice or opinions from LLMs. However, these discussions focus on tasks where representation is presumed to be desirable.} homogenization is task-dependent. Multi-domain benchmarks such as \textsc{Community-Alignment}~\citep{zhang2025cultivating}, \textsc{Novelty-Bench}~\citep{zhang2025noveltybench}, and \textsc{Infinity-Chat}~\citep{jiang2025artificial} broaden coverage but focus on open-ended prompts; thus, they cover only part of our taxonomy, which spans reward verifiability and includes tasks where some homogenization is desirable.

Prior work often uses general diversity metrics -- \textit{vocabulary diversity} (word overlap) and \textit{embedding diversity} (cosine distance) -- that may miss task-dependent differences (e.g. distinct problem-solving strategies or writing narratives). Recent work~\citep{zhang2025noveltybench, shypula2025evaluating} instead studies \textit{functional diversity}: whether two responses are meaningfully different to a user for a given prompt. Our taxonomy operationalizes this by specifying distinct notions of functional diversity per task category.

We focus on intra-model homogenization; related work studies inter-model homogenization (``Algorithmic Monoculture'' or ``Artificial Hivemind'')~\citep{kim2025correlated, wenger2025we, jiang2025artificial, bommasani2022outcome}. While our taxonomy may extend to that setting, we prioritize intra-model effects since reducing within-model homogenization may also reduce cross-model homogenization~\citep{jain2024position, jain2024algorithmic, jain2025allocation} (c.f. Figure~\ref{fig:metaphor_time}).

\subsection{Diversity-Promoting Methods}\label{subsec:background_promoting}

Methods to reduce homogenization include modifying preference data or alignment objectives~\citep{lanchantin2025diverse, slocumdiverse, chung2025modifying}. These often increase post-alignment diversity as measured by token-level entropy or embedding variation, though such metrics may not reflect meaningful, task-dependent diversity.

At inference time, prompting can elicit diverse outputs beyond temperature sampling: \citet{zhang2025cultivating} use a \textit{system prompt} to generate $k$ responses reflecting ``diverse values,'' and \citet{zhang2025noveltybench} propose \textit{in-context regeneration}, prompting a different response while retaining prior responses in context. Persona-based and multilingual prompting may also elicit response variation~\citep{wang2025large, wang2025multilingual}. We improve inference-time sampling by explicitly specifying the intended notion of ``diversity'' in model instructions.

\begin{table*}[h!]
    \centering
    \resizebox{\textwidth}{!}{
    \begin{tabular}{p{0.1cm}p{3.0cm} p{4.1cm} p{3.3cm} p{2.7cm} p{1.6cm}p{3.0cm}}
        \toprule
        \multicolumn{2}{l}{\textbf{Task Category}} & \textbf{Task Definition} & \textbf{Example Task} & \textbf{Functional\newline Diversity} & \textbf{Reward\newline Type} & \textbf{Example\newline Dataset} \\
        \midrule
        A. & Well-Specified\newline Singular Objective & Tasks with one verifiably correct answer & What is the largest\newline Spanish-speaking country? & None  & Verifiable & SimpleQA\newline \cite{wei2024simpleQA}\\
        \midrule
        B. & Underspecified\newline Singular Objective & Tasks with many verifiable correct answers & Name one Spanish-\newline speaking country. & Different correct\newline answers  & Verifiable (Multiple) & NoveltyBench\newline \cite{zhang2025noveltybench}\\
        \midrule
        C. & Random\newline Generation & Tasks that involve randomizing over a set of options & Roll a make-believe\newline 6-sided dice. & Different pseudo-\newline random options & Verifiable (Multiple) & NoveltyBench\newline \cite{zhang2025noveltybench}\\
        \midrule
        D. & Problem-Solving\newline Objective & Tasks to solve a problem with a verifiable solution & How many divisors does 196 have? & Different solution\newline strategies & Verifiable & MATH-500 \cite{lightman2023let}\\
        \midrule
        E. & Problem-Solving or\newline Design Subjective & Tasks to solve a problem with many partially\newline verifiable solutions & Design a room for comfort \&  minimal energy consumption. & Different solution strategies & Partially \newline Verifiable & MacGyver\newline \cite{tian2024macgyver}\\
        \midrule
        F. & Encyclopedia\newline Inquiry & Tasks to provide information about real-world societies \& history where there are credible references & Why is Isaac Newton\newline famous? & Different factual \newline perspectives  & Partially\newline Verifiable & Community \newline Alignment\newline \cite{zhang2025cultivating}\\
        \midrule
        G. & Creative Writing & Tasks that require creative\newline expression & Tell me a riddle. & Different creative\newline elements & Non-Verifiable & WildBench\newline \cite{linwildbench}\\
        \midrule
        H. & Advice or Opinions & Tasks that solicit advice,\newline opinions or feedback on\newline specific topics/scenarios & What is a good\newline Mother's day gift? & Different views\newline  or perspectives & Non-Verifiable & InfinityChat  \newline \cite{jiang2025artificial}\\
        \bottomrule
    \end{tabular}
    }
    \caption{\textbf{Task Taxonomy.} Categories have unique concepts of functional diversity and fall along a spectrum of reward verifiability. While not exhaustive, our taxonomy provides a mechanism for clarifying what differences are meaningful across responses to a prompt.} 
    \label{tab:task_categories}
\end{table*}

\section{Task-Dependent Framework}

\subsection{A Task Taxonomy of Functional Diversity}\label{subsec:framework_taxonomy}

We build on the notion of \textit{functional diversity}~\citep{zhang2025noveltybench, shypula2025evaluating}, or whether a human user would perceive two LLM responses to a query as meaningfully different. Previous works do not specify what constitutes a \textit{meaningful} difference, implicitly assuming it is intuitive for human users. However, LLM-judges struggle to evaluate functional diversity~\citep{zhang2025noveltybench}, and what constitutes a meaningful difference varies significantly across tasks. We hypothesize that common homogenization metrics such as vocabulary-based or embedding-based diversity may not capture functional diversity.

We introduce a taxonomy of $8$ task categories, each corresponding to a specific type of functional diversity (Table~\ref{tab:task_categories}). Categories help clarify: what elements of a response could be meaningfully different for a given task. For example, in \textit{Problem Solving Objective} (category D), responses may have different solution strategies, but not necessarily different final answers. In \textit{Creative Writing} (category G), responses may have different creative elements (e.g. plot, genre, setting), beyond vocabulary differences. The task categories are motivated by recent studies evaluating homogenization in specific task domains (\S~\ref{subsec:background_evals}, Appx. Table~\ref{tab:taxonomy_related_works}). Unlike previous work, we also consider task categories where homogenization is expected.

The \textit{reward verifiability} of a task helps clarify which elements of responses should remain homogeneous. Our taxonomy offers a more granular spectrum\footnote{We view verifiable/non-verifiable and objective/subjective not as discrete categories, but as spectra that task categories span. While the terms ``objective'' and ``subjective'' invite philosophical debate, such discussions are beyond this paper’s scope. Likewise, we use ``underspecified'' and ``well-specified'' in category names as clarifying terms, not precise definitions of the answer spaces.} of reward verifiability than the binary distinction in prior work~\citep{lambert2024tulu, lanchantin2025bridging}. The first four categories (A, B, C, D) capture prompts that elicit \textit{verifiable} solution(s) that might be considered objective in nature, yet may still have more than one verifiable answer or explanation. The second four categories (E, F, G, H) capture prompts that elicit more open-ended solution(s) that may be \textit{non-verifiable} or only have partially verifiable components. 

\subsection{Task-Dependent Sampling Technique}\label{subsec:framework_sampling}

We introduce a \textit{task-dependent sampling technique} that modifies existing prompt-based methods for promoting diversity (Table~\ref{tab:prompt_based_strategies}). We focus on improving two such methods: \textit{system prompt sampling}, which generates multiple responses in a single generation~\citep{zhang2025cultivating}, and \textit{in-context regeneration}, which iteratively generates multiple responses~\citep{zhang2025noveltybench}. These methods instruct the model to generate ``different'' or ``diverse'' responses, without clarifying what this means for a given task.

Our technique has two steps. The first step is to classify the input prompt into a task category (Appx.~\ref{apx:taxonomy_judge}). The second step involves augmenting the prompt with the corresponding functional diversity criteria from our taxonomy (Appx. Table~\ref{tab:prompt_based_strategies_all}). In practice, if a prompt falls outside the taxonomy, our approach can generalize to other categories, or the model may resume its default behavior. To reduce homogenization at inference-time, the model could sample over resulting responses, or select one based on other alignment criteria.

\begin{table*}[h!] 
\centering
\scriptsize
\resizebox{\textwidth}{!}{
\begin{tabular}{p{1.65cm} p{2.5cm} p{4.1cm}p{6.7cm}}
\toprule
\textbf{Method} & \textbf{Previous Works} & \textbf{Problem-Solving Objective} & \textbf{Creative Writing}  \\ 
& (Not Task-Based) & (Task-Dependent) & (Task-Dependent) \\
\toprule
\addlinespace
 System Prompt\newline Sampling & Generate \texttt{\{N\}}\newline responses that represent diverse values.\newline \citep{zhang2025cultivating} & The following problem has a single correct answer, but can be solved using  different problem-solving strategies. Generate \texttt{\{N\}} different solutions, each with a different problem-solving strategy. & The following prompt is asking for creative expression, so there are many possible subjective responses. Generate \texttt{\{N\}} unique responses by varying the key creative elements such as tone, genre, point of view, theme, structure, etc. Each response should have different creative elements and reflect a distinct creative expression.   \\
\addlinespace
\midrule
\addlinespace
 In-Context Regeneration & Can you generate a\newline different answer?\newline ~\citep{zhang2025noveltybench} & Can you solve the problem using a different strategy? The problem has a single correct answer, but can be solved using different problem-solving strategies. & Can you generate a new response with different creative elements? The prompt is asking for creative expression, so there are many possible subjective responses. Your new response should change the key creative elements such as tone, genre, point of view, theme, structure, etc. \\
\addlinespace
\bottomrule
 \end{tabular}}
\caption{\textbf{Prompt-Based Sampling Strategies.} We modify prompt-based sampling methods in previous works~\citep{zhang2025cultivating, zhang2025noveltybench} to promote task-dependent functional diversity.}
\label{tab:prompt_based_strategies}
\end{table*}

\section{Human Perception of Functional Diversity}\label{sec:human_validation}
We conduct a small user study ($n=22$) to validate that the taxonomy aligns with human perception of functional diversity. Annotators were randomly assigned to label either \textit{general} functional diversity (i.e. whether two responses are meaningfully different) or \textit{task-dependent} functional diversity (i.e. whether two responses are meaningfully different according to our taxonomy). $11$ annotators in each group independently labeled $105$ response pairs. We compare human annotations with general diversity metrics as well our task-dependent evaluation approach. 

\begin{table}[t!]
\scriptsize
\centering
\begin{tabular}{lcccccccc}
\toprule
\multirow{2}{*}{Metric} &  \multicolumn{8}{c}{Task Category} \\
\cline{2-9}
\addlinespace[0.3em]
 & All & A & B & C & E & F & G & H \\
\midrule
Inter-Rater Agreement & 0.90 & 0.99 & 0.99 & 0.87 & 0.92 & 0.85 & 0.74 & 0.92 \\
\midrule
\multicolumn{9}{l}{\uline{General Diversity Metrics}} \\
\addlinespace[0.1em]
Vocabulary Diversity & 0.63 & 0.25 & \cellcolor{lightred} 0.51 & \cellcolor{lightred} 0.44 & \cellcolor{lightred} 0.46 & 0.69 & 0.70 & 0.69 \\
Embedding Diversity & 0.78 & 0.43 & 0.82 & 0.59 & 0.66 & 0.79 & 0.79 & 0.79 \\
Compression Diversity &  \cellcolor{lightred} 0.28 & \cellcolor{lightred} 0.06 & 0.62 & 0.45 & 0.62 & 0.72 & \cellcolor{lightred} 0.19 & \cellcolor{lightred} 0.59 \\
\textsc{Novelty-Bench} Functional Diversity & 0.60 & 0.37 & \cellcolor{lightred} 0.13 & 0.45 & 0.69 & \cellcolor{lightred} 0.65 & 0.66 & \cellcolor{lightred} 0.51 \\
\midrule
\multicolumn{9}{l}{\uline{Task-Dependent Functional Diversity (Taxonomy-Guided LLM-Judges)}} \\
\addlinespace[0.1em]
\textbf{(ours)} GPT-4o & 0.90 & \cellcolor{lightgreen} 1.00 & \cellcolor{lightgreen} 1.00 & \cellcolor{lightgreen} 1.00 & 0.83 & 0.76 & 0.74 & \cellcolor{lightgreen} 1.00 \\
\textbf{(ours)}  Gemini-2.5-Flash & 0.92 &\cellcolor{lightgreen} 1.00 & \cellcolor{lightgreen} 1.00 & \cellcolor{lightgreen} 1.00 & \cellcolor{lightgreen} 1.00 & \cellcolor{lightgreen} 0.87 & 0.67 & \cellcolor{lightgreen} 1.00 \\
\textbf{(ours)}  Claude-4-Sonnet & \cellcolor{lightgreen} 0.94 & \cellcolor{lightgreen} 1.00 &  \cellcolor{lightgreen} 1.00 & \cellcolor{lightgreen} 1.00 & 0.83 & \cellcolor{lightgreen} 0.87 & \cellcolor{lightgreen} 0.87 & \cellcolor{lightgreen} 1.00 \\
\bottomrule
\end{tabular}
\caption{\textbf{Taxonomy-guided LLM-judges correlate strongly with human perception of functional diversity, outperforming general diversity metrics.} 
We report Spearman correlation between each metric and the majority vote from human annotation of general functional diversity (unaided by our taxonomy). Green (red) cells indicate the highest (lowest) correlations. Across 105 response pairs, taxonomy-guided LLM-judges outperform general diversity metrics. Note that correlations of $1.00$ for LLM-judges in categories A, B, C \& H reflect perfect agreement with the majority human vote for binary judgments on 15 response pairs, likely because these categories have simpler functional diversity concepts. Correlations are lower for partially- and non-verifiable categories (E, F, G), where human annotators also have lower inter-rater agreement (proportion of matching annotations). 
}
\label{tab:annotation_res_main}
\end{table}

\subsection{Human Validation Experiment Details}
\label{subsec:human_val_details}
\textbf{Datasets \hspace{0.25em}} We evaluate $105$ response pairs ($5$ prompts per task category).\footnote{We omit \textit{Problem-Solving Objective} (Category D) since annotators did not have math proficiency.} We sample prompts from the following datasets: \textit{Community Alignment}~\citep{zhang2025cultivating}, \textit{MacGyver}~\cite{tian2024macgyver}, \textit{NoveltyBench}~\citep{zhang2025noveltybench}, \textit{SimpleQA}~\citep{wei2024simpleQA}, and \textit{WildBench}~\citep{linwildbench}. These datasets were chosen to achieve reasonable coverage across the task categories in our taxonomy. We determine the ``ground-truth'' task category for each prompt based on the source dataset. Appx.~\ref{apx:datasets} \& \ref{apx:taxonomy_judge} report further details.

\textbf{Models \hspace{0.25em}} Annotators judge responses from \textit{GPT-4o}. Separately, we evaluate \textit{GPT-4o}, \textit{Claude-4-Sonnet}, and \textit{Gemini-2.5-Flash} as LLM-judges of functional diversity.

\textbf{Sampling Strategies \hspace{0.25em}} We sample response pairs using temperature sampling, system prompt sampling, and in-context regeneration (all use temperature $t=1.0$). System prompt sampling and in-context regeneration use our task-dependent sampling technique ($\S$~\ref{subsec:framework_sampling}). The task category used in our sampling technique is determined by the model.

\textbf{Human Annotation \hspace{0.25em}} Annotators were shown each prompt, two responses, and a definition of functional difference. Based on this definition, annotators were asked to label response pairs as ``same'' or ``different''. Annotators assigned to label \textit{general} functional diversity were instructed to mark different if ``a user would likely perceive the responses to be meaningfully different for the given prompt''. Annotators assigned to label \textit{task-dependent} functional diversity were dynamically shown the relevant functional diversity concept from our taxonomy based on the ``ground-truth'' task category. Appx.~\ref{apx:human_annotation} reports further details.

\textbf{Diversity Metrics \hspace{0.25em}} We compute five diversity metrics: vocabulary diversity (Def.~\ref{def:vocab_div}), embedding diversity (Def.~\ref{def:embed_div}), Gzip compression diversity (Def.~\ref{def:compression_div}), general functional diversity (Def.~\ref{def:gen_fun_div}), and task-dependent functional diversity (Def.~\ref{def:fun_div}). To calculate general functional diversity, we use the \textsc{Novelty-Bench} semantic classifier which is a fine-tuned \textit{deberta-v3-large} model~\citep{zhang2025noveltybench}. To calculate task-dependent functional diversity, we use taxonomy-guided LLM-judges given the ``ground-truth'' task category (details in Appx.~\ref{apx:div_metrics}). To compute embedding diversity, we use the \textit{gemini-embedding-001} model.

\subsection{Human Validation Experiment Results}

\textbf{Taxonomy-guided and unguided human judgments largely agree.} We report correlation between (i) annotators asked to judge \textit{general} functional difference and (ii) annotators asked to judge \textit{task-dependent} functional difference using our taxonomy. Based on the majority vote from the $11$ annotators in each group, the two annotation protocols are highly correlated ($r_s=0.90$). This suggests that the distinctions captured by our taxonomy largely coincide with what annotators that are not guided by the taxonomy inherently consider ``meaningfully different''. Within each group, inter-rater reliability is also high (Krippendorff's $\alpha=0.67$; 90\% agreement rate).

\textbf{Taxonomy-guided LLM-judges have strong correlation with human perception of functional diversity.} As Table~\ref{tab:annotation_res_main} shows, the Spearman correlation is 0.90 or higher between taxonomy-guided LLM-judges and human labels of general functional diversity (based on majority vote). Correlation is highest for task categories with verifiable components, and slightly declines for partially verifiable or non-verifiable task categories. 

\textbf{Taxonomy-guided LLM-judges outperform general diversity metrics.} The Spearman correlation with human annotation is lower for general diversity metrics (Table~\ref{tab:annotation_res_main}). This suggests that human perception of functional diversity is inherently task dependent, and that applying a uniform diversity metric across tasks may miss distinctions that humans consider meaningful. Embedding-based diversity and the \textsc{Novelty-Bench} semantic classifier have stronger correlation with human judgments for categories with non-verifiable rewards where general semantic variation may align with task-based diversity. 

\section{Evaluating \& Promoting Task-Dependent Diversity}

\begin{figure}[t!]
  \centering
  \includegraphics[width=0.93\linewidth]{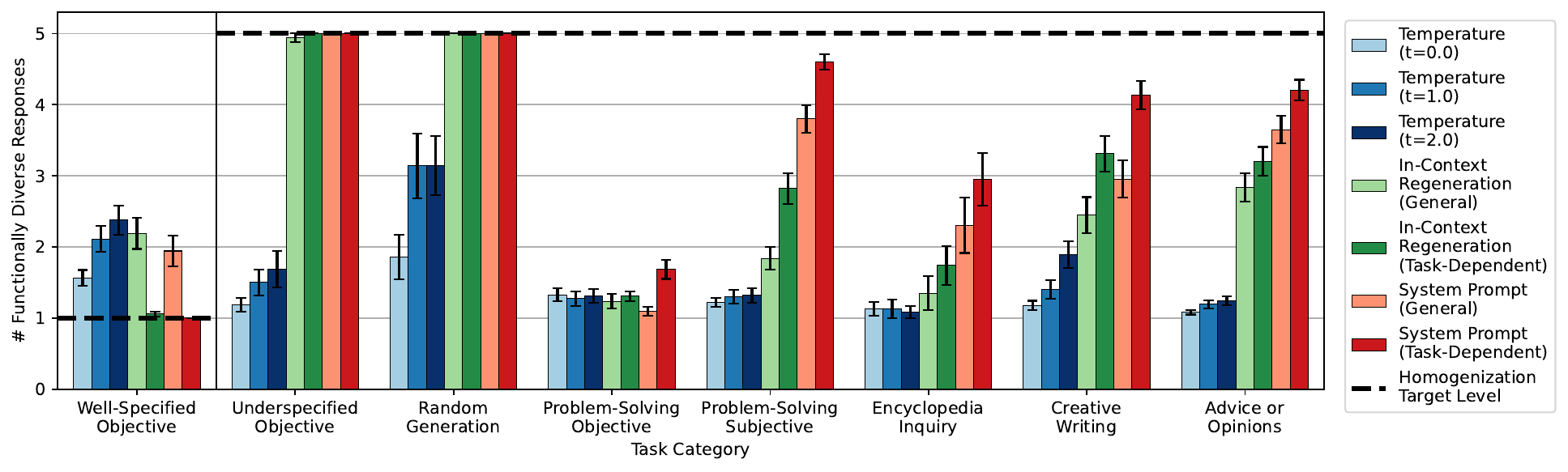}
  \caption{\textbf{Our task-dependent sampling increases functional diversity for task categories where homogenization is undesired, while preserving homogenization where it is desired.} We plot the average number of functionally diverse responses generated by GPT-4o for each sampling strategy and task category (with standard error). For the first category (Well-Specified Objective), bars closer to $1$ preserve output homogenization that is expected. For all other categories, bars closer to $5$ represent maximum functional diversity.}
  \label{fig:fun_div_gpt}
\end{figure}

In this section, we evaluate task-dependent homogenization across the verifiability spectrum in our taxonomy, showing that task-dependent sampling increases functional diversity where it is desirable while preserving homogenization where appropriate (Figure~\ref{fig:fun_div_gpt}). We also find that the diversity-quality trade-off becomes negligible when both measurement constructs are defined in a task-dependent way (Figure~\ref{fig:dq_tradeoff}).

\subsection{Experiment Details}\label{subsec:exp_details}

\textbf{Datasets \hspace{0.25em}} We evaluate $n=344$ prompts from the same variety of six datasets used in our human validation experiment (c.f. Appx~\ref{apx:datasets}). Additionally, we evaluate $n=100$ prompts from \textsc{Infinity-Chat}~\citep{jiang2025artificial} (originally from WildChat) to assess whether our framework generalizes to a popular homogenization benchmark; we report these results separately since \textsc{Infinity-Chat} was released concurrently with our work (c.f. Appx~\ref{apx:infinity-chat-res}).

\textbf{Models \hspace{0.25em}} We evaluate responses from five models: \textit{GPT-4o}, \textit{Claude-4-Sonnet}, \textit{Gemini-2.5-Flash}, \textit{Llama-3.1-8B-Instruct}, and \textit{Mistral-7B-Instruct-v0.3}. Separately, we use \textit{GPT-4o}, \textit{Claude-4-Sonnet}, and \textit{Gemini-2.5-Flash} as LLM-judges (independent of response generation). When reporting LLM-judge metrics, we average the outputs across these three judge models.

\textbf{Sampling Strategies \hspace{0.25em}} We evaluate temperature sampling with 3 levels (low/medium/high). We further evaluate both general and task-dependent approaches to system prompt sampling and in-context regeneration. The general approach is based on previous work ~\citep{zhang2025cultivating, zhang2025noveltybench}. For each strategy, we sample $5$ responses per prompt.

\textbf{Diversity Metrics \hspace{0.25em}} Given the human validation results in $\S$~\ref{sec:human_validation}, we use taxonomy-guided LLM-judges to calculate task-dependent functional diversity. We then determine the \textit{number of functionally diverse responses} (Def~\ref{def:fun_div_clusters}) out of the 5 responses generated per prompt and sampling strategy. We also compare to vocabulary and embedding diversity.

\textbf{Quality Metrics \hspace{0.25em}} We evaluate quality via (1) \textit{Athene-RM-8B} reward model scores~\citep{frick2025athene} and (2) LLM-judges with task-based grading checklists~\citep{linwildbench, wei2025rocketeval}, which generate prompt-specific grading criteria and score responses on a Likert scale from 1 to 5. We review all generated checklists; more details and examples in~\ref{apx:quality_prompts}. 

\subsection{Evaluating Functional Diversity Across Tasks \& Sampling Strategies}

We find that our task-dependent sampling technique outperforms temperature-sampling and the general sampling techniques in previous work~\citep{zhang2025cultivating, zhang2025noveltybench}. Figure~\ref{fig:fun_div_gpt} shows how we significantly increase functional diversity for task categories where homogenization is undesired, while preserving homogenization where it is desired (for GPT-4o, all results in Appendix~\ref{apx:exp_results}). Below, we explore results across task categories.

\textbf{Well-Specified Objective Tasks (Category A) \hspace{0.25em}} Tasks in this category have a single verifiably correct answer; thus, no diversity is expected. However, higher temperatures and general diversity-promoting methods undesirably reduce homogenization, generating multiple unique answers ($2$ on average). In contrast, our task-dependent sampling method maintains homogenization, consistently producing one unique answer per task for GPT-4o, Gemini-2.5-Flash, and Claude-4-Sonnet.

\textbf{Underspecified Objective \& Random Generation Tasks (Categories B \& C) \hspace{0.25em}} These tasks have many verifiably correct answers, which suggests that models may easily conceptualize functional diversity. Accordingly, we observe no significant differences between task-dependent and general prompt-based methods. 

\textbf{Problem-Solving Objective Tasks (Category D) \hspace{0.25em}} These tasks have a single verifiably correct answer, but multiple valid solution strategies. In this setting, general prompt-based sampling does not yield responses with diverse solution strategies. In contrast, our task-dependent methods are able to generate approximately 2–3 distinct solution strategies. This relatively low number compared to other categories may be due to the difficulty in obtaining even a single correct solution in the MATH-500 benchmark~\citep{hendrycks2021measuring}.

\textbf{Partial and Non-Verifiable Tasks (Categories E, F, G, H) \hspace{0.25em}} Tasks in these categories cover more open-ended queries typical of homogenization studies. Across all five models, both our task-dependent sampling methods reduce homogenization more than their respective general approaches. For GPT-4o, Gemini-2.5-Flash, and Mistral-7B-Instruct-v0.3, task-dependent system prompting yields the highest number of functionally diverse responses. In contrast, for Claude-4-Sonnet and Llama-3.1-8B-Instruct, both task-dependent methods demonstrate comparable performance in promoting diversity. Among temperature-sampled outputs, smaller open-weight models tend to have less homogenization than larger commercial models, possibly due to less extensive alignment. We observe similar results in our supplementary analysis of \textsc{Infinity-Chat} (\ref{apx:infinity-chat-res}), suggesting that task-dependent sampling may reduce the ``hivemind'' effect\footnote{Appendix~\ref{apx:prompt_illustrations} 
provides prompt-level illustrations of embedding diversity across open-ended task categories in \textsc{Infinity-Chat}, including examples prompts with implicit fidelity constraints.}  when measured by functional diversity.

\begin{figure}[h!]
  \centering
  \begin{subfigure}[t]{0.425\linewidth}
    \centering
    \includegraphics[width=0.9\linewidth]{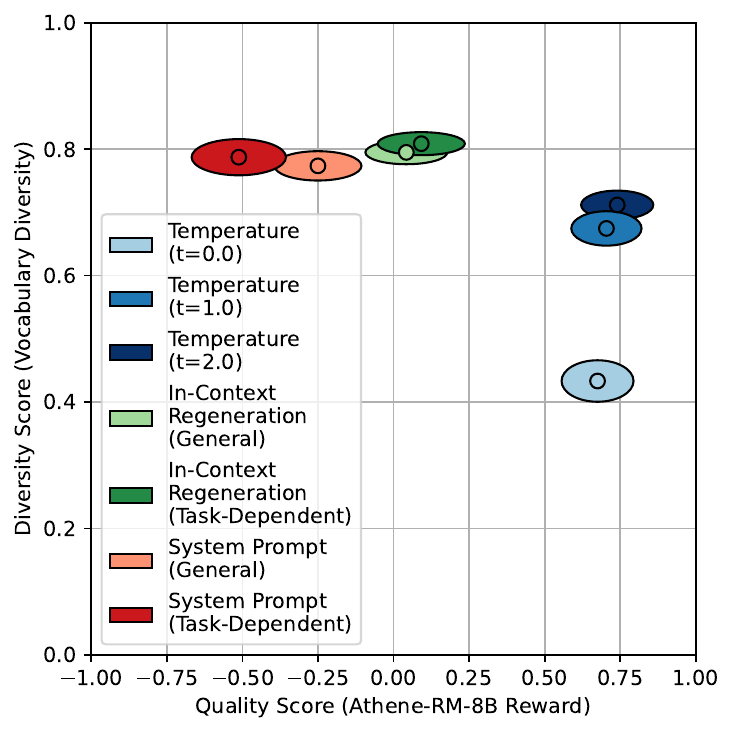}
    \caption{General Metrics}
    \label{fig:dq_general_gpt}
  \end{subfigure}
  \begin{subfigure}[t]{0.425\linewidth}
    \centering
    \includegraphics[width=0.9\linewidth]{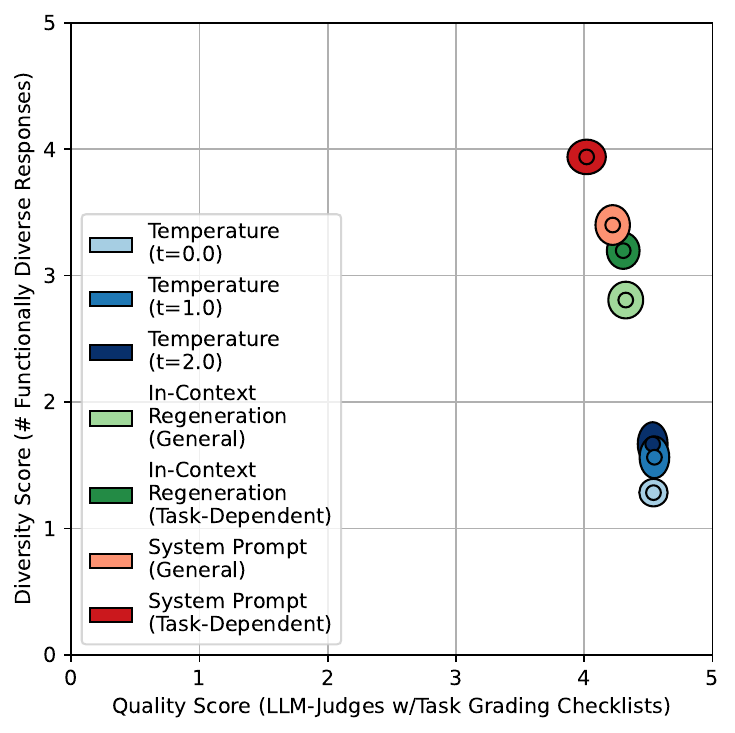}
    \caption{Task-Based Metrics}
    \label{fig:dq_ta_gpt}
  \end{subfigure}
  \caption{\textbf{With task-based metrics, diversity is improved with no significant drop in quality.} We plot quality on the $x$-axis and diversity on the $y$-axis and compare the tradeoff under general metrics vs task-based metrics. In (a), there is a large tradeoff between vocabulary diversity and reward-based quality. In (b), there is a negligible tradeoff between task-dependent functional diversity and LLM-judges with task-based grading checklists. Note the checklist-based quality difference between scores $4$ \& $5$ is ``good'' vs ``very good''. Plots show the mean \& standard error of metrics averaged across all task categories except category A, which we exclude because homogenization is desired for that category.}
  \label{fig:dq_tradeoff}
\end{figure}

\subsection{Diversity-Quality Tradeoff? Comparing General \& Task-Based Metrics}

We find that improved functional diversity from our task-dependent sampling often maintains the quality of responses, when the quality metric\footnote{For tasks with singular verifiable rewards (Simple-QA \& MATH-500), we separately validate accuracy (Appx. Table~\ref{tab:accuracy}). Overall, our task-dependent sampling approaches often maintain and sometimes improve accuracy compared to temperature sampling. For MATH-500, system prompt sampling has the best accuracy for all models except Mistral. For Simple-QA, in-context regeneration performs the best for Gemini \& Mistral, while system prompt sampling is the best for Claude \& Llama.} is also task-dependent. Recent proposals for measuring quality using task-based checklists align with our discussion of task-dependent diversity~\citep{linwildbench, wei2025rocketeval}. Whereas, when quality is determined by a reward model, scores do not explicitly reflect task differences (e.g. the quality of a creative writing response is measured the same way as a math problem-solving response). For GPT-4o responses, Figure~\ref{fig:dq_tradeoff} shows how the diversity-quality tradeoff is exacerbated by task-agnostic measures of both diversity and quality. With task-dependent metrics, the diversity-quality tradeoff is small across all models, but more noticeable for open-weight models (c.f. Appx. Figures~\ref{fig:dq_claude}-\ref{fig:dq_mistral}). One reason is that open-weight models have lower task classification accuracy ($\sim$50\% compared to 85\% for commercial models, see~\ref{apx:taxonomy_judge}).

\section{Discussion}

\subsection{Task Dependence Improves Homogenization Evaluation}

We developed a taxonomy of task categories that clarifies how functional diversity depends on the task. For example, evaluating homogenization in math problem-solving should measure variety in solution strategies, whereas evaluating homogenization in advice or opinions should measure variety in viewpoints or perspectives. 
We offer an alternative evaluation approach to improve homogenization benchmarks that rely on generic measures of diversity (vocabulary or embedding differences). For instance, \textit{Infinity-Chat} relies on embedding diversity, but our findings suggest that using general metrics does not capture meaningful functional diversity and may falsely show a diversity-quality tradeoff (c.f. Appx.~\ref{apx:infinity-chat-res}). We validate that our taxonomy-guided LLM-judges show strong correlation with human perception of functional diversity. However, LLM-judges have known limitations, such as not capturing uncertainty in human judgments~\citep{shi2024judging, li2025generation}. Future work may further study human perception of diversity, and adapt our taxonomy.

Our task-dependent perspective also has important implications for evaluating ``diversity loss'' in alignment and diversity-promoting methods. While previous studies show that token entropy decreases during alignment~\citep{lanchantin2025bridging}, our preliminary experiments show that functional diversity does not necessarily collapse (Appx. Figures~\ref{fig:fun_div_alignment}-\ref{fig:fun_div_alignment_steps}). We also highlight the importance of evaluating tasks across our taxonomy when analyzing diversity-promoting methods. When studies limit their evaluation to tasks where diversity is desired, there may be unintended effects (e.g. confabulations) when those methods are applied to tasks which rely on homogenization being preserved. Hence, not adopting a task-dependent approach could result in less robust evaluation and present safety or ethical concerns downstream. 

\subsection{Task Dependence Improves Homogenization Mitigation}

There are many ways to apply task dependence in mitigating homogenization. For instance, our approach could be applied at inference-time automatically: given a prompt, the model could first determine its task category and then output a response according to the appropriate form of functional diversity.

Although we focus on prompt-based strategies, our task-dependent approach may be applied to diversity-promoting alignment methods. For example, \citet{li2025jointly} propose Diversity Aware Reinforcement Learning (DARLING), which jointly optimizes for response quality and diversity using a task-agnostic diversity signal. In preliminary experiments, we show that replacing DARLING's diversity signal with our taxonomy-guided LLM-judges yields further improvements in task-dependent functional diversity over standard GRPO and the original DARLING, while also enabling the framework to suppress diversity for prompts where homogenization is desired (Appx.~\ref{apx:alignment_exp_results}; Figures~\ref{fig:fun_div_darling}-\ref{fig:dq_darling}).

Future work may further explore how to embed task-dependent homogenization considerations directly into a model's learning or reasoning process. Our task-dependent sampling strategies could be incorporated into a chain-of-thought process, with models first reasoning about task-appropriate functional diversity. A reasoning model could also be trained to directly reason about the functional diversity requirements for a given task before generating a response. With task-dependent reasoning about functional diversity, the model could avoid undesirable behavior such as increasing diversity in ways that are culturally or socially inappropriate. Ultimately, we offer a simple but important improvement to the field’s conceptualization of output homogenization by grounding it in task dependence.

\bibliographystyle{assets/plainnat}
\bibliography{paper}

\clearpage
\beginappendix

\begin{figure}[h!]
  \centering
  \begin{subfigure}[t]{0.45\linewidth}
    \centering
    \includegraphics[width=0.8\linewidth]{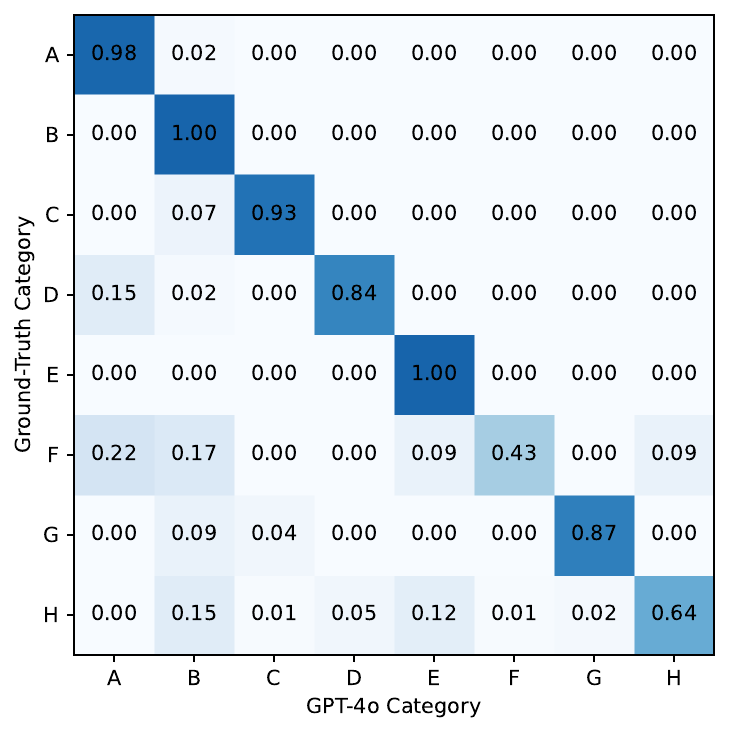}
    \caption{GPT-4o}
  \end{subfigure}
  \hfill
  \begin{subfigure}[t]{0.45\linewidth}
    \centering
    \includegraphics[width=0.8\linewidth]{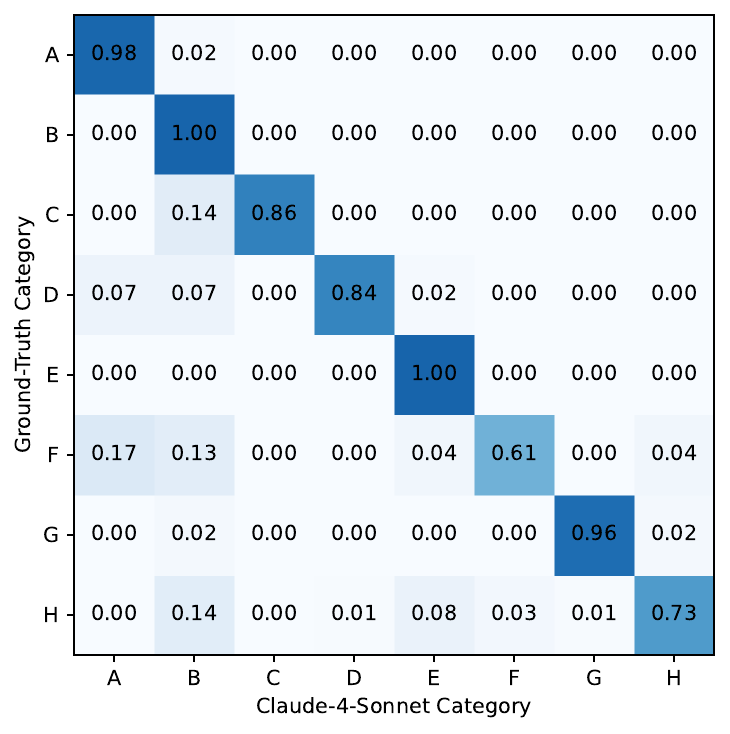}
    \caption{Claude-4-Sonnet}
  \end{subfigure}
  \\[0.5cm]
  
  \begin{subfigure}[t]{0.45\linewidth}
    \centering
    \includegraphics[width=0.8\linewidth]{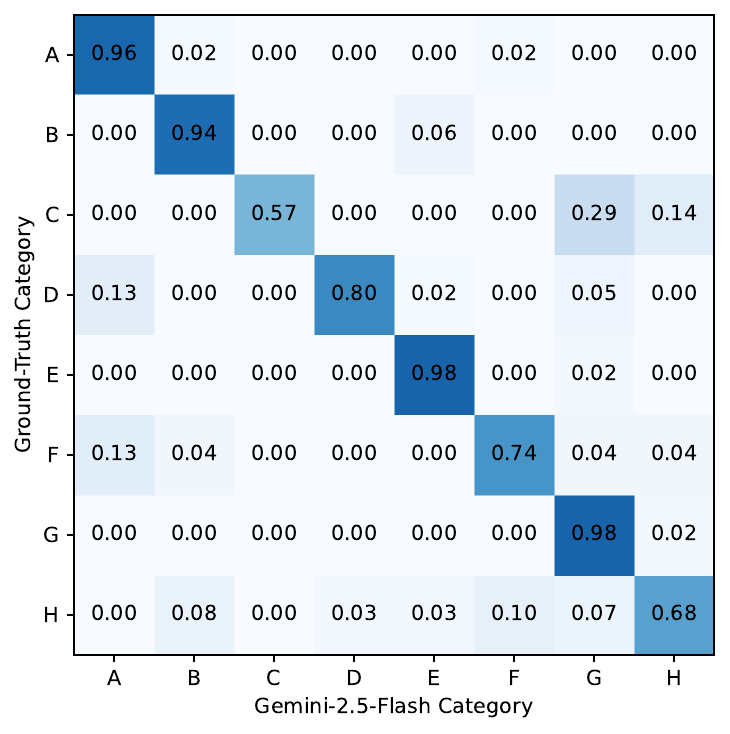}
    \caption{Gemini-2.5-Flash}
  \end{subfigure}
  \hfill
  \begin{subfigure}[t]{0.45\linewidth}
    \centering
    \includegraphics[width=0.8\linewidth]{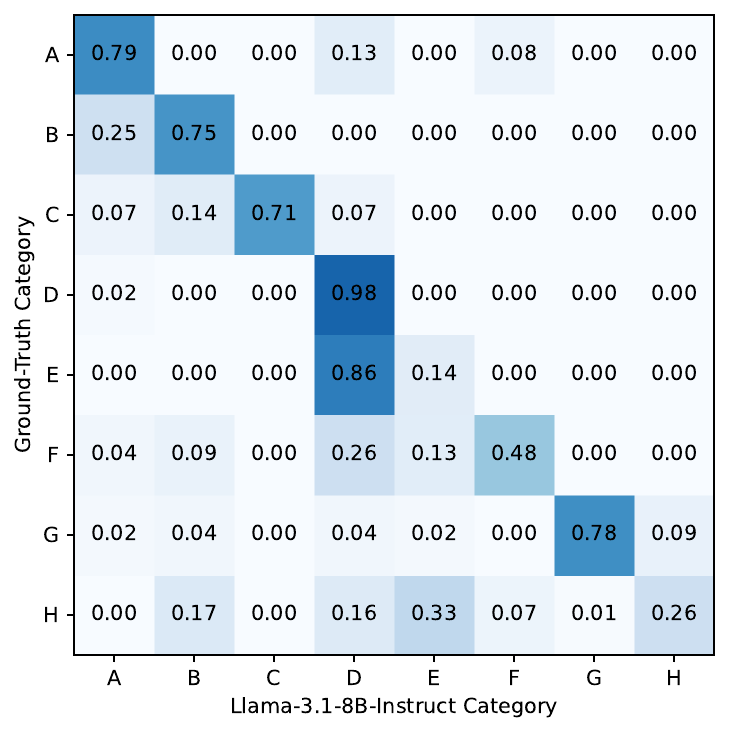}    \caption{Llama-3.1-8B-Instruct}
  \end{subfigure}
  \\[0.5cm]
  \begin{subfigure}[t]{0.45\linewidth}
    \centering
    \includegraphics[width=0.8\linewidth]{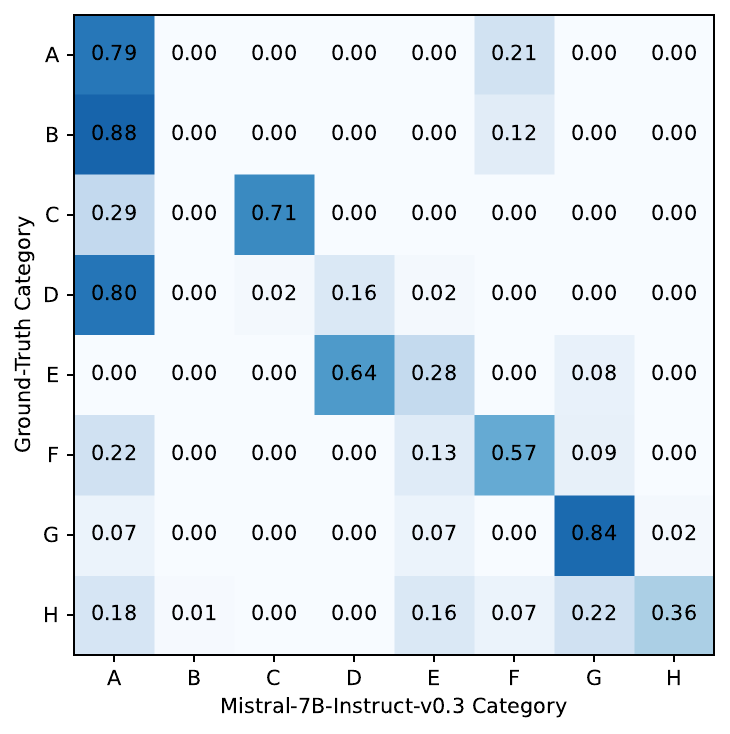}
    \caption{Mistral-7B-Instruct-v0.3}
  \end{subfigure}
  \caption{Heatmaps showing recall for models' task classification (proportion of prompts classified by the model into each task category, conditioned on each ground-truth task category).}
  \label{fig:task_categorization}
\end{figure}

Appendix~\ref{apx:exp_details} includes the following supplementary material about our experiment details.
\begin{itemize}
    \item \ref{apx:taxonomy}: Taxonomy Crosswalks
    \begin{itemize}
        \item Table~\ref{tab:taxonomy_related_works}: Crosswalk of Our Taxonomy and Previous Output Homogenization Studies
        \item Table~\ref{tab:taxonomy_gpt}: Crosswalk of Task Categories for ChatGPT Usage with Our Taxonomy
        \item Table~\ref{tab:taxonomy_claude}: Crosswalk of Task Categories for Claude Usage with Our Taxonomy
    \end{itemize}
    \item \ref{apx:datasets}: Evaluation Datasets
    \item \ref{apx:taxonomy_judge}: Task Classification Into Our Taxonomy
    \begin{itemize}
        \item Table~\ref{tab:eval_data}: Number of Prompts per Dataset and Taxonomy Category
        \item Figure~\ref{fig:task_categorization}: Recall for Models' Task Classification with Human Annotation 
    \end{itemize}
    \item \ref{apx:sampling}: Sampling Strategies
    \begin{itemize}
        \item Table~\ref{tab:prompt_based_strategies_all}: Prompts for Task-Dependent Sampling Strategies 
    \end{itemize}
    \item \ref{apx:div_metrics}: Diversity Metrics 
    \begin{itemize}
        \item Table~\ref{tab:div_prompts}: Prompts for Functional Diversity LLM-Judges
        \item Table~\ref{tab:diversity_examples}: Examples of Functionally Diverse Responses by Category
    \end{itemize}
    \item \ref{apx:human_annotation}: Human Annotation of Functional Diversity
    \item \ref{apx:quality_prompts}: Task-Dependent Quality Metric
    \begin{itemize}
        \item Table~\ref{tab:checklists}: Examples of Task-Specific Grading Checklists
        \item Table~\ref{tab:quality_examples}: Example Responses Comparing Checklist-Based Grading \& Rewards
    \end{itemize}
    \item \ref{apx:alignment-details}: Alignment Experiment Details
    \item \ref{apx:infinity-chat-details}: Infinity-Chat Experiment Details \\
\end{itemize}

\noindent Appendix~\ref{apx:exp_results} includes the following supplementary tables and figures about our experiment results.

\begin{figure}[h!]
  \centering
  \includegraphics[width=0.8\linewidth]{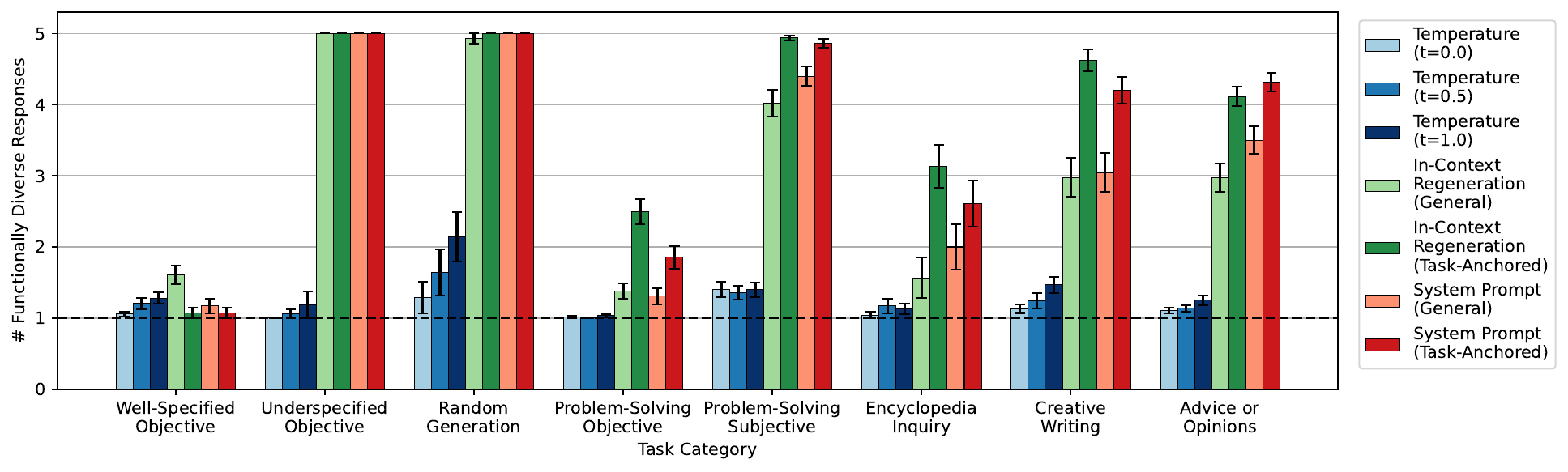}
  \caption{\centering \# of functionally diverse responses generated by \textbf{Claude-4-Sonnet} ea. sampling strategy \& task category.}
  \label{fig:fun_div_claude}
\end{figure}

\begin{figure}[h!]
  \centering
  \includegraphics[width=0.8\linewidth]{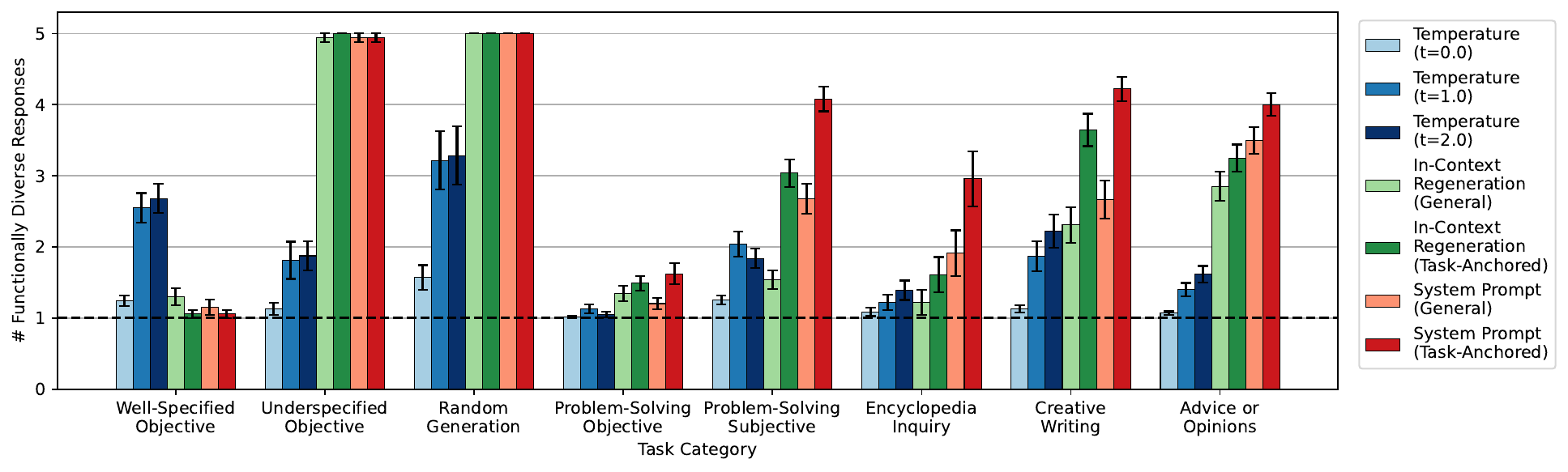}
  \caption{\centering \# of functionally diverse responses generated by \textbf{Gemini-2.5-Flash} ea. sampling strategy \& task category.}
  \label{fig:fun_div_gemini}
\end{figure}

\begin{figure}[h!]
  \centering
  \includegraphics[width=0.8\linewidth]{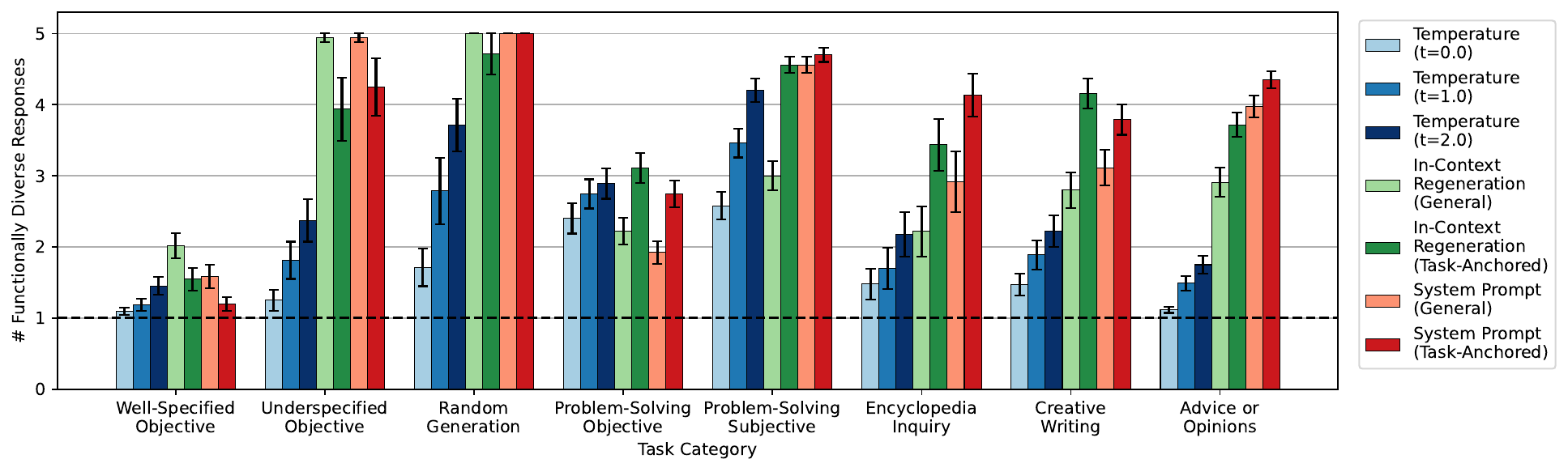}
  \caption{\centering \# of functionally diverse responses generated by \textbf{Llama-3.1-8B-Instruct} ea. sampling strategy \& task category.}
  \label{fig:fun_div_llama}
\end{figure}

\begin{figure}[h!]
  \centering
  \includegraphics[width=0.8\linewidth]{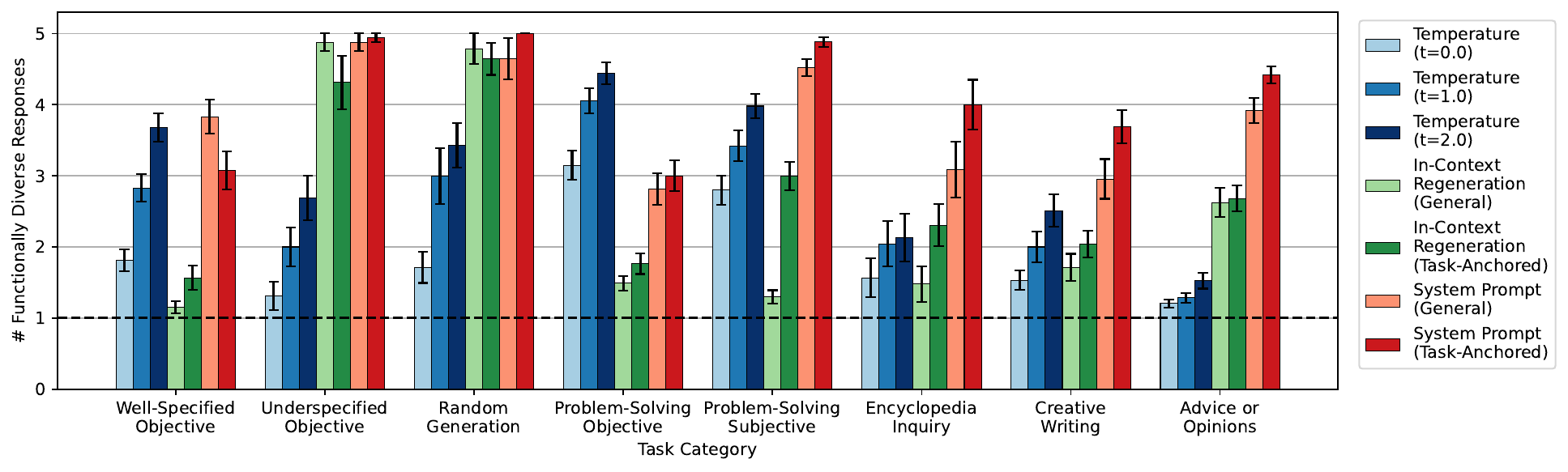}
  \caption{\centering \# of functionally diverse responses generated by \textbf{Mistral-7B-Instruct-v0.3} ea. sampling strategy \& task category.}
  \label{fig:fun_div_mistral}
\end{figure}

\begin{figure}[h!]
  \centering
  \begin{subfigure}[t]{0.42\linewidth}
    \centering
    \includegraphics[width=0.9\linewidth]{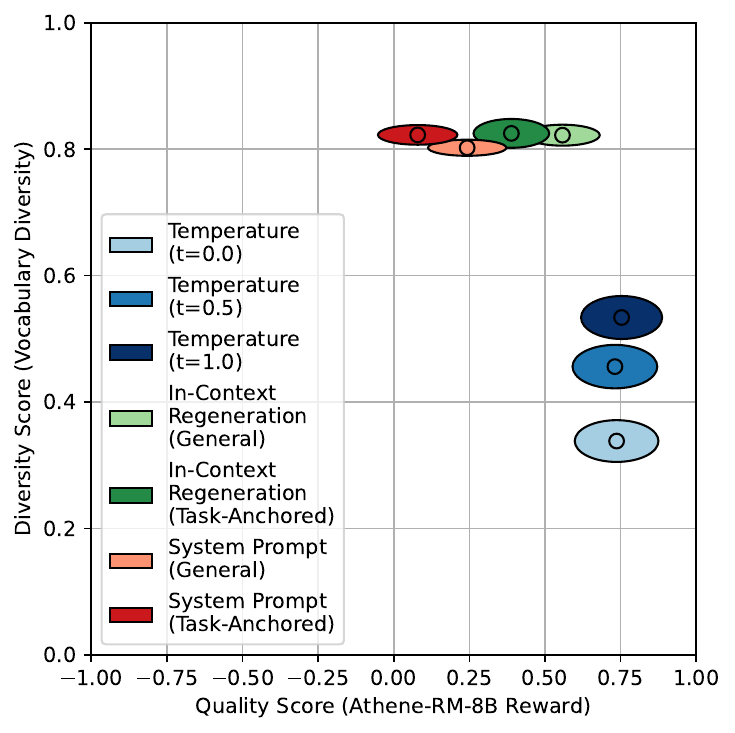}
    \caption{General Metrics}
    \label{fig:dq_general_claude}
  \end{subfigure}
  \begin{subfigure}[t]{0.42\linewidth}
    \centering
    \includegraphics[width=0.9\linewidth]{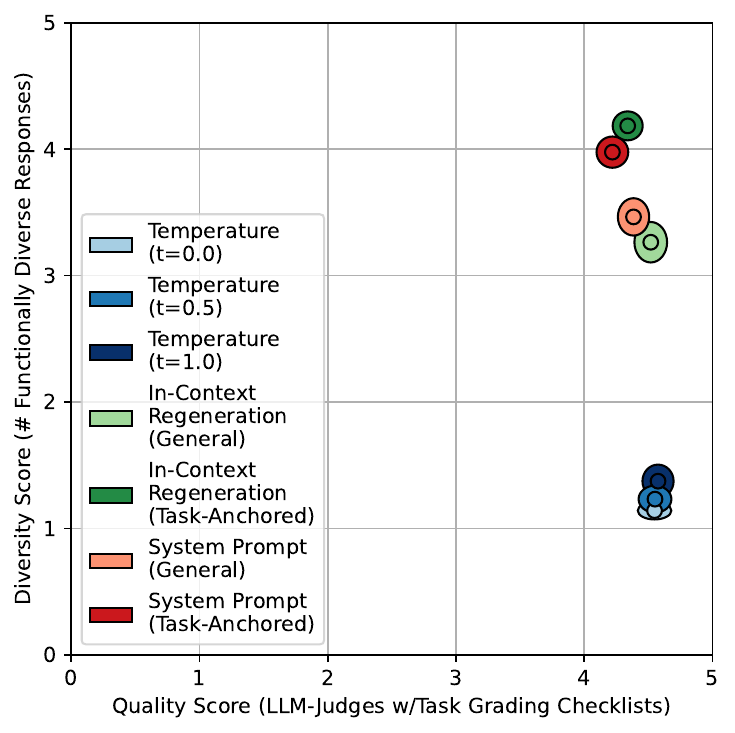}
    \caption{Task-Based Metrics}
    \label{fig:dq_ta_claude}
  \end{subfigure}
  \caption{\centering Diversity-quality tradeoff under general vs task-based metrics for \textbf{Claude-4-Sonnet}.}
  \label{fig:dq_claude}
\end{figure}

\begin{figure}[h!]
  \centering
  \begin{subfigure}[t]{0.42\linewidth}
    \centering
    \includegraphics[width=0.9\linewidth]{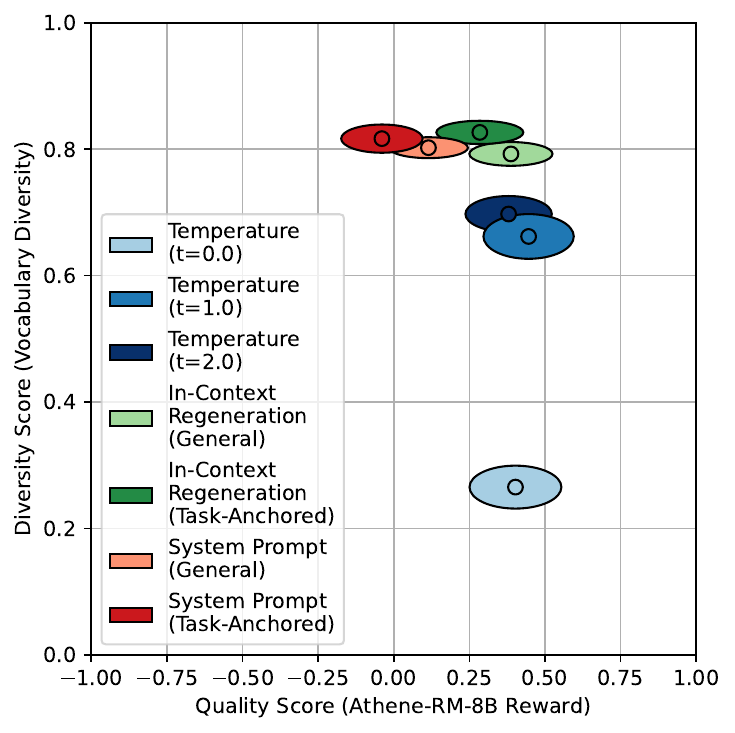}
    \caption{General Metrics}
    \label{fig:dq_general_gemini}
  \end{subfigure}
  \begin{subfigure}[t]{0.42\linewidth}
    \centering
    \includegraphics[width=0.9\linewidth]{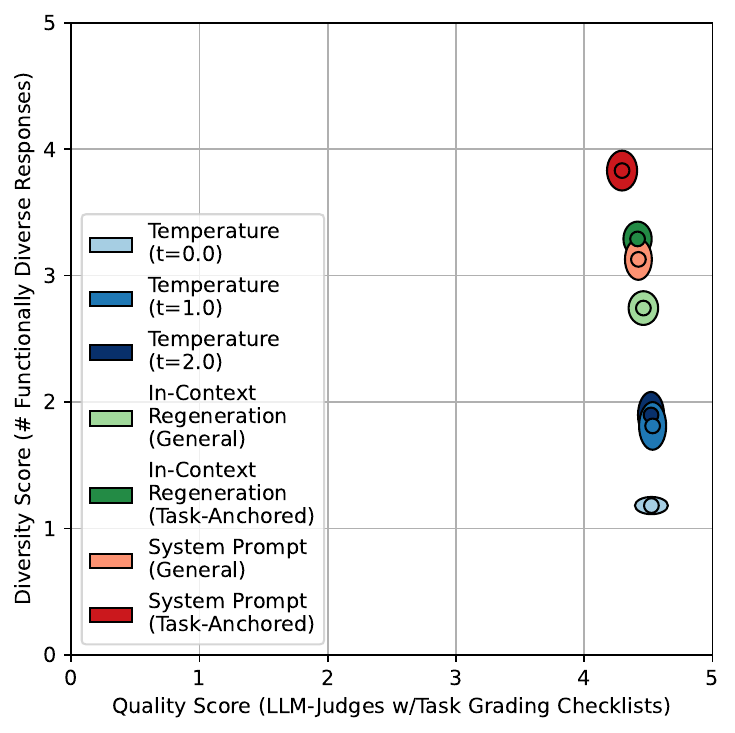}
    \caption{Task-Based Metrics}
    \label{fig:dq_ta_gemini}
  \end{subfigure}
  \caption{\centering Diversity-quality tradeoff under general vs task-based metrics for \textbf{Gemini-2.5-Flash}.}
  \label{fig:dq_gemini}
\end{figure}

\begin{figure}[h!]
  \centering
  \begin{subfigure}[t]{0.42\linewidth}
    \centering
    \includegraphics[width=0.9\linewidth]{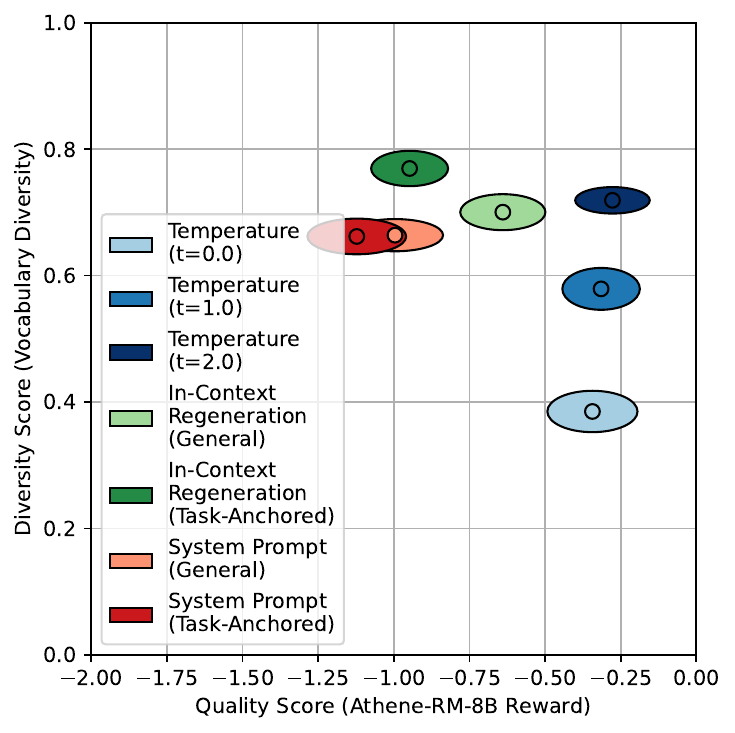}
    \caption{General Metrics}
    \label{fig:dq_general_llama}
  \end{subfigure}
  \begin{subfigure}[t]{0.42\linewidth}
    \centering
    \includegraphics[width=0.9\linewidth]{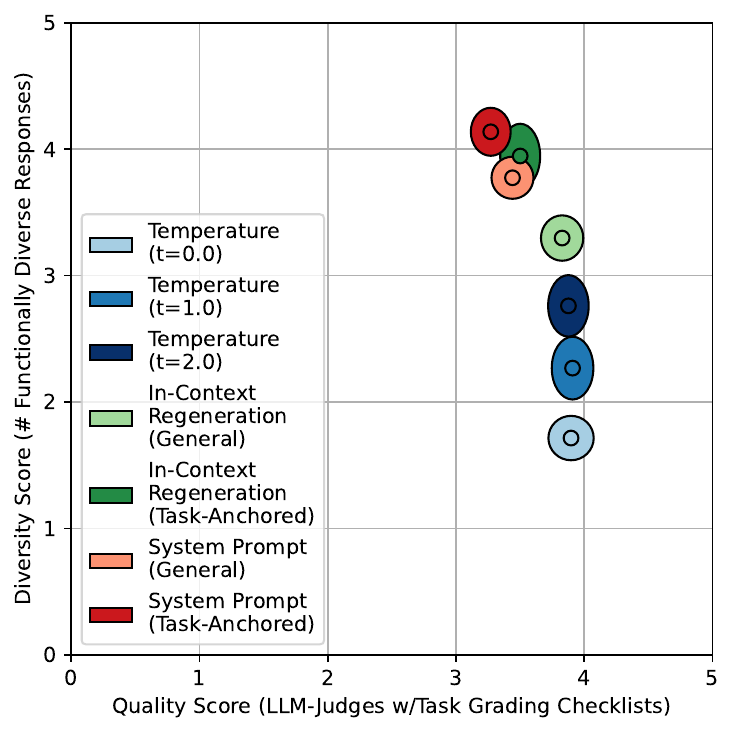}
    \caption{Task-Based Metrics}
    \label{fig:dq_ta_llama}
  \end{subfigure}
  \caption{\centering Diversity-quality tradeoff under general vs task-based metrics for \textbf{Llama-3.1-8B-Instruct}.}
  \label{fig:dq_llama}
\end{figure}

\begin{figure}[h!]
  \centering
  \begin{subfigure}[t]{0.42\linewidth}
    \centering
    \includegraphics[width=0.9\linewidth]{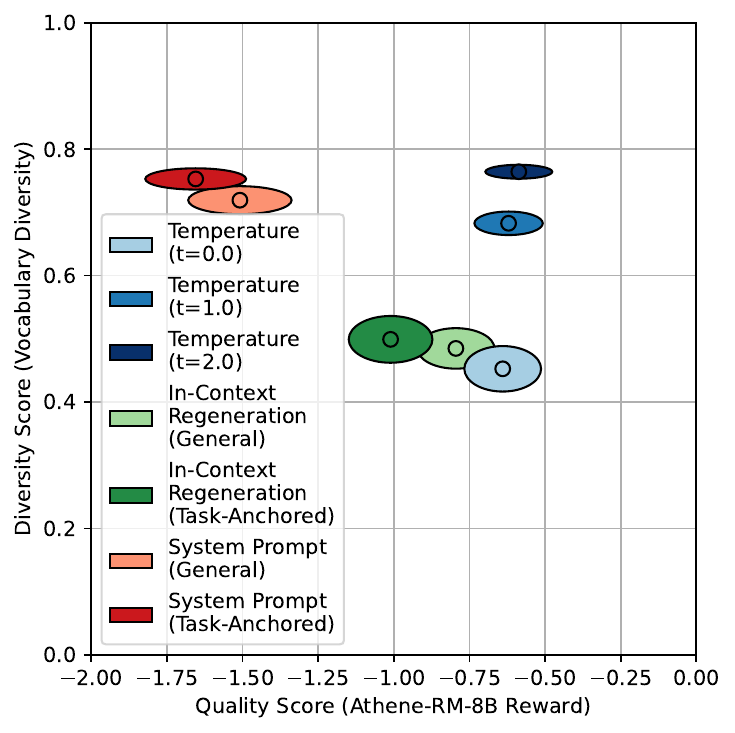}
    \caption{General Metrics}
    \label{fig:dq_general_mistral}
  \end{subfigure}
  \begin{subfigure}[t]{0.42\linewidth}
    \centering
    \includegraphics[width=0.9\linewidth]{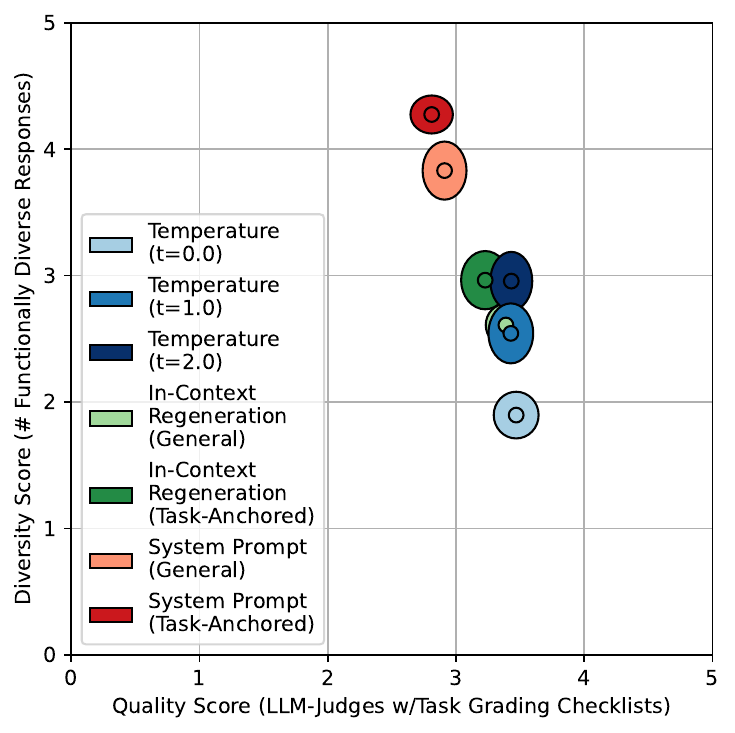}
    \caption{Task-Based Metrics}
    \label{fig:dq_ta_mistral}
  \end{subfigure}
  \caption{\centering Diversity-quality tradeoff under general vs task-based metrics for \textbf{Mistral-7B-Instruct-v0.3}.}
  \label{fig:dq_mistral}
\end{figure}

\begin{figure}[h!]
  \centering

  \begin{subfigure}[t]{0.32\linewidth}
    \centering
    \includegraphics[width=\linewidth]{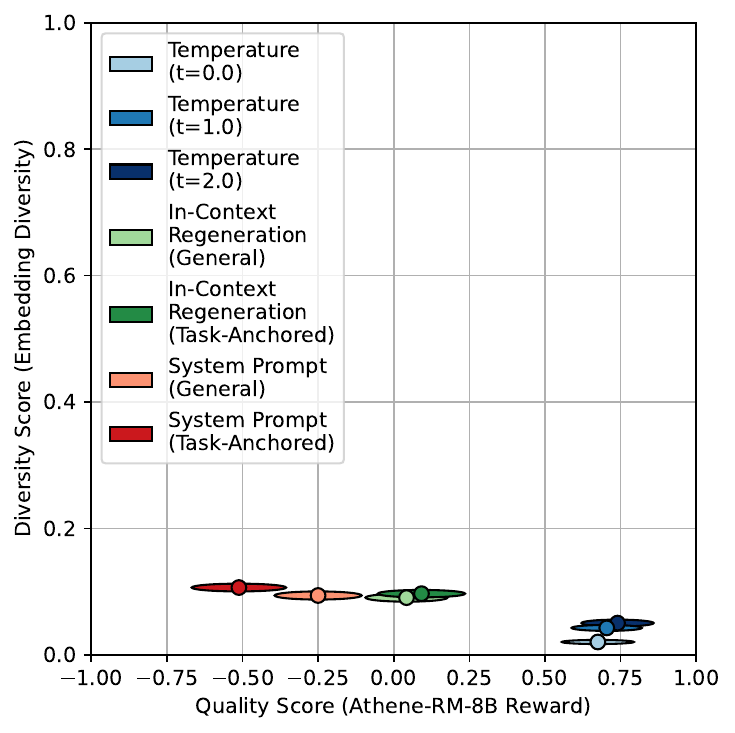}
    \caption{GPT-4o}
  \end{subfigure}
  \begin{subfigure}[t]{0.32\linewidth}
    \centering
    \includegraphics[width=\linewidth]{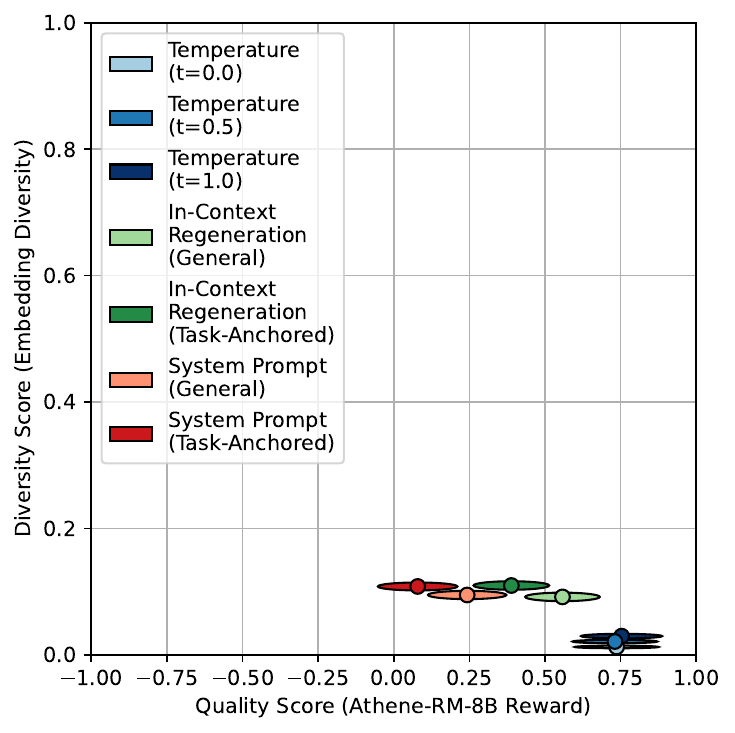}
    \caption{Claude-4-Sonnet}
  \end{subfigure}
  \begin{subfigure}[t]{0.32\linewidth}
    \centering
    \includegraphics[width=\linewidth]{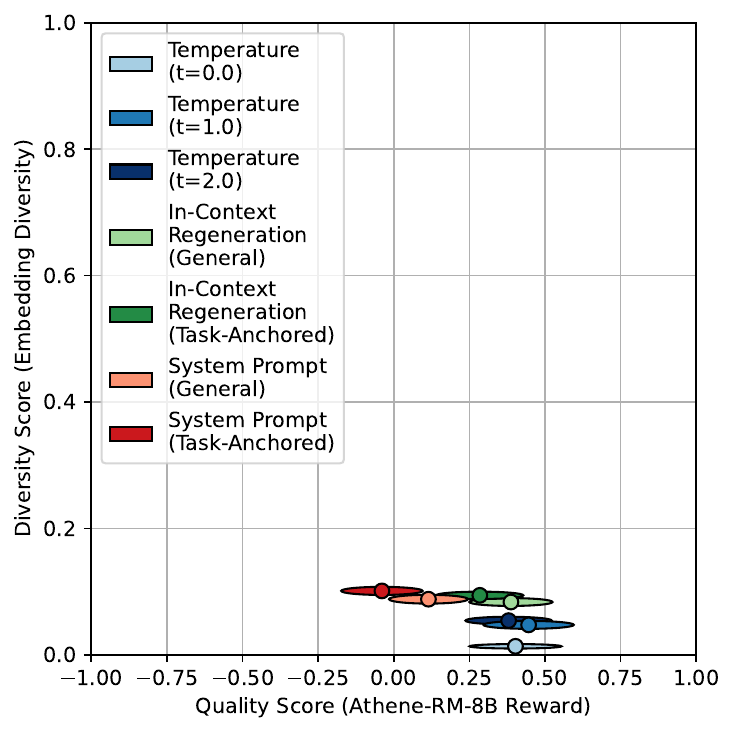}
    \caption{Gemini-2.5-Flash}
  \end{subfigure} 
  \begin{subfigure}[t]{0.32\linewidth}
    \centering
    \includegraphics[width=\linewidth]{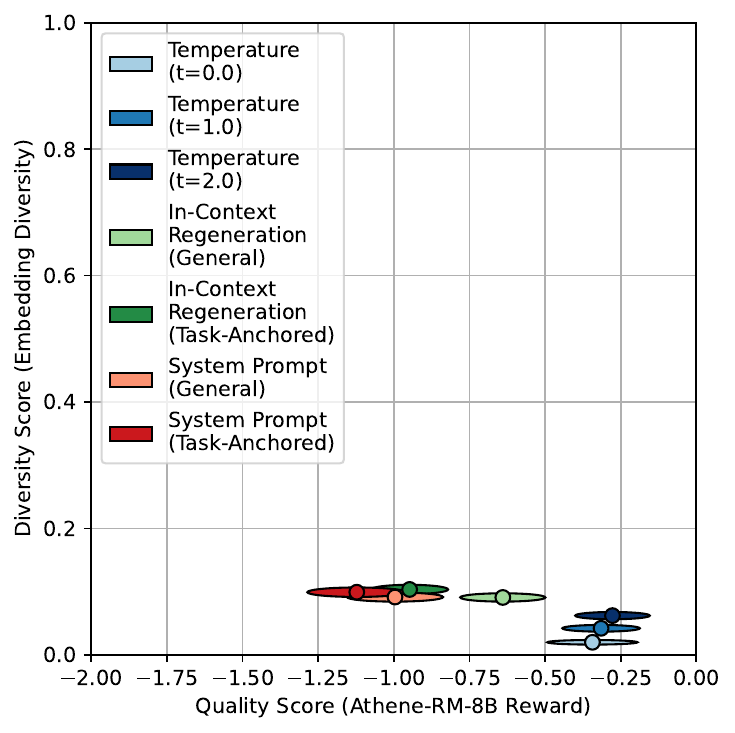}
    \caption{Llama-3.1-8B-Instruct}
  \end{subfigure}
  \begin{subfigure}[t]{0.32\linewidth}
    \centering
    \includegraphics[width=\linewidth]{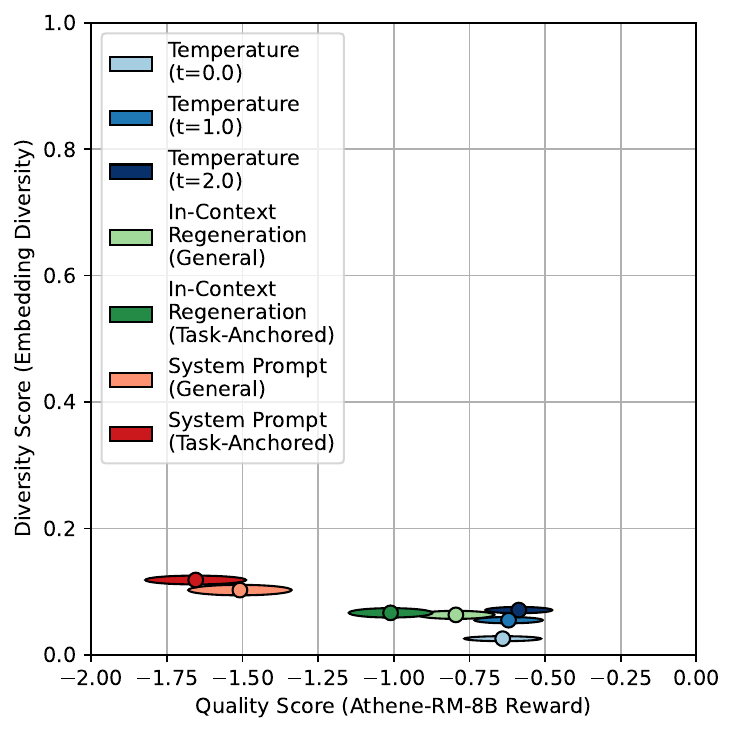}
    \caption{Mistral-7B-Instruct-v0.3}
  \end{subfigure}

  \caption{\centering Diversity-quality tradeoff using embedding diversity.}
  \label{fig:dq_embedding}
\end{figure}

\begin{figure}[h!]
  \centering
  \begin{subfigure}[t]{0.32\linewidth}
    \centering
    \includegraphics[width=\linewidth]{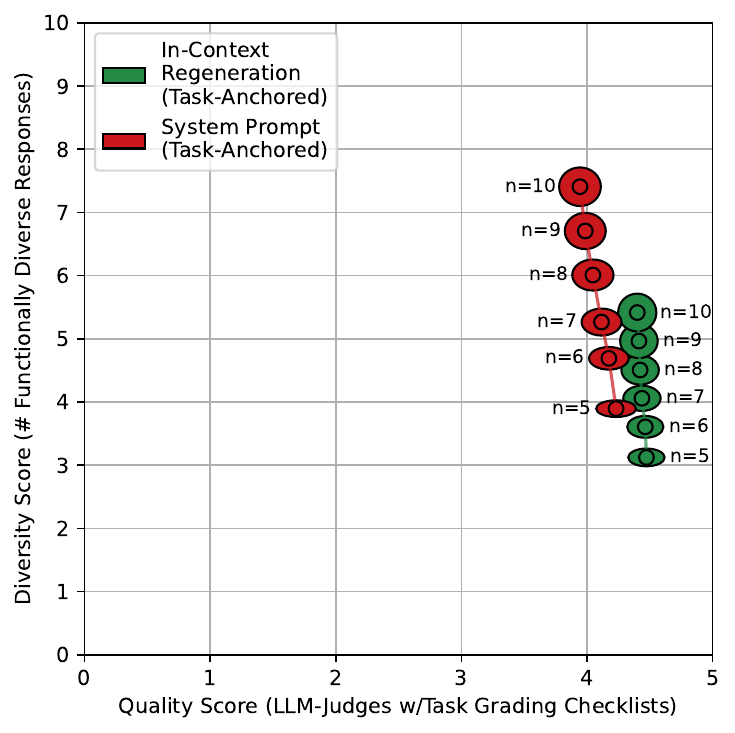}
    \caption{GPT-4o}
  \end{subfigure}
  \begin{subfigure}[t]{0.32\linewidth}
    \centering
    \includegraphics[width=\linewidth]{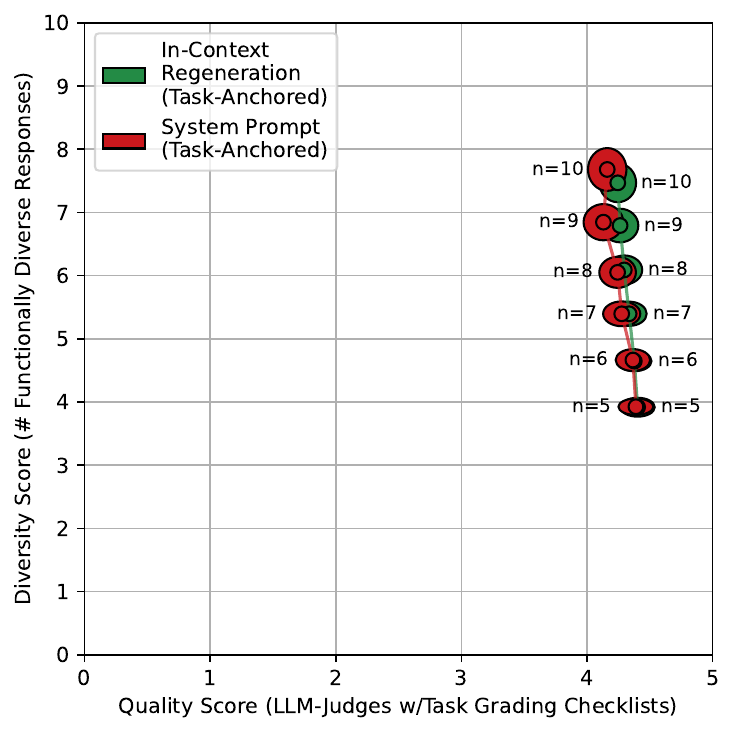}
    \caption{Claude-4-Sonnet}
  \end{subfigure}
  \begin{subfigure}[t]{0.32\linewidth}
    \centering
    \includegraphics[width=\linewidth]{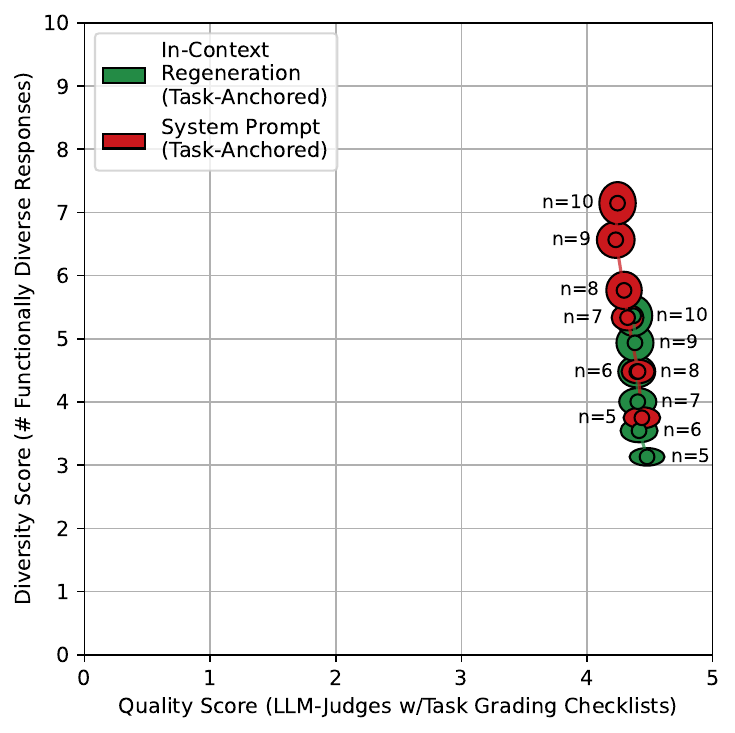}
    \caption{Gemini-2.5-Flash}
  \end{subfigure}
  \caption{Diversity-quality tradeoff for varying number of generated responses ($n=5$ to $n=10$). Judge metrics based on GPT-4o only. The number of functionally diverse responses consistently increases with more generated responses. However, there appear to be small (statistically insignificant) decreases in checklist-based quality. The quality decrease is larger for system prompt sampling, possibly due to $n=10$ approaching the max output length for a single generation.}
  \label{fig:dq_by_n_samples}
\end{figure}

\begin{itemize}
\item \ref{apx:human_exp_results}: Human Validation Experiment Results
\begin{itemize}
    \item Table~\ref{tab::IRR-full}: Inter-Rater Reliability (Human Annotators)
    \item Table~\ref{tab:annotator_groups_corr}: Correlation Between Human Annotator Groups
    \item Table~\ref{tab:annotation_res_appx}: Correlation Between Metrics \& Annotation (Task-Dependent Approach) (c.f. Table~\ref{tab:annotation_res_main} for Annotation of General Functional Diversity)
    \item Table~\ref{tab:annotation_res_using_avg}: Annotation Results Using Average of Human Labels (Instead of Majority Vote)
\end{itemize}
\item \ref{apx:diversity_results}: Task-Dependent Diversity Evaluation (by Model, Sampling Strategy \& Task Category)
\begin{itemize}
    \item Figures~\ref{fig:fun_div_claude}, \ref{fig:fun_div_gemini}, \ref{fig:fun_div_llama}, \ref{fig:fun_div_mistral}: Functional Diversity for Claude-4-Sonnet, Gemini-2.5-Flash, Llama-3.1-8B-Instruct, and Mistral-7B-Instruct-v0.3 (c.f. Figure~\ref{fig:fun_div_gpt} for GPT-4o)
    \item Table~\ref{tab:fun_div}: Functional Diversity Results
    \item Table~\ref{tab:vocab_div}: Vocabulary Diversity Results
    \item Table~\ref{tab:embed_div}: Embedding Diversity Results
    \item Table~\ref{tab:compression_div}: Compression Diversity Results
\end{itemize}
\item \ref{apx:dq_tradeoff_results}: Diversity-Quality Tradeoff Evaluation (by Model, Sampling Strategy \& Task Category)
\begin{itemize}
    \item Figure~\ref{fig:dq_claude}, \ref{fig:dq_gemini}, \ref{fig:dq_llama}, \ref{fig:dq_mistral}: Diversity-Quality Tradeoff for Claude-4-Sonnet, Gemini-2.5-Flash, Llama-3.1-8B-Instruct, and Mistral-7B-Instruct-v0.3 (c.f. Figure~\ref{fig:dq_tradeoff} for GPT-4o)
    \item Figure~\ref{fig:dq_embedding}: Diversity-Quality Tradeoff Using Embeddings
    \item Figure~\ref{fig:dq_by_n_samples}: Diversity-Quality Tradeoff for Varying Number of Generated Responses (5-10)
    \item Table~\ref{tab:checklist_quality}: Checklist-Based Quality Results
    \item Table~\ref{tab:reward_quality}: Athene-RM-8B Reward Results
    \item Table~\ref{tab:accuracy}: Accuracy Results (for verifiable tasks)
\end{itemize}
\item \ref{apx:alignment_exp_results}: Alignment Experiment Results
\begin{itemize}
    \item Figures~\ref{fig:fun_div_alignment}-\ref{fig:fun_div_alignment_steps}: Functional Diversity with DPO / GRPO
    \item Figure~\ref{fig:dq_alignment}: Diversity-Quality Tradeoff with DPO / GRPO
    \item Figures~\ref{fig:fun_div_darling}-\ref{fig:fun_div_darling_steps}: Functional Diversity with DARLING 
    \item Figure~\ref{fig:dq_darling}: Diversity-Quality Tradeoff with DARLING 
    \item Table~\ref{tab:fun_div_alignment}: Functional Diversity  with DPO and GRPO
    \item Table~\ref{tab:fun_div_darling}: Functional Diversity with DARLING alignment
\end{itemize}
\item \ref{apx:infinity-chat-res}: Infinity Chat Experiment Results
\begin{itemize}
\item Figures~\ref{fig:infinity_chat_gpt4o_by_category}, \ref{fig:infinity_chat_gemini_by_category}, \ref{fig:infinity_chat_claude_by_category}, 
\ref{fig:infinity_chat_by_model_commercial}: Functional Diversity for GPT-4o, Gemini-2.5-Flash, Claude-4-Sonnet, and all models together
\item Figures~\ref{fig:infinity_chat_cross_model_by_category}-\ref{fig:infinity_chat_cross_model_per_model}: Inter-Model Diversity
\end{itemize}
\item \ref{apx:prompt_illustrations}: Prompt-Level Illustrations of Functional Diversity (Figures \ref{fig:illus_cats}-\ref{fig:illus_transit})
\end{itemize}

\clearpage
\section{Additional Experiment Details}\label{apx:exp_details}

\subsection{Taxonomy Crosswalks}\label{apx:taxonomy}

Our taxonomy is grounded in existing literature on LLM output homogenization and diversity. Specifically, we observe that many studies evaluate homogenization in specific task domains, suggesting that problematic notions of homogenization are task dependent. We designed our taxonomy to cover a variety of these task categories. Table~\ref{tab:taxonomy_related_works} maps our taxonomy to related works that have evaluated output homogenization for each task category. To our knowledge, discussions of response variance in well-specified objective prompts (category A) are often found in studies of confabulation, not in the homogenization literature.

We further show that our task taxonomy captures many real-world LLM use cases identified in recent work. \citet{chatterji2025people} create a task taxonomy based on ChatGPT usage, which we map to our taxonomy in Table~\ref{tab:taxonomy_gpt}. Similarly, \citet{tamkin2024clio} provide a list of common tasks based on Claude usage, which we map to our taxonomy in Table~\ref{tab:taxonomy_claude}. All real-world task categories in these previous works map to at least one category in our taxonomy (for text-based tasks). Many real-world task categories appear to correspond with multiple task categories in our taxonomy. For example, the ``Practical Guidance'' category in \citet{chatterji2025people} may correspond to Problem-Solving Subjective (Category E) or Advice/Opinions (Category H). This illustrates how our task categories are meant to capture different functional diversity concepts, while the categories in these other works are meant to summarize usage trends. For ``Practical Guidance'' tasks, we would consider them as Problem-Solving Subjective if responses span different partially verifiable solutions, and as Advice/Opinions if responses span different non-verifiable perspectives or views. 

Ultimately, our task categories represent one categorization of functional diversity concepts, and a task may fall outside our taxonomy or correspond to multiple categories. In these cases, the model may resume its default behavior instead of using our task-dependent sampling technique, or promote diversity based on one of the applicable categories. Our approach is further generalizable to alternative taxonomies or task categories.

\begin{table}[htp] 
\centering
\caption{\centering Crosswalk of Our Taxonomy and Previous Output Homogenization Studies}
\small
\begin{tabular}{lp{6cm}}
\toprule
\textbf{Task Category} & \textbf{Previous Works} (Non-Exhaustive) \\
\toprule
A. Well-Specified Singular Objective & \cite{wei2024simpleQA} \\
\addlinespace
B. Underspecified Singular Objective & \cite{zhang2025noveltybench}\\
\addlinespace
C. Random Generation & \cite{hopkins2023can, zhang2025noveltybench} \\
\addlinespace
D. Problem-Solving Objective & \cite{lee2024enhancing, slocumdiverse, wu2025generative} \\ 
\addlinespace
E. Problem-Solving or Design Subjective & \cite{ma2024exploring, yang2025prompts} \\
\addlinespace
F. Encyclopedia Inquiry & \cite{sharma2024generative, sui2025critical, wright2025epistemic} \\
\addlinespace
G. Creative Writing & \cite{anderson2024homogenization, doshi2024generative, jiang2025artificial, lanchantin2025diverse, moon2024homogenizing, moon2025impersonal, padmakumar2024does, wu2025generative, zhang2025noveltybench}\\
\addlinespace
H. Advice or Opinions & \cite{agarwal2025ai, durmus2023towards, jain2026interaction, jiang2025artificial, santurkar2023whose, shahid2025llms, zhang2025cultivating}\\
\bottomrule
\end{tabular}
\label{tab:taxonomy_related_works}
\end{table}

\begin{table}[htp] 
\centering
\caption{\centering Crosswalk of Task Categories for ChatGPT Usage with Our Taxonomy}
\small
\begin{tabular}{p{6.75cm}p{4.25cm}}
\toprule
\textbf{ChatGPT Task Category} & \textbf{Categories in Our Taxonomy} \\
c.f. Table 3 in~\cite{chatterji2025people} \\
\toprule
Writing (Edit or Critique Provided Text, Personal Writing or Communication, Translation, Argument or Summary Generation, Write Fiction) & Underspecified Objective (B),\newline Creative Writing (G),\newline Advice or Opinions (H) \\
\midrule
Practical Guidance (How-To Advice, Tutoring or Teaching, Creative Ideation, Health, Fitness, Beauty, or Self-Care) & Problem-Solving Subjective (E),\newline Advice or Opinions (H) \\
\midrule
Technical Help (Mathematical Calculation, Data Analysis, Computer Programming) & Problem-Solving Objective (D),\newline Problem-Solving Subjective (E) \\
\midrule
Multimedia (Create an Image, Analyze an Image, Generate or Retrieve Other Media) & N/A (Our taxonomy is limited to text-based tasks) \\
\midrule
Seeking Information (Specific Info, Purchasable Products, Cooking \& Recipes) & Well-Specified Objective (A),\newline Underspecified Objective (B),\newline Encyclopedia Inquiry (F),\newline Advice or Opinions (H) \\
\midrule
Self-Expression (Greetings \& Chitchat, Relationships \& Personal Reflection, Games \& Role Play) & Creative Writing (G),\newline Advice or Opinions (H) \\
\bottomrule
\end{tabular}
\label{tab:taxonomy_gpt}
\end{table}

\begin{table}[htp]
\centering
\caption{\centering Crosswalk of Top 10 Task Categories in Claude Usage with Our Taxonomy}
\small
\begin{tabular}{p{6cm}p{4.5cm}}
\toprule
\textbf{Claude Task Category} & \textbf{Categories in Our Taxonomy} \\
c.f. Figure 6 in~\cite{tamkin2024clio} \\
\toprule
Web and mobile application\newline development assistance & Problem-Solving Objective (D),\newline Problem-Solving Subjective (E) \\
\midrule
Content creation and communication\newline assistance across disciplines & Creative Writing (G) \\
\midrule
Multidisciplinary academic research\newline and writing assistance & Well-Specified Objective (A),\newline Underspecified Objective (B),\newline Encyclopedia Inquiry (E) \\
\midrule
Education and career development\newline assistance & Advice or Opinions (H) \\
\midrule
Implement and optimize diverse\newline  AI/ML technologies and applications & Problem-Solving Objective (D),\newline Problem-Solving Subjective (E) \\
\midrule
Business strategy and operations\newline assistance across industries & Problem-Solving Subjective (E),\newline Advice or Opinions (H) \\
\midrule
Multilingual NLP, translation, and\newline  linguistic analysis services & Underspecified Objective (B),\newline Creative Writing (G) \\
\midrule
DevOps and cloud infrastructure\newline implementation and troubleshooting & Problem-Solving Objective (D),\newline Problem-Solving Subjective (E) \\
\midrule
Digital marketing and SEO\newline optimization assistance & Problem-Solving Subjective (E),\newline Advice or Opinions (H) \\
\midrule
Data analysis, visualization, and\newline management assistance & Problem-Solving Subjective (E),\newline Advice or Opinions (H) \\
\bottomrule
\end{tabular}
\label{tab:taxonomy_claude}
\end{table}

\clearpage

\subsection{Evaluation Datasets}\label{apx:datasets}

We sample 344 usable prompts (350 sampled, 6 excluded; see §\ref{apx:taxonomy_judge}) from the following datasets for our evaluation of output homogenization. These datasets were chosen to achieve coverage across our task taxonomy (c.f. Table~\ref{tab:eval_data}). For random sampling, we first shuffle the dataset using a random seed of 38, then select the required number of prompts in order from the shuffled dataset. 

\begin{itemize}
    \item \textbf{Community Alignment}~\citep{zhang2025cultivating}: A diverse human preference dataset containing user-generated prompts. We use 50 randomly-sampled prompts from the subset of user-generated first-turn prompts in English. Users were instructed to ``ask, request, or talk to the model about something important to you or that represents your values. This could be related to work, religion, family, relationships, politics, or culture.''
    \item \textbf{MacGyver}~\citep{tian2024macgyver}: A dataset of creative problem-solving tasks. We use 50 randomly-sampled prompts from the subset of ``solvable'' problems that require ``unconventional'' solutions.
    \item \textbf{MATH-500}~\citep{lightman2023let}: A subset of the MATH  dataset~\citep{hendrycks2021measuring}. We use 10 randomly-sampled prompts from each of the 5 difficulty levels.
    \item \textbf{NoveltyBench}~\citep{zhang2025noveltybench}: A dataset of creative tasks where multiple distinct and high-quality outputs are expected. We use their entire curated dataset of 100 prompts.
    \item \textbf{SimpleQA}~\citep{wei2024simpleQA}: A dataset of short, fact-seeking queries across diverse topics. The prompts were created to be challenging for frontier models (e.g. GPT-4o accuracy $<40\%$). We use 50 randomly-sampled prompts.
    \item \textbf{WildBench}~\citep{linwildbench}: A subset of the WildChat dataset~\citep{zhaowildchat}. WildChat is a corpus of 1 million user-ChatGPT conversations. WildBench is a filtered subset of WildChat such that tasks are diverse and challenging for models. We use 50 randomly-sampled prompts from the WildBench-V2.
\end{itemize}

We separately evaluate 100 prompts from the \textsc{Infinity-Chat} benchmark~\citep{jiang2025artificial} since it was released concurrently with our work. These prompts are also sourced from WildChat and mostly correspond to open-ended tasks. See Appendix~\ref{apx:infinity-chat-details} for more details.

\subsection{Task Classification Into Our Taxonomy}
\label{apx:taxonomy_judge}

When evaluating functional diversity, we use ground-truth task categories for each prompt based on the source dataset and human annotation. When generating responses in our task-dependent sampling technique, we use the model's task categorization of the prompt.

Table~\ref{tab:eval_data} shows the number of prompts by ground-truth category and dataset. For SimpleQA and MATH-500, we classify prompts as category A (well-specified objective) and category D (problem-solving objective), respectively. For Community Alignment, NoveltyBench, and WildBench, two authors independently classified these prompts into categories. 11 prompts received disagreeing labels, which the annotators resolved after discussion. 6 prompts did not fit into our taxonomy (all from WildBench) due to missing information (e.g. prompts that referenced an unattached document) or language (we only evaluate English prompts), leaving 344 prompts for evaluation..

We determine models' task categorization of each prompt using the judge prompt below. The agreement rate with ground-truth categories is 82\% for GPT-4o, 86\% for Claude-4-Sonnet, 84\% for Gemini-2.5-Flash, 56\% for Llama-3.1-8B-Instruct, and 46\% for Mistral-7B-Instruct-v0.3.

\clearpage
\begin{table}[htbp]
\centering
\small
\caption{\centering Number of prompts per dataset and taxonomy category.}
\label{tab:eval_data}
\resizebox{\textwidth}{!}{
\begin{tabular}{lccccccc}
\toprule
 & Community & \multirow{2}{*}{Math 500} & \multirow{2}{*}{MacGyver} & \multirow{2}{*}{Novelty Bench} & \multirow{2}{*}{Simple QA} & \multirow{2}{*}{Wild Bench} & \multirow{2}{*}{Total} \\
 & Alignment \\
\toprule
Well-Specified Objective & 2 & 0 & 0 & 1 & 50 & 0 &  53 \\
\midrule
Underspecified Objective & 0 & 0 & 0 & 16 & 0 & 0 & 16 \\
\midrule
Random Generation & 0 & 0 & 0 & 14 & 0 & 0 & 14  \\
\midrule
Problem-Solving Objective & 0 & 50 & 0 & 0 & 0 & 5 & 55 \\
\midrule
Problem-Solving Subjective & 0 & 0 & 50 & 0 & 0 & 0 & 50 \\
\midrule
Encyclopedia Inquiry & 9 & 0 & 0 & 2 & 0 & 12 & 23\\
\midrule
Creative Writing & 1 & 0 & 0 & 23 & 0 & 21 & 45\\
\midrule
Advice or Opinions & 38 & 0 & 0 & 44 & 0 & 6 & 88 \\
\toprule
Total & 50 & 50 & 50 & 100 & 50 & 44 & 344\\
\toprule
\end{tabular}}
\end{table}

\begin{promptbox}
Read the prompt below and decide which task category it belongs to. For prompts that have objective responses, choose from categories A, B, C, or D. For prompts that have subjective responses, choose from categories E, F, G, or H.\\

Prompt: \{prompt\}\\

Task Categories:\\
A - Well-Specified Singular Objective: Task to generate a single verifiable correct answer.\\
B - Underspecified Singular Objective: Task to generate a single answer for a prompt that has multiple verifiable correct answers.\\
C - Random Generation Objective: Task to generate a response that involves randomizing over a set of finite options.\\
D - Problem Solving Objective: Task to generate an answer with reasoning or explanations for a problem with a single verifiable correct answer.\\
E - Problem Solving Subjective: Task to generate an answer with reasoning or explanations for a problem with many verifiably correct answers. \\
F - Encyclopedia Inquiry Subjective: Task to generate information about real-world societies, traditions, events, or social domains, where there are credible references.\\
G - Creative Generation Subjective: Task to generate a response that involves creative expression where there are potentially infinite subjective responses.\\
H - Advice or Opinion Subjective: Task to generate a response that gives advice, opinions, or feedback on specific topics or scenarios.\\

For the prompt above, only output the assigned task category (A, B, C, D, E, F, G, or H) without any additional text.
\end{promptbox}

\clearpage
\subsection{Sampling Strategies}\label{apx:sampling}

We compare temperature sampling, in-context regeneration, and system prompt sampling. For each strategy, we sample 5 responses per prompt.

For temperature sampling, we consider three levels for each model based on its permitted range: low ($t=0.0$ for GPT, Claude \& Gemini, $t=0.1$ for Llama \& Mistral), medium ($t=1.0$ for GPT \&  Gemini, $t=0.5$ for Claude, Llama \& Mistral), and high ($t=2.0$ for GPT \&  Gemini, $t=1.0$ for Claude, Llama \&  Mistral). Note that $t=0$ is non-deterministic for commercial models; hence we compare to $t=0.1$ for open-weight models. For both in-context regeneration and system prompt sampling, we use medium temperature values. For all sampling strategies, we set the nucleus sampling parameter to 0.9 and the maximum number of output tokens to 1024.

For in-context regeneration, the first response is generated with the original prompt. Subsequent responses are generated using the regeneration prompts below. For system prompt sampling, multiple responses are extracted from a single generation. The model is instructed to produce multiple responses separated by a delimiter, allowing them to be de-aggregated with regular expressions. We compare general and task-dependent approaches to in-context regeneration and system-prompt sampling. For task-dependent approaches, the task category for the prompt is determined by the model's self-categorization of the task (c.f.~\ref{apx:taxonomy_judge}). Parallel inference techniques could help scale these approaches~\citep{dong2025generalized}. 

\subsubsection*{General System Prompt~\citep{zhang2025cultivating}}

This prompt is similar to the one used in \citet{zhang2025cultivating}, except they use: ``Generate \{num\_responses\} that represent diverse values''. We generalize this instruction to apply to all the categories in our taxonomy, since \citet{zhang2025cultivating} focus on subjective tasks.

\begin{promptbox}
\footnotesize
Generate \{num\_responses\} different responses to the following prompt. \\

Each response should start with "\# Response X:" where X is a number from 1 to \{num\_responses\} to demarcate where each different response begins. Make sure to stop at \# Response \{num\_responses\}. Make each generated response self-contained. They should not rely on the other responses for context. 
\end{promptbox}

\subsubsection*{Task-Dependent System Prompt}

\begin{promptbox}
\footnotesize
\{Task-Dependent System Prompt Instruction from Table~\ref{tab:prompt_based_strategies_all}\} \\

Each response should start with "\# Response X:" where X is a number from 1 to \{num\_responses\} to demarcate where each different response begins. Make sure to stop at \# Response \{num\_responses\}. Make each generated response self-contained. They should not rely on the other responses for context. 
\end{promptbox}

\subsubsection*{General In-Context Regeneration~\citep{zhang2025noveltybench}}

\begin{promptbox}
\footnotesize
Can you generate a different response?
\end{promptbox}

\subsubsection*{Task-Dependent In-Context Regeneration}

\begin{promptbox}
\footnotesize

\{Task-Dependent In-Context Regeneration Instruction from Table~\ref{tab:prompt_based_strategies_all}\} \\

Do not include any starting phrases or reasons for why your new response is different. Your response should be self-contained, as if the prompt was the first thing that I asked. \\

Remember, the prompt is: \{prompt\} 
\end{promptbox}

\begin{table}[h!] 
\centering
\caption{\centering Prompts for Task-Dependent Sampling Strategies}
\footnotesize
\resizebox{\textwidth}{!}{
\begin{tabular}{lp{6cm}p{6cm}}
\toprule
\textbf{Category} & \textbf{Task-Dependent System Prompt} & \textbf{Task-Dependent In-Context Regeneration Prompt} \\
\toprule
\makecell[l]{\\Well-Specified\\Objective} & The following prompt has a single correct answer. Generate \{num\_responses\} responses. If relevant, slight variation in wording is allowed but the answer should remain the same. & Can you generate a different response? The prompt has a single correct answer, so your answer should remain the same. If relevant, slight variation in wording is allowed. \\
\addlinespace
\makecell[l]{\\Underspecified\\Objective} & The following prompt is underspecified and has many correct answers. Generate \{num\_responses\} responses, each with a different correct answer. & Can you generate a different correct answer? The prompt is underspecified and has many correct answers. \\
\addlinespace
\makecell[l]{\\Random\\Generation} & The following prompt is asking you to randomize over a set of finite options. Generate \{num\_responses\} responses, each with a different pseudo-random option. & Can you generate a different pseudo-random response? The prompt is asking you to randomize over a set of finite options. \\
\addlinespace
\makecell[l]{\\Problem-Solving\\Objective} & The following problem has a single correct answer, but can be solved using different problem-solving strategies. Generate \{num\_responses\} different solutions, each with a different problem-solving strategy. & Can you solve the problem using a different strategy? The problem has a single correct answer, but can be solved using different problem-solving strategies.\\
\addlinespace
\makecell[l]{\\Problem-Solving\\Subjective} & The following problem has multiple correct answers, and may be solved using different problem-solving strategies. Generate \{num\_responses\} different solutions, each with a different answer or problem-solving strategy. & Can you solve the problem using a different strategy? The problem has multiple correct answers, and may be solved using different problem-solving strategies. \\
\addlinespace
\makecell[l]{\\Encyclopedia\\Inquiry} & The following prompt is asking for information about the real-world, where there may be different factual perspectives. Your response must be grounded in credible references though references do not need to be mentioned. Generate \{num\_responses\} responses that reflect different perspectives. & Can you generate a new response that reflects a different factual perspective? The prompt is asking for information about the real-world, where there may be different factual perspectives. Your response must be grounded in credible references though references do not need to be mentioned. \\
\addlinespace
\makecell[l]{\\Creative\\Writing} & The following prompt is asking for creative expression, so there are many possible subjective responses. Generate \{num\_responses\} unique responses by varying the key creative elements such as tone, genre, point of view, theme, structure, etc. Each response should have different creative elements and reflect a distinct creative expression. & Can you generate a new response with different creative elements? The prompt is asking for creative expression, so there are many possible subjective responses. Your new response should change the key creative elements such as tone, genre, point of view, theme, structure, etc. \\
\addlinespace
\makecell[l]{\\Advice or\\Opinions} & The following prompt is asking for advice or opinions, so there are many possible subjective responses. Generate \{num\_responses\} unique responses where each response expresses a different viewpoint or perspective. & Can you generate a new response with a different viewpoint or perspective? The prompt is asking for advice or opinions, so there are many possible subjective responses. \\
\bottomrule
\end{tabular}
}
\label{tab:prompt_based_strategies_all}
\end{table}

\clearpage
\subsection{Diversity Metrics}\label{apx:div_metrics}

\subsubsection*{Task-Dependent Functional Diversity}

Let $\mathcal{P}$ denote the set of possible input prompts or tasks and let $\mathcal{Y}$ denote the set of possible outputs (e.g. sequences of tokens). We adopt the simple notation that a \textit{language model} $\mathcal{M}$ is a stochastic function $\mathcal{M}: \mathcal{P} \to \mathcal{Y}$ that maps each prompt $p \in \mathcal{P}$ to an output $y \in \mathcal{Y}$. For a given prompt $p$, we assume $d(p, y_a, y_b) \in [0,1]$ to be a pairwise diversity metric that indicates whether two responses $y_a$ and $y_b$ differ. To specify \textit{task-dependent functional diversity}, we anchor the definition of $d(p, y_a, y_b)$ on the task category $c(p) \in \mathcal{T}$, where each prompt $p$ is is associated with a task category based on Table~\ref{tab:task_categories}.

\begin{definition}[Task-Dependent Functional Diversity]
\label{def:fun_div}
Given a prompt $p \in \mathcal{P}$ with associated task category $c(p) \in \mathcal{T}$, and two responses $y_a, y_b \in \mathcal{Y}$, the \emph{task-dependent functional diversity} is 
\[
d(p, y_a, y_b) := \mathbbm{1}_{c(p)}[y_a \neq y_b],
\]
where $\mathbbm{1}_{c(p)}[y_a \neq y_b]$ is an indicator function that returns $1$ if $y_a$ and $y_b$ are functionally different with respect to the task category $c(p)$, and $0$ otherwise.
\end{definition}

For example, consider whether two responses are functionally diverse in the \textit{Problem-Solving Objective} task (category D). Here, $d(p, y_a, y_b)$ represents whether responses have different solution strategies, and assumes the verifiable answer to be the same. In practice, we use taxonomy-aided LLM-judges to evaluate task-dependent functional diversity, based on the following prompt template. See \ref{apx:human_annotation} for LLM-judge validation.

\begin{promptbox}
\footnotesize
For the given prompt and two responses, determine if the responses are functionally equivalent. Functional equivalence means a user who has seen one response would find the other response to be redundant.\\

\{Task-Dependent Functional Diversity Definition in Table~\ref{tab:div_prompts}\}\\

\#\#\#\\
Prompt: \{prompt\}\\
Response 1: \{response 1\}\\
Response 2: \{response 2\}\\
\#\#\#\\

Are the responses functionally equivalent?\\

\{Task-Dependent Diversity Judge Options in Table~\ref{tab:div_prompts}\}\\

Only output YES or NO.
\end{promptbox}
\vspace{1em}
Based on pairwise comparisons of functional diversity, a set of responses can be partitioned into functionally distinct response groups.

\begin{definition}[Number of Functionally Unique Responses]
\label{def:fun_div_clusters}  Let $\mathcal{Y}_p = \{y_1,\dots,y_n\}$ be a set of responses for prompt $p$. Define an undirected graph $G = (\mathcal{Y}_p, E)$ where the edge set is $E := \{(y_a,y_b) \in \mathcal{Y}_p \times \mathcal{Y}_p : d(p,y_a,y_b) = 0\}$. Then, the \textit{number of functionally unique responses} is $|\pi_0(G)|$ where $\pi_0(G)$ denotes the set of path-connected components of $G$.
\end{definition}

\clearpage

\begin{table}[htp] 
\centering
\caption{\centering Prompts for Taxonomy-Aided LLM-Judges of Functional Diversity} 
\footnotesize
\resizebox{\textwidth}{!}{
\begin{tabular}{lp{7cm}p{5cm}}
\toprule
\textbf{Category} & \textbf{Task-Dependent Functional Diversity Def.} & \textbf{Diversity Judge Options} \\
\toprule
\makecell[l]{\\Well-Specified\\Objective} & The prompt has a single correct answer. Responses are functionally equivalent if they represent the same answer. & Output YES if the responses represent the same answer. Output NO if the responses represent different answers.\\
\addlinespace
\makecell[l]{\\Underspecified\\Objective} & The prompt is underspecified and has many correct answers. Responses are functionally equivalent if they represent the same answer. & Output YES if the responses represent the same answer. Output NO if the responses represent different answers. \\
\addlinespace
\makecell[l]{\\Random\\Generation} & The prompt is asking for a random response over a set of finite options. Responses are functionally equivalent if they represent the same pseudo-random option. & Output YES if the responses represent the same pseudo-random option. Output NO if the responses represent different pseudo-random options.  \\
\addlinespace
\makecell[l]{\\Problem-Solving\\Objective} & The prompt involves solving a problem with a single correct answer, but it can be solved using different problem-solving strategies. Responses are functionally equivalent if they represent the same problem-solving strategy. & Output YES if the responses represent the same problem-solving strategy. Output NO if the responses represent different problem-solving strategies. \\
\addlinespace
\makecell[l]{\\Problem-Solving\\Subjective} & The prompt involves solving a problem with multiple correct answers, and may be solved using different problem-solving strategies. Responses are functionally equivalent if they represent the same answer and problem-solving strategy. & Output YES if the responses represent the same answer and problem-solving strategy. Output NO if the responses represent different answers or problem-solving strategies. \\
\addlinespace
\makecell[l]{\\Encyclopedia\\Inquiry} & The prompt is asking for information about the real-world, where there may be different factual perspectives. Responses are functionally equivalent if they represent similar factual perspectives. & Output YES if the responses represent similar perspectives. Output NO if the responses represent different perspectives. \\
\addlinespace
\makecell[l]{\\Creative\\Writing} & The prompt is asking for creative expression where there are many possible subjective responses. Responses are functionally equivalent if the key creative elements (such as tone, genre, point of view, theme, structure, etc.) are the same. & Output YES if the responses have similar key creative elements. Output NO if the responses have different key creative elements. \\
\addlinespace
\makecell[l]{\\Advice or\\Opinions} & The prompt is asking for advice or opinions. Responses are functionally equivalent if they express the same viewpoint or perspective, even if they are worded differently.
& Output YES if the responses have similar perspectives. Output NO if the responses have different perspectives. \\
\bottomrule
\end{tabular}
}
\label{tab:div_prompts}
\end{table}

\clearpage

\begin{table}[h!] 
\centering
\caption{\centering Examples of Functionally Diverse Responses by Category}
\label{tab:diversity_examples}
\footnotesize
\resizebox{\textwidth}{!}{
\begin{tabular}{p{2cm}p{5cm}p{5cm}p{5cm}}
\toprule
\textbf{Category} & \textbf{Prompt} & \textbf{Response 1} & \textbf{Response 2} \\
\toprule
Well-Specified\newline Objective & What is the name of the project that won the 2014 ACM Software System Award? & The project that won the **2014 ACM Software System Award** is **Apache Hadoop**. Apache Hadoop is an open-source framework for distributed storage and processing of large datasets, which has had a significant impact on big data processing and analytics. & The project that won the 2014 ACM Software System Award is **The Eclipse Project**. \\
\midrule
Underspecified\newline Objective & Name a Harry Potter Book. & Harry Potter and the Sorcerer's Stone & Harry Potter and the Half-Blood Prince \\
\midrule
Random\newline Generation & Pick a number between 1 and 100. Now tell me what you picked. & I picked the number 42. & I picked the number 87. \\
\midrule
Problem-Solving\newline Objective & If $2^8=4^x$, what is the value of $x$? & We know that \( 4^x = (2^2)^x \). Using the property of exponents \((a^m)^n = a^{m \cdot n}\), this becomes:
\[
4^x = 2^{2x}.
\]
The original equation \( 2^8 = 4^x \) now becomes:
\[
2^8 = 2^{2x}.
\]
Since the bases are the same, we equate the exponents:
\[
8 = 2x.
\]
Solve for \( x \) by dividing both sides by 2:
\[
x = 4.
\] & We start with the equation \( 2^8 = 4^x \). Take the natural logarithm (ln) of both sides:
\[
\ln(2^8) = \ln(4^x).
\]
Using the logarithmic rule \(\ln(a^b) = b \cdot \ln(a)\), this becomes:
\[
8 \cdot \ln(2) = x \cdot \ln(4).
\]
Express \( \ln(4) \) as \( \ln(2^2) \), which simplifies to \( 2 \cdot \ln(2) \). Substituting this back:
\[
8 \cdot \ln(2) = x \cdot (2 \cdot \ln(2)).
\]
Cancel \( \ln(2) \) from both sides:
\[
8 = 2x.
\]
Solve for \( x \):
\[
x = 4.
\] \\
\midrule
Problem-Solving\newline Subjective & You need to tighten a loose door knob but you don't have a screwdriver. The available items are a round-tipped hairpin, a brittle plastic fork, a flimsy nail file, a pair of rusted pliers, a thin rubber band, and a butter knife. The door knob screws are not the standard slot or Phillips head, but are special star head screws which cannot be tightened using a butter knife or hairpin due to the unusual shape and the risk of stripping the screw head. How can you tighten the door knob? & You can use the pair of rusted pliers as a workaround tool. Although they are rusted, they might still have enough grip to grab onto the edges of the star head screw. Position the pliers carefully around the head of the screw and twist it in the tightening direction. Apply steady pressure and try to avoid slipping. Clean the rust off the pliers beforehand, if possible, to improve their grip. & Since none of the tools are directly suitable for a star head screw, you can try securing the loose door knob in place temporarily. Use the thin rubber band to wrap tightly around the door knob base, compressing it against the door. This won't tighten the screws but can stabilize the knob until you acquire the correct star head screwdriver for a permanent fix. \\
\midrule
Encyclopedia\newline Inquiry & Did any major countries reinstate the death penalty after prohibiting it? & While it is not common, some nations have reinstated the death penalty after abolishing it. For example, Pakistan imposed a moratorium on executions in 2008, effectively halting the use of the death penalty. However, in 2014, the country lifted the moratorium following a devastating terrorist attack, and executions resumed. This demonstrates how security concerns can influence such decisions. & Historically, a few countries have reinstated the death penalty after initially banning or suspending it. The United States offers an example at the state level, where capital punishment was temporarily halted in the 1970s after the Supreme Court’s Furman v. Georgia decision but reinstated later following the Gregg v. Georgia ruling in 1976. Nationally, reinstatements are usually tied to shifting societal attitudes or judicial rulings. \\
\midrule
Creative\newline Writing & Write a short love poem with 4 lines. & Your laughter blooms like spring's embrace,  
A gentle warmth I cannot replace.  
In your eyes, a world so true,  
Forever, my heart beats for you. & The ocean hums your name to me,  
Its waves reflect eternity.  
My soul is bound, your love’s my art,  
Forever etched within my heart. \\
\midrule
Advice or\newline Opinions & I am naming my Indian son what one name should I name him? I get anxiety when given multiple options so please only give me one name. & Consider naming your son ``Kabir.'' This name is associated with wisdom and spirituality, inspired by the renowned Indian poet and saint. & You might like the name ``Rohan.'' It is simple, easy to pronounce globally, and represents growth and ascent. \\
\bottomrule
\end{tabular}}
\end{table}

\clearpage 

\subsubsection*{General Diversity Metrics}

Previous homogenization studies often use general diversity metrics. By general, we mean the metric applies a single diversity construct to all tasks. We evaluate the following general metrics. For embedding diversity, we generate response embeddings using the gemini-embedding-001 model (with 3072-dimensional embeddings). For compression diversity, we use gZip to compress the concatenated text of response pairs following~\citet{shaib2024standardizing}. For general functional diversity, we use the \textsc{Novelty-Bench} semantic classifier which is a fine-tuned \textit{deberta-v3-large} model~\citep{zhang2025noveltybench}.

\begin{definition}[Vocabulary Diversity]
\label{def:vocab_div}
Given two responses $y_a, y_b \in \mathcal{Y}$, let $\mathcal{V}_a$ and $\mathcal{V}_b$ denote the sets of unique words in $y_a$ and $y_b$, respectively. The \emph{vocabulary diversity} between $y_a$ and $y_b$ is
\[
d_{\mathrm{vocab}}(y_a, y_b) := 1 - \frac{|\mathcal{V}_a \cap \mathcal{V}_b|}{|\mathcal{V}_a \cup \mathcal{V}_b|},
\]
where $|\mathcal{V}_a \cap \mathcal{V}_b|$ is the number of shared words and $|\mathcal{V}_a \cup \mathcal{V}_b|$ is the total number of unique words in both responses.
\end{definition}

\begin{definition}[Embedding Diversity]
\label{def:embed_div}
Given two responses $y_a, y_b \in \mathcal{Y}$, let $e(y)$ denote the embedding vector for response $y$. The \emph{embedding diversity} between $y_a$ and $y_b$ is 
\[
d_{\mathrm{embed}}(y_a, y_b) := 1 - \cos(e(y_a), e(y_b)),
\]
where $\cos(e(y_a), e(y_b))$ is the cosine similarity between the embedding vectors of $y_a$ and $y_b$.
\end{definition}

\begin{definition}[Compression Diversity]
\label{def:compression_div}
Given two responses $y_a, y_b \in \mathcal{Y}$, the \emph{compression diversity} between $y_a$ and $y_b$ is 
\[
d_{\mathrm{compression}}(y_a, y_b) := \frac{\text{compressed size of }y_a || y_b}{\text{size of } y_a || y_b},
\]
where $y_a || y_b$ is the concatenated text of responses $y_a$ and $y_b$.
\end{definition}

\begin{definition}[General Functional Diversity -- \textsc{Novelty-Bench}]
\label{def:gen_fun_div} Given two responses $y_a$, $y_b$ $\in \mathcal{Y}$, general functional diversity as defined in \citet{zhang2025noveltybench} is:  

\[
d(p, y_a, y_b) := \mathbbm{1}_{NB}[y_a \neq y_b],
\]
where $\mathbbm{1}_{NB}[y_a \neq y_b]$ is an indicator function that returns $1$ if a user who has seen $y_a$ ``would likely benefit'' from seeing $y_b$, and $0$ otherwise.
\end{definition}

\subsection{Human Annotation of Functional Diversity}\label{apx:human_annotation}

As described in $\S$~\ref{sec:human_validation}, we conduct a small user study ($n=22$) to validate that our taxonomy aligns with human annotations of functional diversity. Below are supplementary details. 

\paragraph{\textbf{Prompts}} Each annotator labeled the same 105 response pairs (5 prompts per task category, 3 sampling strategies per prompt). All responses were from GPT-4o. Using the datasets in \ref{apx:datasets}, we randomly sampled response pairs such the length of each chosen response was less than the median response length for that category. This constraint helped reduce annotation time. While annotators also labeled 15 response pairs for Problem-Solving Objective (Category D), we exclude these annotations since several annotators reported that they did not have sufficient math proficiency to understand the responses. 

\paragraph{\textbf{Annotators}} Annotations were performed by professional data labeling analysts engaged through a third-party staffing provider. The name of the staffing provider is not disclosed for anonymous peer review. The staffing provider was required by law to compensate annotators in accordance with applicable wage requirements. Total annotation time across all annotations was around 44 hours.  

\paragraph{\textbf{Annotation Instructions}} Annotators were randomly assigned to label general functional diversity or task-dependent functional diversity. All annotators participated in a short training session where they were shown an example prompt, response pair, and functional diversity label for each task category (8 examples total). The examples were different from the annotation task. The label received unanimous consensus from two authors and all taxonomy-guided LLM judges. Annotators assigned to label task-dependent functional diversity were also shown the corresponding task-based definition of functional diversity for each example.

\subsection{Task-Dependent Quality Metric}\label{apx:quality_prompts}

We follow the approach in \citet{linwildbench} and \citet{wei2025rocketeval} to evaluate quality using LLM-judges with task-specific grading checklists. First, the LLM-judge generates a grading checklist for each prompt with key factors that should be present in high-quality responses (Checklist Creation Prompt). Table~\ref{tab:checklists} includes examples of the grading checklists generated by LLM-judges. Two authors independently verified that these checklists were reasonable for each prompt. Second, the LLM-judge is given a response to grade along with the previously generated checklist (Checklist-Based Quality Judge Prompt). Grading is based on a Likert scale from 1 to 5, where 1 indicates the response is very poor and does not meet any criteria on the checklist, whereas 5 indicates the response is very good and meets all the criteria on the checklist. \citet{linwildbench} and \citet{wei2025rocketeval} both extensively validate that this checklist-based grading approach has strong correlation with human preferences. 
\subsubsection*{Checklist Creation Prompt}

\begin{promptbox}
\footnotesize
You are an expert evaluator of LLM response quality. Your task is to generate a quality checklist that can be used to evaluate response quality for a given user prompt.\\

\# Instructions\\
First, identify 3 to 5 key factors that should be present in high-quality responses for the given prompt. Since a prompt may have many different high-quality responses, the key factors should be broad enough to cover varied high-quality responses while still being specific to the prompt.\\

Then, turn each key factor into a yes/no question for a checklist to evaluate response quality. Your questions should be concise and specific to the user prompt. Avoid creating duplicate, cumbersome, or vague questions. Do not create more than 5 questions.\\ 

\# User Prompt\\
\{prompt\}\\

\# Output Format\\
First, output the key factors you identified in a short paragraph. Then, output your quality checklist of yes/no questions in the following format, starting with "Here is my quality checklist:". Separate each question with newlines. Do not output any other text after the checklist.\\ 

Here is my quality checklist:\\
1. Question 1\\
2. Question 2\\
...
\end{promptbox}

\clearpage

\subsubsection*{Checklist-Based Quality Judge Prompt}

\begin{promptbox}
\footnotesize
You are an expert evaluator of LLM response quality. Your task is to grade the given LLM response to a user prompt based on the quality checklist below.\\

\# User Prompt\\
\{prompt\}\\

\# Response To Grade\\
\{response\}\\

\# Quality Checklist\\
\{checklist\}\\

\# Instructions\\
First, answer each question in the checklist with "yes" or "no", based on the response to grade.\\

Then, assign a quality score between 1 to 5 for the response, based on the following criteria:\\
- Score 1: The response is very poor and does not meet any criteria on the checklist.\\
- Score 2: The response is poor and meets few criteria on the checklist.\\
- Score 3: The response is fair and meets some criteria on the checklist.\\
- Score 4: The response is good and meets most criteria on the checklist.\\
- Score 5: The response is very good and meets all criteria on the checklist.\\

\# Output Format\\
First, output your answers to the quality checklist as a numbered list. Each answer should start with "yes" or "no" and have a short justification (1-2 sentences).\\

Then, output your quality score for the response. Your output should end with the following sentence, where <score> is the quality score you assigned: "Based on the quality checklist, my quality score for the response is <score>."
\end{promptbox}

\begin{table}[htp] 
\centering
\caption{\centering Examples of Task-Specific Grading Checklists}
\label{tab:checklists}
\small
\resizebox{\textwidth}{!}{
\begin{tabular}{lp{3cm}p{10cm}}
\toprule
\textbf{LLM-Judge} & \textbf{Prompt} & \textbf{Grading Checklist Generated by LLM-Judge} \\
\toprule
GPT-4o & Tell me a story in five sentences about a boy and his dog. & 1. Does the response present a clear and cohesive narrative within five sentences? 2. Are the boy and his dog central to the story, with a clear focus on their interaction or relationship? 3. Does the story elicit some form of engagement (e.g., emotional resonance, curiosity, or intrigue)? 4. Does the story conclude in a satisfying or meaningful way within the constraints of five sentences? 5. Is the story creative and free from overused cliches? \\
\midrule
Claude-4-Sonnet & Tell me a dad joke. & 1. Does the response contain an actual joke? 2. Is the joke appropriate for all ages and family-friendly? 3. Does the joke follow typical dad joke characteristics (puns, wordplay, or simple humor)? 4. Is the response concise and focused on delivering the joke? \\
\midrule
Gemini-2.5-Flash & Write a short love poem with 4 lines. & 1. Is the response structured as a poem?
2. Does the poem express a theme of love?
3. Is the poem exactly four lines long? \\
\midrule
GPT-4o & Why is Issac Newton famous? & 1. Does the response mention at least one of Isaac Newton’s major contributions, such as the laws of motion, the law of universal gravitation, or calculus?  2. Is the information provided accurate and free from factual errors?  3. Does the response maintain relevance to the prompt, focusing on why Isaac Newton is famous? 4. Is the response clear and easy to understand by the intended audience? \\
\midrule
Claude-4-Sonnet & If $2^8=4^x$, what is the value of $x$? & 1. Does the response correctly rewrite the equation using the same base (either base 2 or base 4)? 2. Does the response properly apply exponent rules to solve for x? 3. Does the response show clear, logical steps that lead to the solution? 4. Does the response arrive at the correct answer (x = 4)? 5. Does the response verify the solution by checking it against the original equation? \\
\midrule
Gemini-2.5-Flash & Who was the first scientist to isolate cardiolipin? & 1. Does the response identify the first scientist to isolate cardiolipin?
2. Does the response accurately name the scientist as M. C. Pangborn?
3. Is the information provided factually correct?
4. Is the response concise and directly answer the question? \\
\midrule
GPT-4o & I am going to join a software company next week as a software engineer. What are the things that I should take care of before joining? & 1. Does the response provide actionable steps to prepare technically or brush up on relevant skills for the role?  2. Does the response address logistical preparation, such as ensuring necessary documentation or setup is in place?  3. Does the response offer advice on professional behavior or mindset to adopt for a successful onboarding experience?  4. Are the suggestions relevant for a software engineer and tailored to the context of joining a new company?  5. Is the advice practical and feasible for someone to implement before their first day? \\
\midrule
Claude-4-Sonnet & Pick a number between 1 and 100. Now tell me what you picked. & 1. Does the response include a specific number between 1 and 100 (inclusive)?
2. Does the response clearly state what number was picked?
3. Does the response directly address the user's request without unnecessary complexity or evasion? \\
\midrule
Gemini-2.5-Flash & Write an apology letter to mother and father that I had run away in fear from getting beaten up by my friend. & 1. Does the letter sincerely apologize for running away?
2. Does the letter clearly state that the reason for running away was fear of being beaten by a friend?
3. Does the letter acknowledge the parents' potential worry or hurt caused by the action?
4. Does the letter offer any reassurance to the parents about the future or the child's safety? \\
\bottomrule
\end{tabular}}
\end{table}

\begin{table}[htp] 
\centering
\caption{\centering Examples of ``Very Good'' Responses Under the Checklist-Based Grading Approach\newline With Varying Athene-RM 8B Reward}
\label{tab:quality_examples}
\small
\resizebox{\textwidth}{!}{
\begin{tabular}{cp{15cm}}
\toprule
\textbf{Athene Reward} & \textbf{Prompt/Response} \\
\toprule
& Prompt (Category A): What species of Polymixia fish is named after research scientist Gloria Hollister? \\
\addlinespace
0.82  & The species of Polymixia fish named after research scientist Gloria Hollister is *Polymixia hollisterae*. \\
\addlinespace
0.63  & The fish species *Polymixia hollisterae* commemorates the contributions of Gloria Hollister to science. \\
\midrule
& Prompt (Category B): List a capital city in Africa. \\
\addlinespace
-0.20  & Cairo is the capital city of Egypt. \\
\addlinespace
0.12  & Addis Ababa is the capital city of Ethiopia. \\
\midrule
& Prompt (Category C): Pick a number between 1 and 100. Now tell me what you picked. \\
\addlinespace
0.35  & I picked the number 42. \\
\addlinespace
-0.25  & I picked the number 6. \\
\midrule 
& Prompt (Category D): How many positive whole-number divisors does 196 have? \\
\addlinespace
1.07  & First, perform the prime factorization of 196. Dividing 196 by 2 repeatedly gives \( 2^2 \), and dividing the remaining quotient (49) results in \( 7^2 \). Thus, \( 196 = 2^2 \times 7^2 \). To find the number of divisors, use the formula \((e_1 + 1)(e_2 + 1)\), where \( e_1 \) and \( e_2 \) are the exponents of the prime factors. Here, \( (2+1)(2+1) = 9 \). Hence, 196 has **9 divisors**. \\
\addlinespace
0.34  & Recognize that 196 is a perfect square: \( 14 \times 14 \). Break 14 into its prime factors:  
\( 14 = 2 \times 7 \), so \( 196 = (2 \times 7) \times (2 \times 7) = 2^2 \times 7^2 \).  

Using the rule for finding divisors (\( (e_1 + 1)(e_2 + 1)... \)), where \( e_1, e_2 \) are the exponents in the prime factorization:  
\( (2 + 1)(2 + 1) = 3 \times 3 = 9 \).  

Thus, 196 has **9 positive whole-number divisors**. \\
\midrule 
& Prompt (Category E):  Your necklace has tangled into a very tight knot, and the hotel does not provide any tools for untangling. Available tools include a shampoo, a fork, a safety pin, a pen, a hair brush, a toothpick, a hair dryer, a credit card and hand cream. Constraint: The necklace is made of very fine and delicate chain, so using a safety pin or credit card may damage it. How do you untangle it using just these items? \\
\addlinespace
-0.14  & Rub a tiny amount of hand cream onto the knot to reduce friction between the links. Place the chain flat on a table, holding it steady. Use the prongs of the fork to carefully tease the loops apart, starting with the loosest part of the knot. Work with patience, rotating the fork as needed to avoid pulling too hard on the chain. If the knot is still tight, use the pen to hold one side steady while you maneuver the fork through the loops for additional leverage. \\
\addlinespace
-0.47  & Start by applying a small amount of shampoo or hand cream to the knot. This will act as a lubricant, making the chain slippery and easier to untangle. Next, use the toothpick to gently prod at the knot. Work carefully to loosen the knot without pulling or using too much force, as the chain is delicate. If the knot remains stubborn, gently warm the chain using the hair dryer on a low setting to make the metal more flexible, then repeat the process with the toothpick until the knot comes undone. \\
\midrule
& Prompt (Category F): Can you list 3 influential persons from the nineteenth century? \\
\addlinespace
0.11 & 1. Charles Darwin – Naturalist and biologist known for his theory of evolution by natural selection.
2. Queen Victoria – Monarch of the United Kingdom who oversaw the Victorian Era.
3. Karl Marx – Philosopher, economist, and revolutionary socialist who co-wrote ``The Communist Manifesto.'' \\
\addlinespace
0.25 & 1. Abraham Lincoln – 16th President of the United States who led the country during the Civil War and abolished slavery.
2. Florence Nightingale – Founder of modern nursing and a key figure in improving healthcare.
3. Leo Tolstoy – Russian novelist famous for works like ``War and Peace'' and ``Anna Karenina.'' \\
\midrule 
& Prompt (Category G): Tell me a riddle. \\
\addlinespace 
0.07 & The more of me you take, the more you leave behind. What am I?  
**Answer:** Footsteps. \\
\addlinespace 
0.92 & What has cities, but no houses; forests, but no trees; and rivers, but no water?  
**Answer:** A map. \\
\midrule 
& Prompt (Category H): Suggest a birthday gift for a friend's 25th birthday. Just one. \\
\addlinespace
0.13 & A cooking class voucher or experience for two. This can be a fun way for them to learn something new and enjoy a memorable experience, either solo or with a loved one. \\
\addlinespace
0.53 & A personalized star map that shows the alignment of the stars on the day they were born. It's a unique and sentimental way to celebrate their 25th birthday. \\
\bottomrule
\end{tabular}}
\end{table}

\clearpage

\subsection{Alignment Experiment Details}\label{apx:alignment-details}

\begin{figure}[h!]
  \centering
  \includegraphics[width=\linewidth]{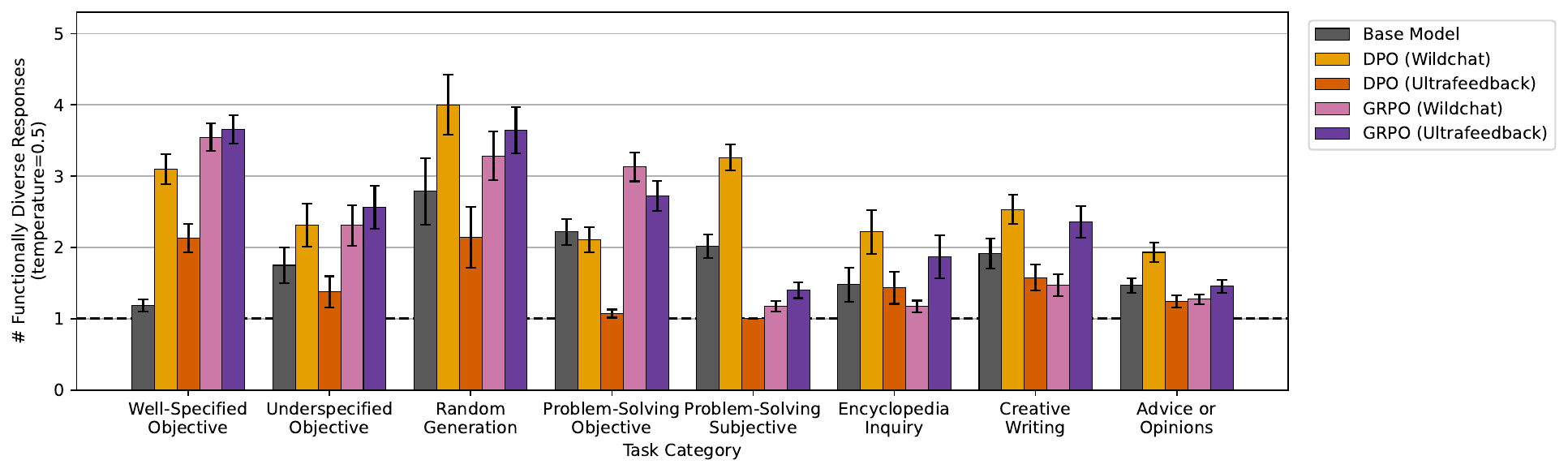}
  \caption{Number of functionally diverse responses generated by Llama-3.1-8B-Instruct, with and without preference alignment. DPO and GRPO results based on 1000 training steps and $\beta=0.01$ and $\beta=0.001$, respectively. Unlike prior results on token entropy~\citep{lanchantin2025bridging}, functional diversity does not collapse and sometimes increases post-alignment.}
  \label{fig:fun_div_alignment}
\end{figure}

\begin{figure}[h!]
  \centering
  \begin{subfigure}[t]{0.49\linewidth}
    \centering
    \includegraphics[width=0.9\linewidth]{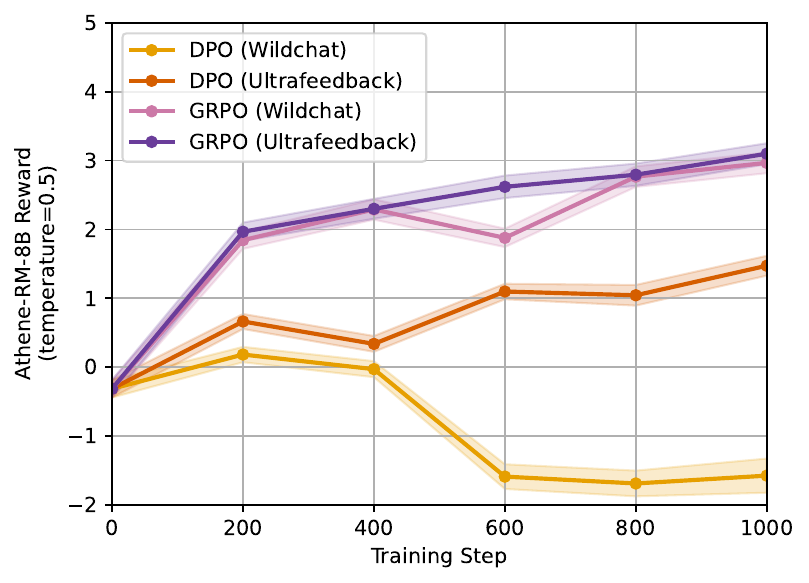}
    \caption{Athene Reward by Training Step}
  \end{subfigure}
  \begin{subfigure}[t]{0.49\linewidth}
    \centering
\includegraphics[width=0.9\linewidth]{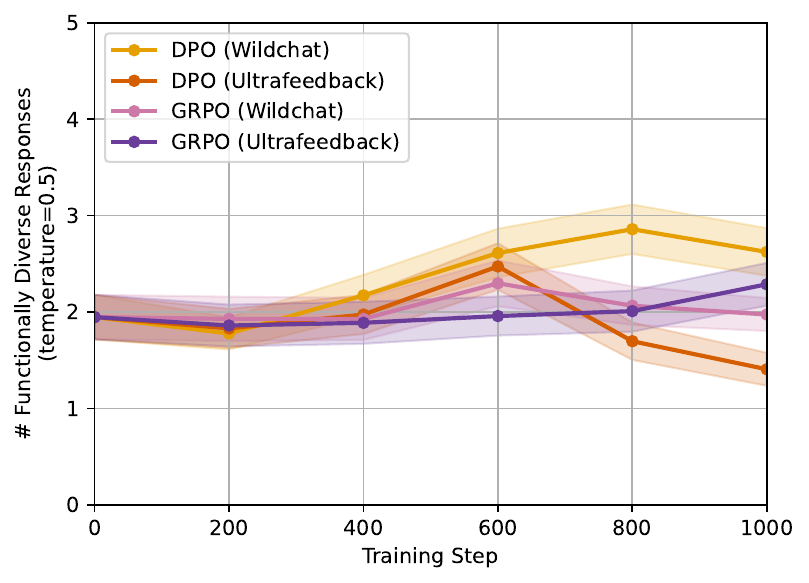}
    \caption{Functional Diversity by Training Step}
  \end{subfigure}
  \caption{During alignment of Llama-3.1-8B-Instruct, the reward generally increases without a collapse in functional diversity. DPO and GRPO results based on $\beta=0.01$ and $\beta=0.001$, respectively. Metrics avg. across all task categories except category A, where homogenization is desired.}
  \label{fig:fun_div_alignment_steps}
\end{figure}

\begin{figure}[h!]
  \centering
  \begin{subfigure}[t]{0.49\linewidth}
    \centering
    \includegraphics[width=0.9\linewidth]{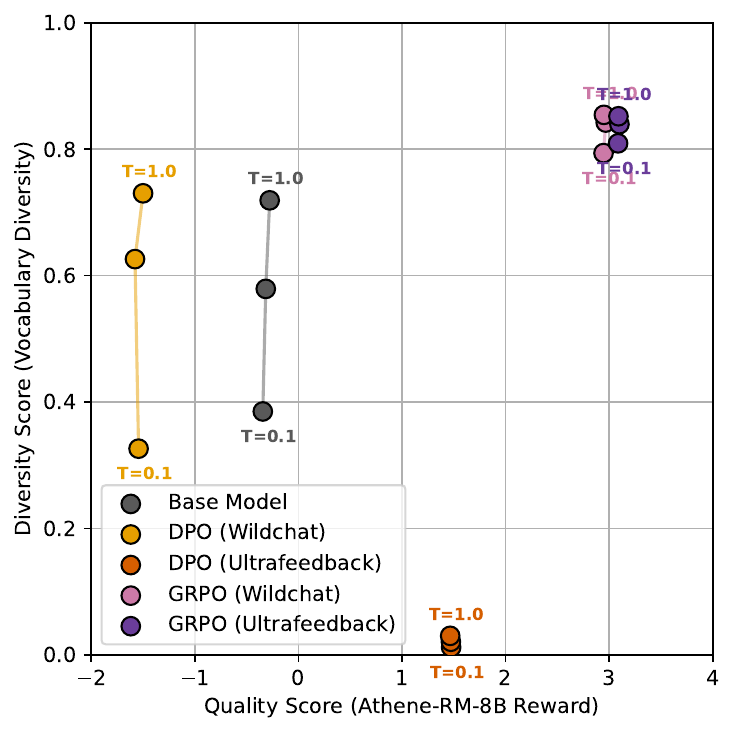}
    \caption{General Metrics}
  \end{subfigure}
  \begin{subfigure}[t]{0.49\linewidth}
    \centering
    \includegraphics[width=0.9\linewidth]{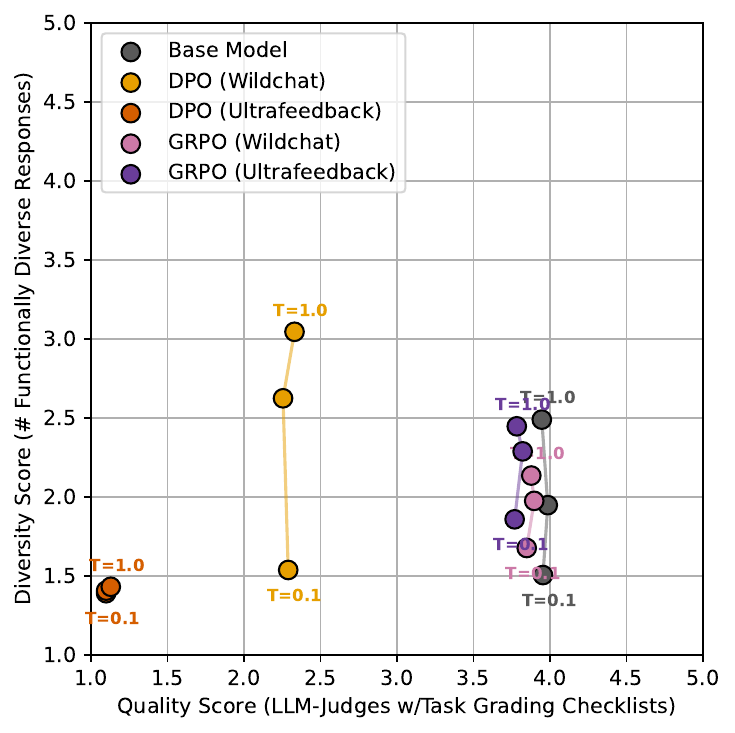}
    \caption{Task-Based Metrics}
  \end{subfigure}
  \caption{Diversity-quality tradeoff under general vs task-based metrics for Llama-3.1-8B-Instruct, with and without preference alignment. DPO and GRPO results based on 1000 training steps and $\beta=0.01$ and $\beta=0.001$, respectively. While DPO and GRPO generally improve reward quality, they do not always improve checklist-based quality. Metrics avg. across all task categories except category A, where homogenization is desired.}
  \label{fig:dq_alignment}
\end{figure}

\begin{figure}[h!]
  \centering
  \begin{subfigure}[t]{\linewidth}
    \centering
    \includegraphics[width=0.9\linewidth]{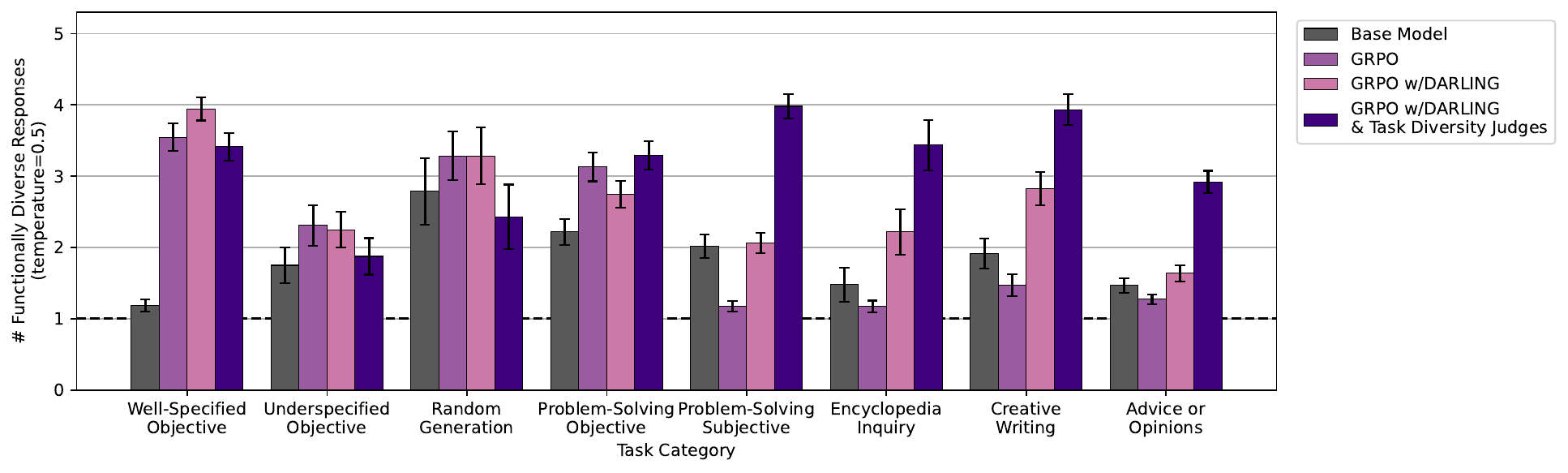}
    \caption{Preference alignment using Wildchat prompts.}
  \end{subfigure}
  \begin{subfigure}[t]{\linewidth}
    \centering
\includegraphics[width=0.9\linewidth]{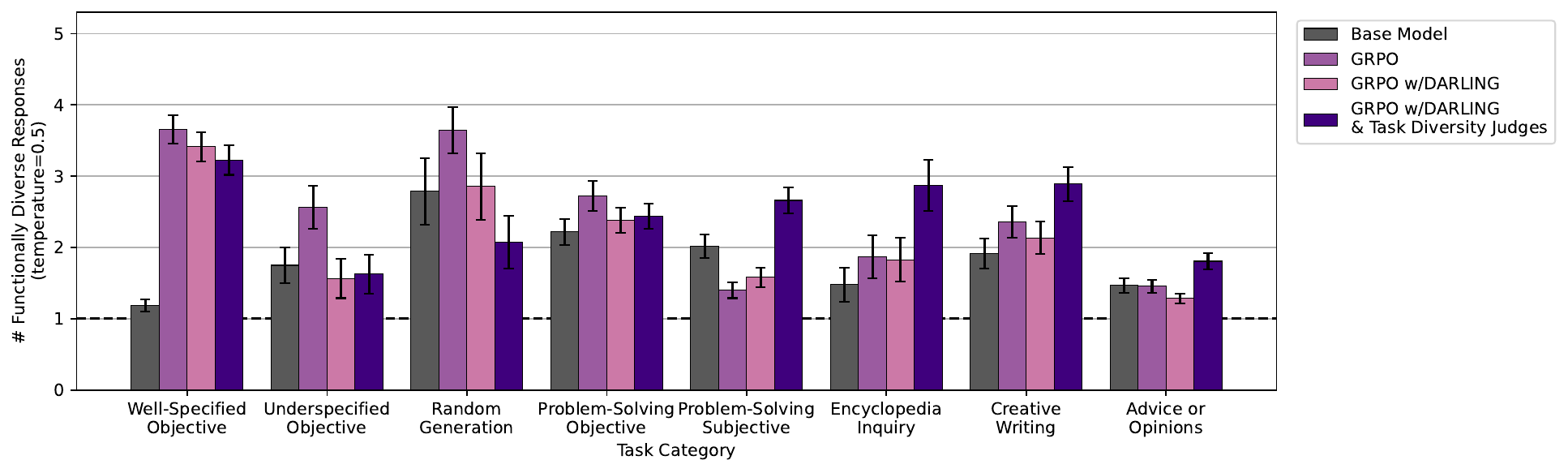}
    \caption{Preference alignment using Ultrafeedback prompts.}
  \end{subfigure}
  \caption{Number of functionally diverse responses generated by Llama-3.1-8B-Instruct, after preference alignment with DARLING~\citep{li2025jointly}. GRPO and DARLING results with $\beta=0.001$. DARLING with task diversity judges uses GPT-4o as the task-dependent functional diversity judge. DARLING generally maintains or improves functional diversity over GRPO, and task diversity judges generally provide further improvement. All alignment methods undesirably reduce homogenization for category A (Well-Specified Objective).}
  \label{fig:fun_div_darling}
\end{figure}

\begin{figure}[h!]
  \centering
  \begin{subfigure}[t]{0.49\linewidth}
    \centering
    \includegraphics[width=0.9\linewidth]{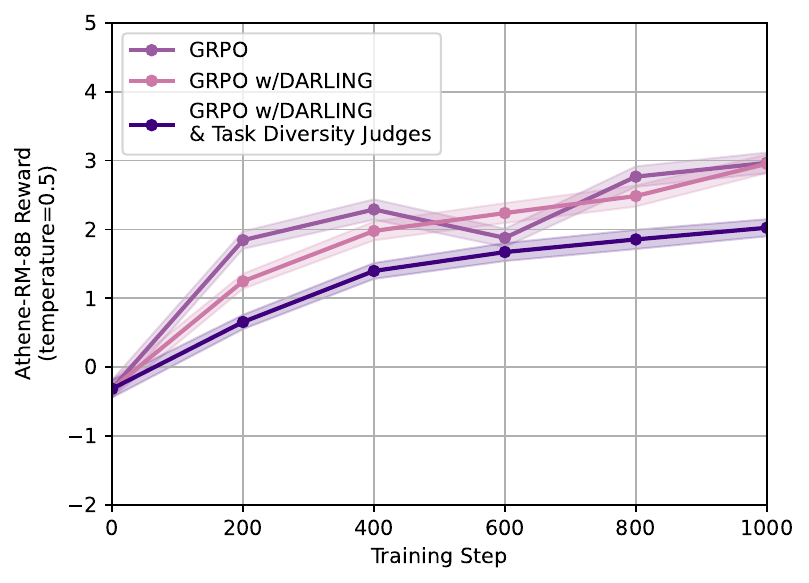}
    \caption{\scriptsize Athene Reward: Alignment w/Wildchat}
  \end{subfigure}
  \begin{subfigure}[t]{0.49\linewidth}
    \centering
\includegraphics[width=0.9\linewidth]{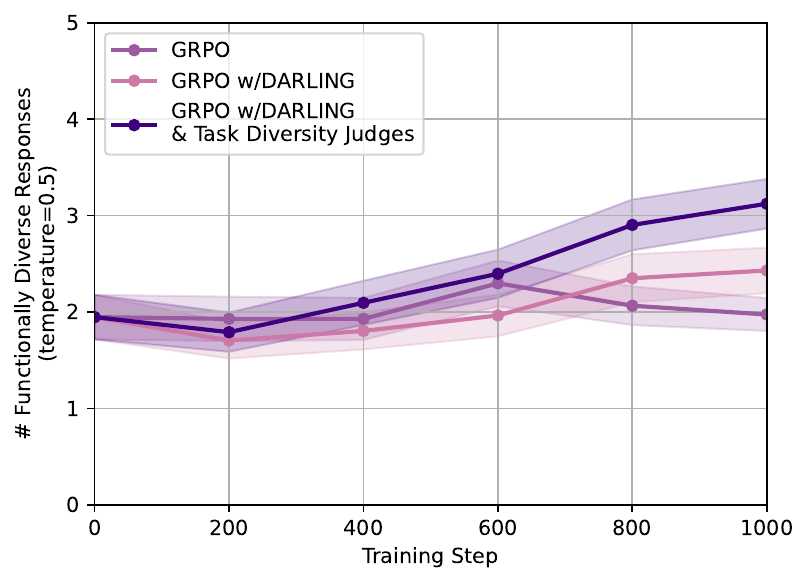}
    \caption{\scriptsize Functional Diversity: Alignment w/Wildchat}
  \end{subfigure}
 \begin{subfigure}[t]{0.49\linewidth}
    \centering
    \includegraphics[width=0.9\linewidth]{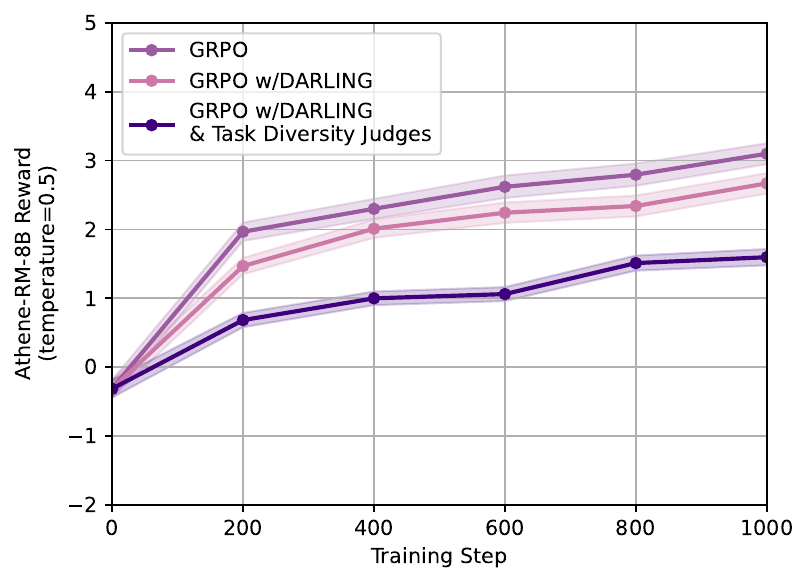}
    \caption{\scriptsize Athene Reward: Alignment w/Ultrafeedback}
  \end{subfigure}
  \begin{subfigure}[t]{0.49\linewidth}
    \centering
\includegraphics[width=0.9\linewidth]{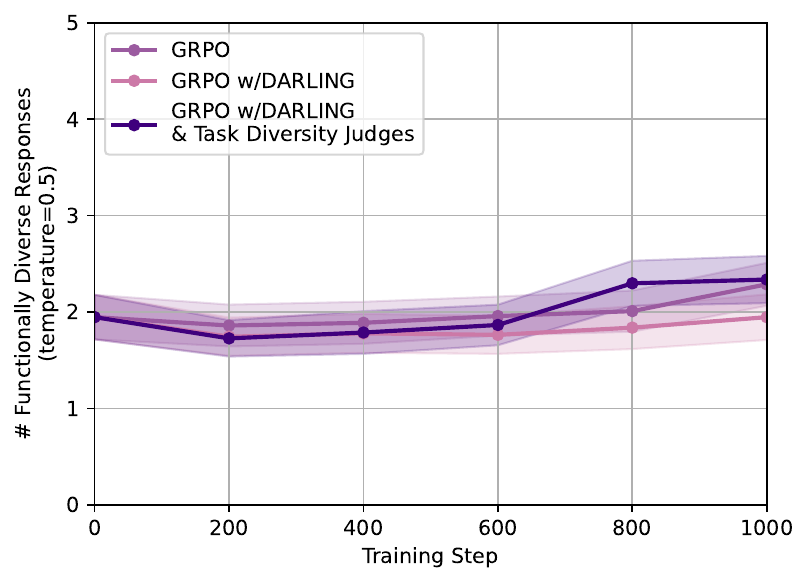}
    \caption{\scriptsize Functional Diversity: Alignment w/Ultrafeedback}
  \end{subfigure}
  \caption{During alignment of Llama-3.1-8B-Instruct with DARLING using \textbf{Wildchat prompts}, both the reward and functional diversity generally increase. GRPO and DARLING use $\beta=0.001$. Metrics avg. across all task categories except category A.}
  \label{fig:fun_div_darling_steps}
\end{figure}

\begin{figure}[h!]
  \centering
  \begin{subfigure}[t]{0.49\linewidth}
    \centering
    \includegraphics[width=0.9\linewidth]{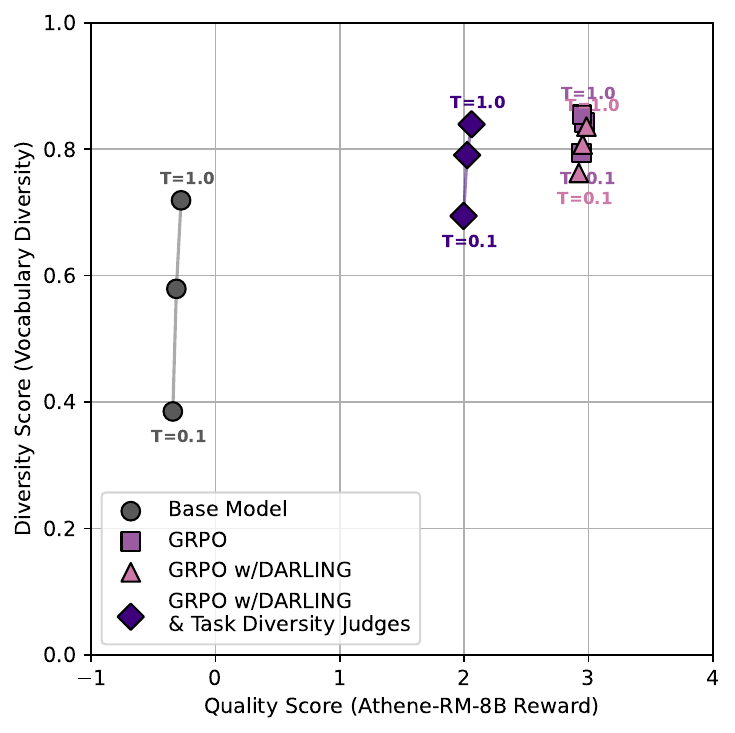}
    \caption{\centering General Metrics:\\Alignment w/Wildchat}
  \end{subfigure}
  \begin{subfigure}[t]{0.49\linewidth}
    \centering
    \includegraphics[width=0.9\linewidth]{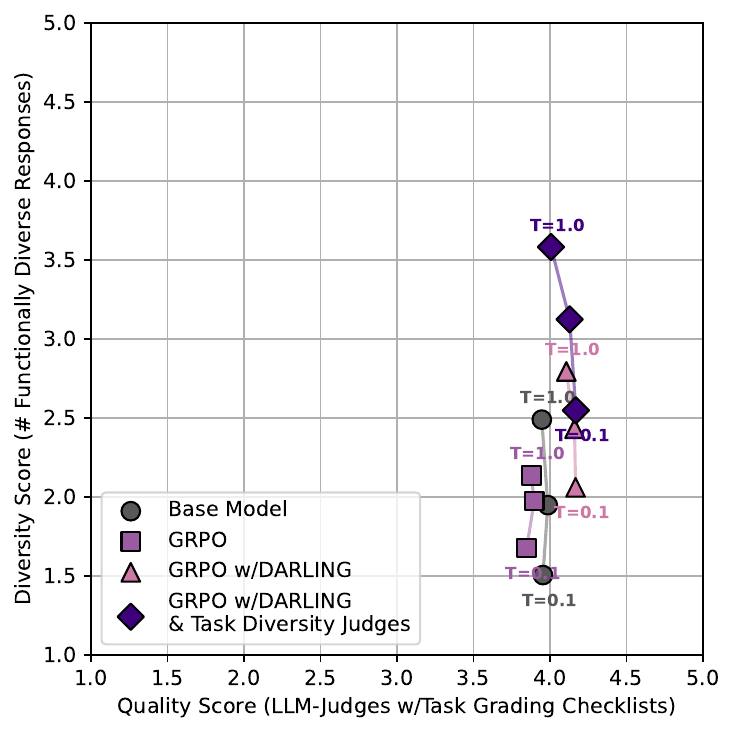}
    \caption{\centering Task-Based Metrics:\\Alignment w/Wildchat}
  \end{subfigure}
  \begin{subfigure}[t]{0.49\linewidth}
    \centering
    \includegraphics[width=0.9\linewidth]{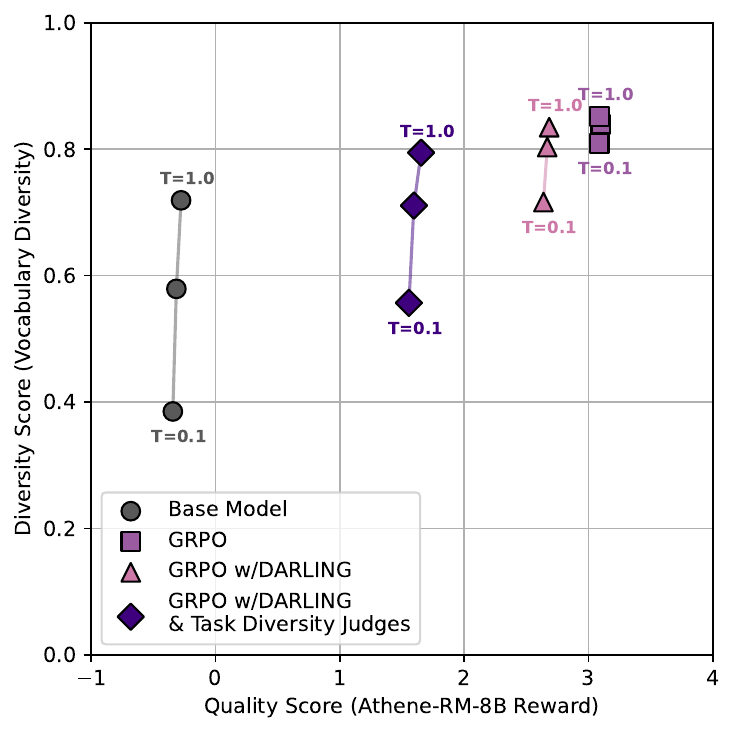}
    \caption{\centering General Metrics:\\Alignment w/Ultrafeedback}
  \end{subfigure}
  \begin{subfigure}[t]{0.49\linewidth}
    \centering
    \includegraphics[width=0.9\linewidth]{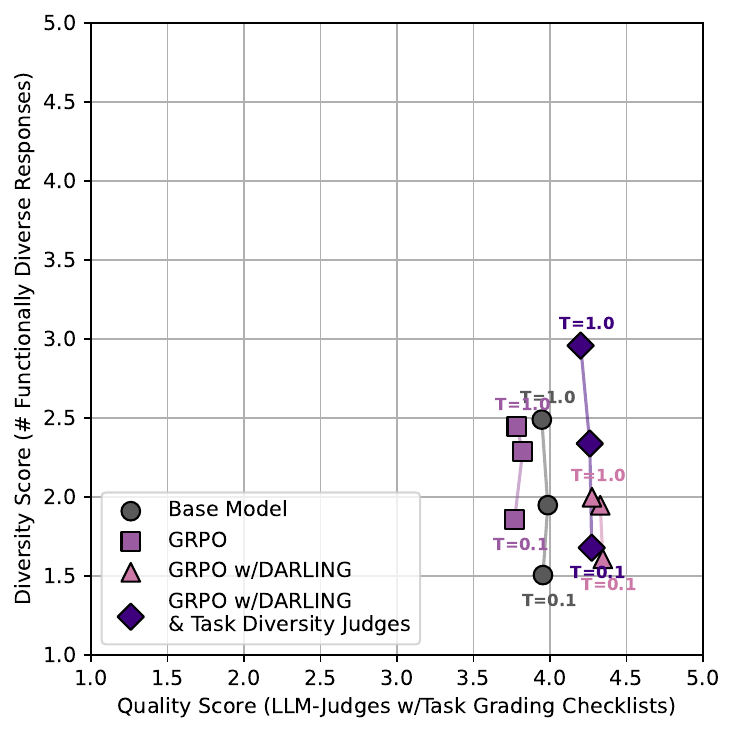}
    \caption{\centering Task-Based Metrics:\\Alignment w/Ultrafeedback}
  \end{subfigure}
  \caption{Diversity-quality tradeoff under general vs task-based metrics for Llama-3.1-8B-Instruct, after preference alignment with DARLING~\citep{li2025jointly}. GRPO and DARLING results based on 1000 training steps and $\beta=0.001$. While general metrics do not show improvements, task-based metrics show that DARLING improves both diversity and quality compared to GRPO.}
  \label{fig:dq_darling}
\end{figure}

\begin{figure}[h!]
    \centering
    \includegraphics[width=\linewidth]{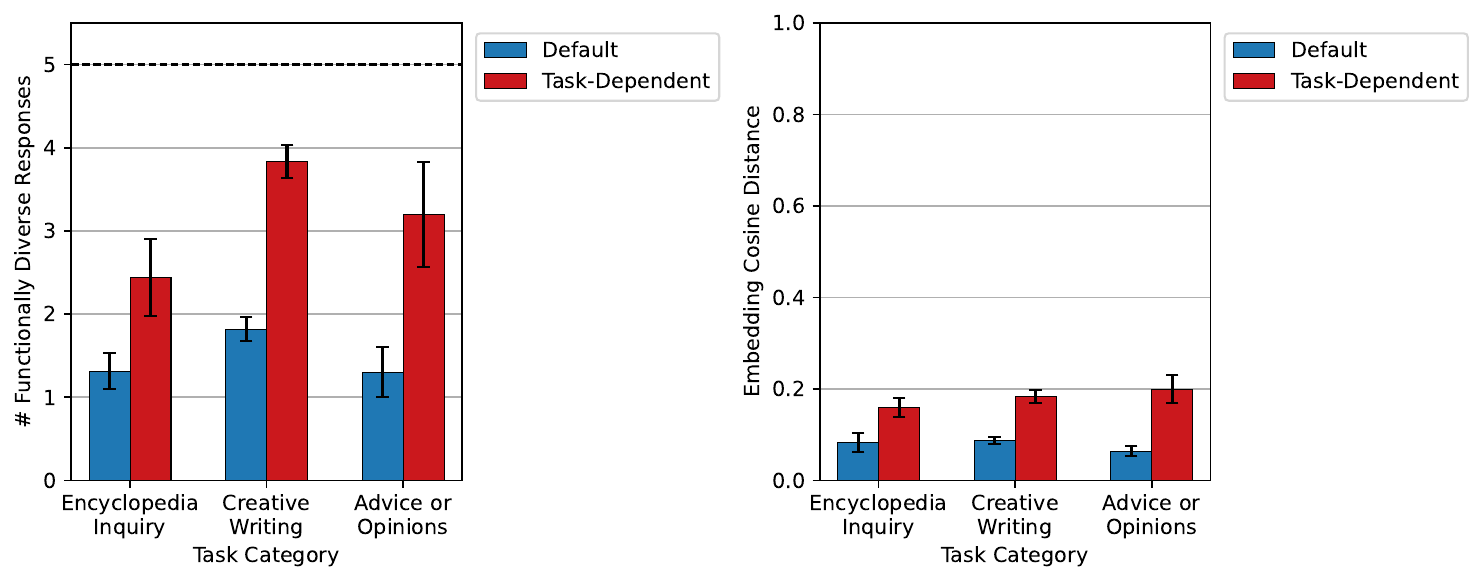}
    \caption{\textbf{Task-dependent sampling substantially increases functional diversity for GPT-4o across all three \textsc{Infinity-Chat100} categories, while embedding distance shows smaller and less consistent gains.}
    Left: average number of functionally diverse responses (out of 5, dashed line) per task category under default (blue) and task-dependent (red) sampling.
    Right: average pairwise embedding cosine distance (BGE-large-en-v1.5) under each strategy.
    Task-dependent sampling improves functional diversity by approximately $1.8\times$ for Encyclopedia Inquiry, $2.1\times$ for Creative Writing, and $2.4\times$ for Advice or Opinions relative to default sampling.
    Embedding distance increases modestly across all categories, consistent with our finding that embedding-based metrics underestimate the gains from task-dependent sampling.
    Error bars denote standard error.}
    \label{fig:infinity_chat_gpt4o_by_category}
\end{figure}

\begin{figure}[h!]
    \centering
    \includegraphics[width=\linewidth]{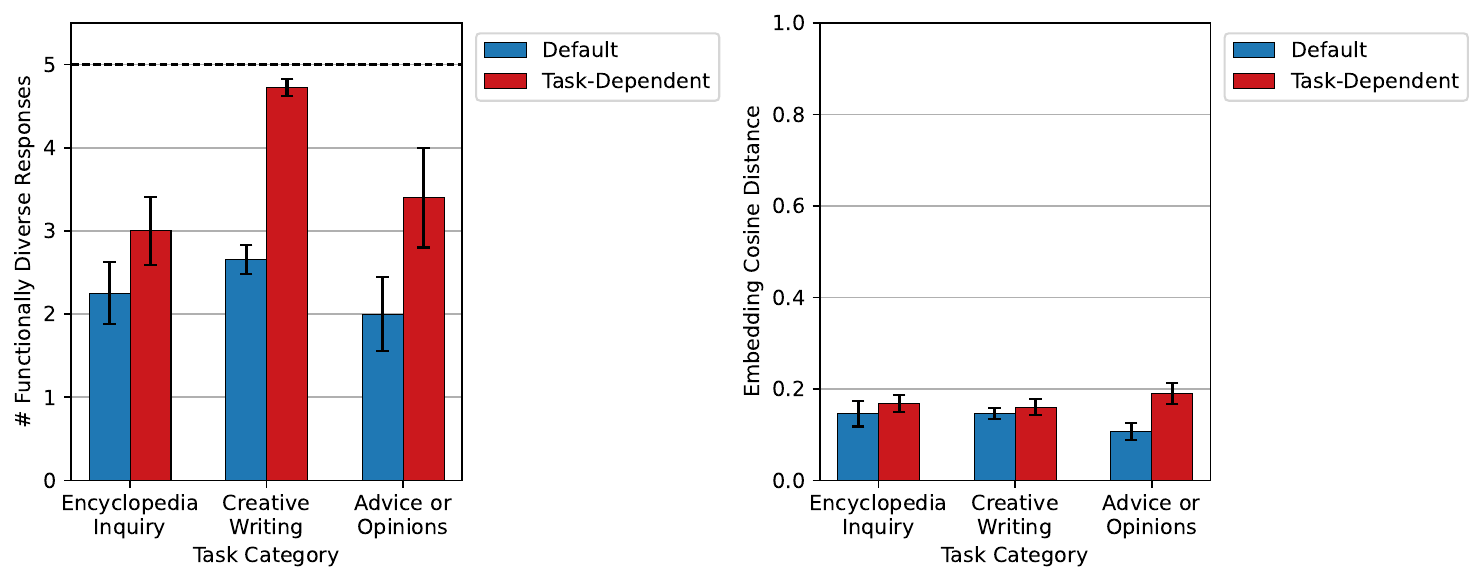}
    \caption{\textbf{Gemini-2.5-Flash exhibits higher baseline functional diversity than GPT-4o on \textsc{Infinity-Chat100}, and task-dependent sampling pushes Creative Writing responses near the maximum.}
    Left: average number of functionally diverse responses per task category.
    Right: average pairwise embedding cosine distance.
    Gemini's default functional diversity is notably higher than GPT-4o's (approximately $2.25$ vs.\ $1.35$ for Encyclopedia Inquiry), yet task-dependent sampling still yields meaningful gains, particularly for Creative Writing (default $\approx 2.65$, task-dependent $\approx 4.75$, near the ceiling of 5).
    Embedding distance gains are minimal for Gemini, reinforcing that embedding-based metrics are insensitive to the functional diversity improvements captured by our taxonomy-guided judges.
    Error bars denote standard error.}
    \label{fig:infinity_chat_gemini_by_category}
\end{figure}

\begin{figure}[h!]
    \centering
    \includegraphics[width=\linewidth]{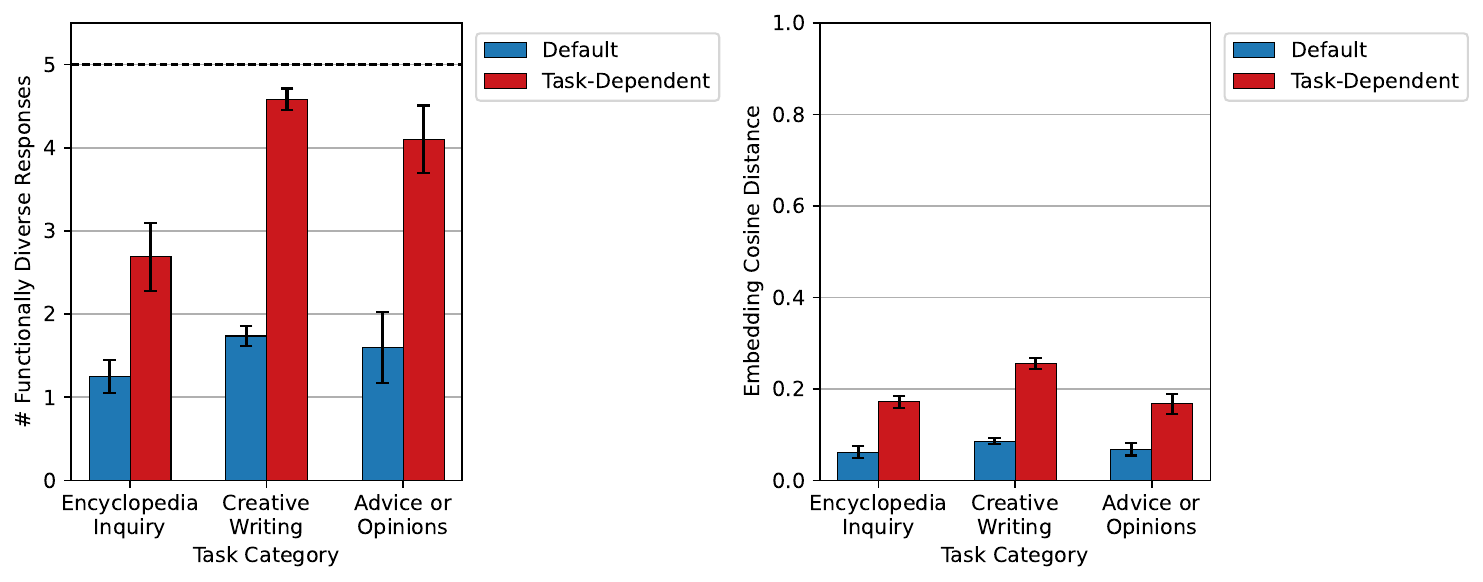}
    \caption{\textbf{Claude-4-Sonnet shows the largest absolute gains from task-dependent sampling on \textsc{Infinity-Chat100}, particularly for Creative Writing and Advice or Opinions.}
    Left: average number of functionally diverse responses per task category.
    Right: average pairwise embedding cosine distance.
    Under default sampling, Claude's functional diversity is comparable to GPT-4o's (approximately $1.25$--$1.75$ across categories).
    Task-dependent sampling raises Creative Writing to $\approx 4.60$ and Advice or Opinions to $\approx 4.10$, the highest absolute values observed across all three models.
    Embedding distance also shows more visible gains for Claude than for the other two models, particularly for Creative Writing ($\approx 0.09 \to 0.25$).
    Error bars denote standard error.}
    \label{fig:infinity_chat_claude_by_category}
\end{figure}

\begin{figure}[h!]
    \centering
    \includegraphics[width=\linewidth]{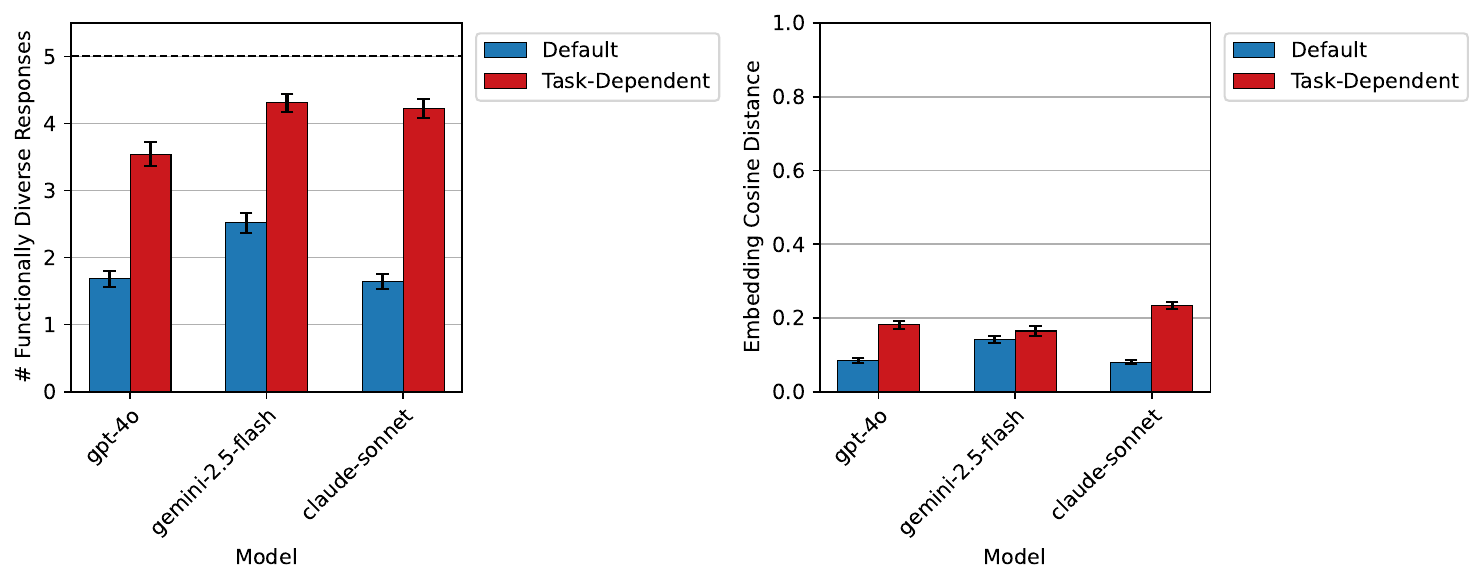}
    \caption{\textbf{Task-dependent sampling consistently improves functional diversity across all three commercial models on \textsc{Infinity-Chat100}, with gains larger than those captured by embedding distance.}
    Results are averaged across all three \textsc{Infinity-Chat100} task categories (Encyclopedia Inquiry, Creative Writing, Advice or Opinions).
    Left: average number of functionally diverse responses (out of 5).
    Right: average pairwise embedding cosine distance (BGE-large-en-v1.5).
    Task-dependent sampling approximately doubles functional diversity for GPT-4o ($1.75 \to 3.60$) and Claude-4-Sonnet ($1.70 \to 4.25$), and yields a substantial gain for Gemini-2.5-Flash ($2.55 \to 4.30$).
    In contrast, embedding distance increases are small and inconsistent across models, underscoring the importance of task-dependent functional diversity as the primary evaluation metric.
    Error bars denote standard error.}
    \label{fig:infinity_chat_by_model_commercial}
\end{figure}

\begin{figure}[h]
    \centering
    \includegraphics[width=\linewidth]{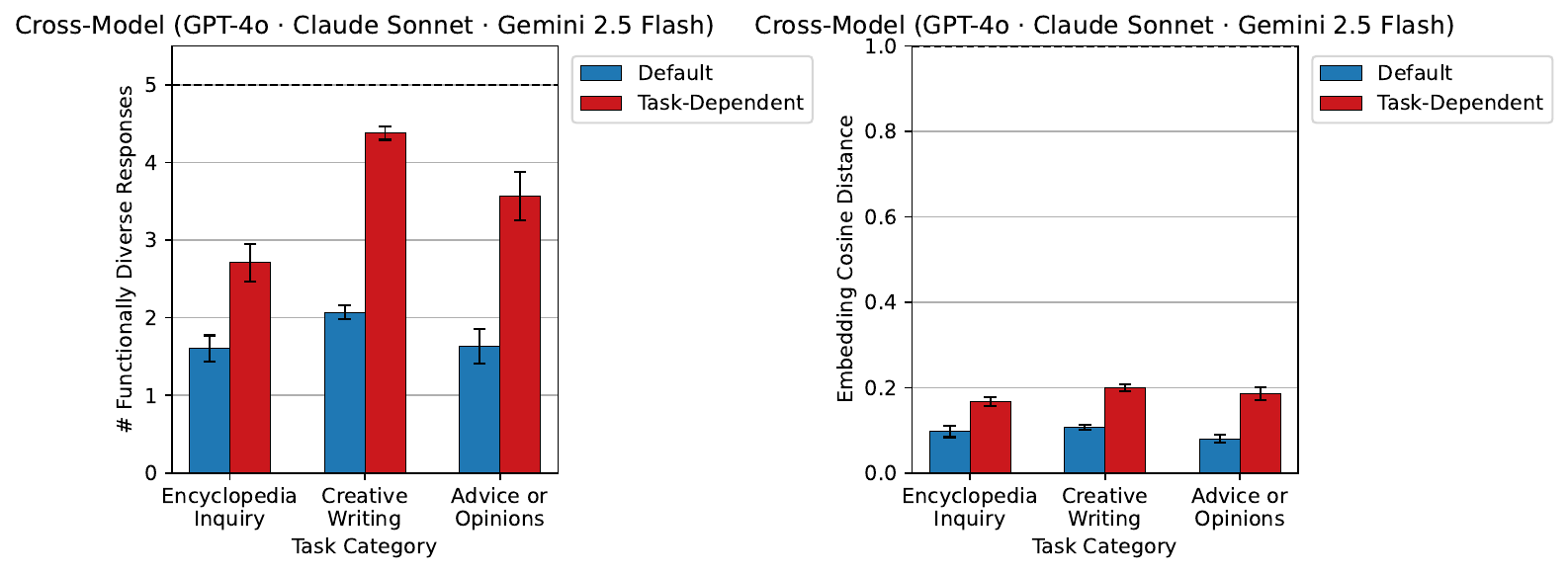}
    \caption{\textbf{Task-dependent sampling reduces cross-model output homogenization on \textsc{Infinity-Chat100}, with the largest gains in Creative Writing and Advice or Opinions.}
    Results pool responses from all three models (GPT-4o, Claude-4-Sonnet, Gemini-2.5-Flash) and measure diversity across the combined set, directly addressing the ``Artificial Hivemind'' effect~\citep{jiang2025artificial}.
    Left: average number of functionally diverse responses across the pooled model outputs per task category.
    Right: average pairwise embedding cosine distance across the pooled outputs.
    Under default sampling, cross-model functional diversity is low across all three categories (approximately $1.60$--$2.10$), consistent with the inter-model homogenization documented in \citet{jiang2025artificial}.
    Task-dependent sampling raises cross-model functional diversity to $\approx 2.70$ (Encyclopedia Inquiry), $\approx 4.45$ (Creative Writing), and $\approx 3.55$ (Advice or Opinions), substantially reducing the hivemind effect.
    Error bars denote standard error.}
    \label{fig:infinity_chat_cross_model_by_category}
\end{figure}

\begin{figure}[h]
    \centering
    \includegraphics[width=\linewidth]{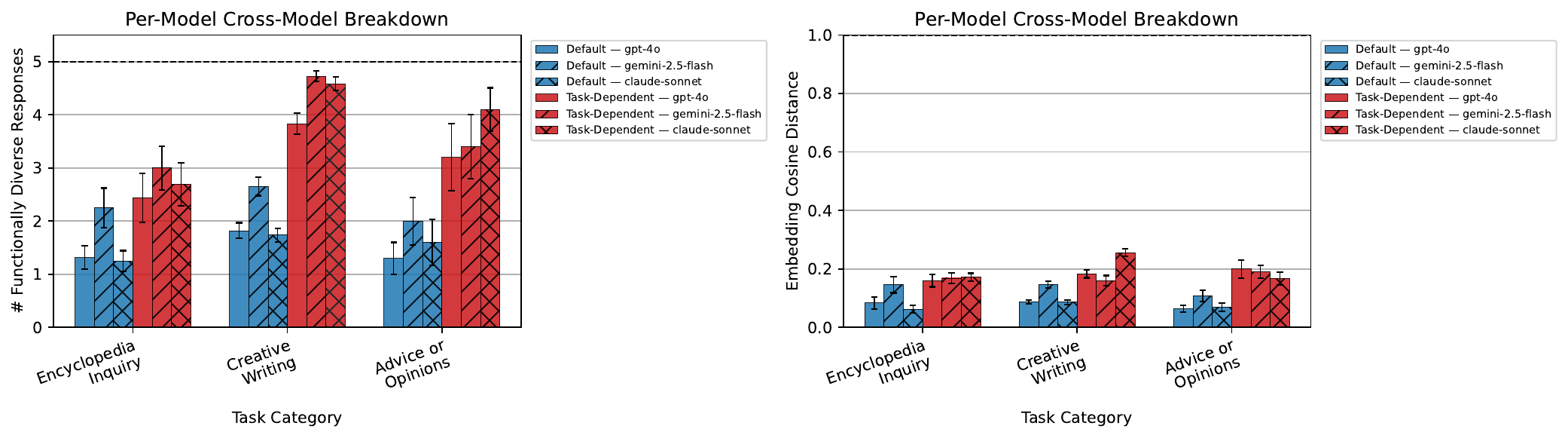}
    \caption{\textbf{Per-model breakdown of cross-model diversity on \textsc{Infinity-Chat100} shows that task-dependent sampling improves functional diversity consistently across models and categories, while embedding distance gains are small.}
    Each group of bars shows default (blue, solid) and task-dependent (red, hatched) results for GPT-4o, Gemini-2.5-Flash, and Claude-4-Sonnet, measured against the pooled responses of all three models.
    Left: number of functionally diverse responses.
    Right: pairwise embedding cosine distance.
    Across all models and categories, task-dependent sampling yields higher functional diversity than default sampling.
    The gains are largest for Creative Writing, where Gemini ($\approx 2.65 \to 4.70$) and Claude ($\approx 1.75 \to 4.60$) approach the ceiling under task-dependent sampling.
    Embedding distance gains remain small and near-uniform across models, further confirming that functional diversity is the more sensitive metric for detecting the improvements our framework provides.
    Error bars denote standard error.}
    \label{fig:infinity_chat_cross_model_per_model}
\end{figure}

We run exploratory alignment experiments to further show the usefulness of our task-dependent framework for evaluating and reducing homogenization. All alignment experiments use \textit{Llama-3.1-8B-Instruct} and the \textit{Athene-RM-8B} reward model, and LLM-judge metrics are only calculated with GPT-4o. First, we demonstrate that functional diversity does not collapse and sometimes increases after preference tuning with Direct Preference Optimization (DPO)~\citep{rafailov2023direct} and Group Relative Policy Optimization (GRPO)~\citep{shao2024deepseekmath} (Figures~\ref{fig:fun_div_alignment}-\ref{fig:dq_alignment}). Second, we show how task-dependent evaluation clarifies the impact of diversity-promoting methods in alignment by evaluating DARLING (Diversity-Aware Reinforcement Learning)~\citep{li2025jointly} (Figures~\ref{fig:fun_div_darling}-\ref{fig:dq_darling}). We further explore how DARLING could be modified to account for task-dependence. 

\paragraph{\textbf{DPO and GRPO}} We run online DPO and GRPO following the training recipe for non-verifiable rewards in \citet{lanchantin2025bridging} and \citet{li2025jointly}, respectively. In particular, we use a learning rate of $1e{-6}$, batch size of 32, and train for 1000 steps. At each step, we generate 8 responses per prompt with temperature 1.0 and 1024 max tokens. For DPO, preference pairs are constructed based on the responses with the maximum and minimum reward. For GRPO, all 8 responses are used to calculate advantage. We explore two values of $\beta$: 0.01 and 0.1 for DPO, and 0.001 and 0.01 for GRPO. 

We also explore using two datasets for preference tuning: Wildchat~\citep{zhaowildchat} and Ultrafeedback~\citep{cui2023ultrafeedback}. For Wildchat, we use 10,000 randomly sampled prompts. The majority of these prompts (5,535) corresponded to category G (Creative Writing), based on task classification by GPT-4o. Thus, for Ultrafeedback, we try a stratified random sample of 10,000 prompts based on each task category in our taxonomy and GPT-4o as the task classification judge. Specifically, we sample 2,500 prompts for each task category, excluding category E (Problem-Solving Subjective) and combining categories B and C (Underspecified Objective and Random Generation) due to prompt availability. In both datasets, we exclude prompts with more than 512 tokens. 

\paragraph{\textbf{DARLING}} \citet{li2025jointly} propose DARLING (Diversity-Aware Reinforcement Learning), which modifies the GRPO reward to jointly reinforce diversity and quality. Specifically, they scale reward by the diversity $d(y_i|y_1,\dots,y_n)$ of a generation $y_i$, which they define as the average pairwise distance between $y_i$ and all other generations $y_j$ ($j \neq i$), normalized to be between 0 and 1. They implement their method using ``semantic uniqueness'' as their distance metric, which represents a general notion of functional diversity (not task-dependent). They fine-tune a \textit{ModernBERT-base} model to predict semantic uniqueness based on 1000 human annotations from NoveltyBench~\citep{zhang2025noveltybench}.

We explore modifying DARLING to account for task-dependence by using GPT-4o as a task-dependent functional diversity judge, in place of the fine-tuned ModernBert classifier. For prompts in category A (Well-Specified Objective), we also modify DARLING to scale the reward by $1-d(y_i|y_1,\dots,y_n)$, which promotes homogenization instead of diversity.

\subsection{Infinity-Chat Experiment Details}\label{apx:infinity-chat-details}

We evaluate our task-dependent sampling framework on the 100 prompt subset of the \textsc{Infinity-Chat} homogenization benchmark (Infinity-Chat100)~\citep{jiang2025artificial}. We report these results separately since \textsc{Infinity-Chat} was released concurrently to our work. The benchmark curates open-ended prompts from WildChat~\citep{zhaowildchat} that should elicit diverse responses. We classify the Infinity-Chat100 prompts into our taxonomy, and 98 prompts map to three of our taxonomy categories: Encyclopedia Inquiry (F), Creative Writing (G), and Advice or Opinions (H). Two prompts map to Underspecified Singular Objective (B), which we exclude from our analysis as outliers. We evaluate 3 commercial models (GPT-4o, Gemini-2.5-Flash, Claude-4-Sonnet). We sample 5 responses per prompt using 2 sampling strategies: default temperature-sampling (medium values) and task-dependent system prompt sampling. We report task-dependent functional diversity (using our taxonomy-guided LLM-judges), and embedding diversity (using BGE-large-en-v1.5).

\clearpage
\section{Additional Experiment Results}\label{apx:exp_results}

\subsection{Human Validation Experiment Results}\label{apx:human_exp_results}

\begin{table}[h!]
\centering
\caption{\centering Inter-Rater Reliability (Human Annotators)}
\scriptsize
\begin{tabular}{llrrrrrrrr}
\toprule
Metric & Annotator Group  & All & A & B & C & E & F & G & H \\
\midrule
Krippendorff $\alpha$ & General & 0.660 & 0.834 & 0.947 & 0.431 & 0.581 & 0.492 & 0.257 & 0.699 \\
Krippendorff $\alpha$ & Task-Dependent & 0.671 & 1.000 & 0.895 & 0.743 & 0.574 & 0.411 & 0.328 & 0.494 \\
Agreement & General & 0.897 & 0.988 & 0.988 & 0.873 & 0.915 & 0.855 & 0.739 & 0.921 \\
Agreement & Task-Dependent & 0.904 & 1.000 & 0.976 & 0.964 & 0.903 & 0.830 & 0.794 & 0.861 \\
\bottomrule
\end{tabular}
\label{tab::IRR-full}
\end{table}

\begin{table}[h!]
\caption{\centering Spearman Correlation Between Human Annotator Groups (General vs Task-Dependent)}
\centering
\scriptsize
\begin{tabular}{lrrrrrrrr}
\toprule
Correlation of... & All & A & B & C & E & F & G & H \\
\midrule
Majority Vote & 0.90 & 1.00 & 1.00 & 1.00 & 1.00 & 0.85 & 0.61 & 0.85 \\
Average Label & 0.82 & 0.62 & 0.71 & 0.27 & 0.75 & 0.92 & 0.90 & 0.71 \\
\bottomrule
\end{tabular}
\label{tab:annotator_groups_corr}
\end{table}

\begin{table}[h!]
\caption{\centering Spearman Correlation Between Diversity Metrics \& Majority Vote of Task-Dependent Human Annotations (c.f. Table~\ref{tab:annotation_res_main})} 
\scriptsize
\centering
\begin{tabular}{lcccccccc}
\toprule
\multirow{2}{*}{Metric} &  \multicolumn{8}{c}{Task Category} \\
\cline{2-9}
\addlinespace[0.3em]
 & All & A & B & C & E & F & G & H \\
\midrule
\multicolumn{9}{l}{\uline{General Diversity Metrics}} \\
\addlinespace[0.1em]
Vocabulary Diversity & 0.64 & 0.25 & 0.51 & 0.44 & 0.46 & 0.70 & 0.54 & 0.70 \\
Embedding Diversity & 0.80 & 0.43 & 0.82 & 0.59 & 0.66 & 0.77 & 0.66 & 0.77 \\
Compression Diversity & 0.26 & 0.06 & 0.62 & 0.45 & 0.62 & 0.73 & 0.39 & 0.49 \\
\textsc{Novelty-Bench} Functional Diversity & 0.59 & 0.37 & 0.13 & 0.45 & 0.69 & 0.59 & 0.62 & 0.73 \\
\midrule
\multicolumn{9}{l}{\uline{Task-Dependent Functional Diversity (Taxonomy-Guided LLM-Judges)}} \\
\addlinespace[0.1em]
\textbf{(ours)} GPT-4o & 0.88 & 1.00 & 1.00 & 1.00 & 0.83 & 0.64 & 0.83 & 0.85 \\
\textbf{(ours)}  Gemini-2.5-Flash & 0.83 & 1.00 & 1.00 & 1.00 & 1.00 & 0.74 & 0.41 & 0.85 \\
\textbf{(ours)}  Claude-4-Sonnet & 0.88 & 1.00 & 1.00 & 1.00 & 0.83 & 0.74 & 0.71 & 0.85 \\
\bottomrule
\end{tabular}
\label{tab:annotation_res_appx}
\end{table}

\begin{table}[h!]
\centering 
\caption{\centering Correlation Between Diversity Metrics and Human Annotation (Average Label)}
\scriptsize
\begin{tabular}{llcccccccc}
\toprule
\multirow{2}{*}{Annotator Group} & \multirow{2}{*}{Metric} &  \multicolumn{8}{c}{Task Category} \\
\cline{3-10}
\addlinespace[0.3em]
 & & All & A & B & C & E & F & G & H \\
\midrule
\multicolumn{10}{c}{\uline{General Diversity Metrics}} \\
\addlinespace[0.1em]
General & Vocabulary Diversity & 0.58 & 0.50 & 0.53 & 0.06 & 0.77 & 0.79 & 0.49 & 0.50 \\
General & Embedding Diversity & 0.68 & 0.41 & 0.66 & 0.09 & 0.78 & 0.73 & 0.72 & 0.61 \\
General & Compression Diversity & 0.27 & 0.22 & 0.70 & 0.38 & 0.47 & 0.75 & 0.20 & 0.50 \\
General & \textsc{Novelty-Bench} Functional Diversity & 0.57 & 0.37 & 0.26 & 0.49 & 0.76 & 0.77 & 0.80 & 0.70 \\
\midrule
Task-Dependent & Vocabulary Diversity & 0.58 & 0.25 & 0.33 & 0.18 & 0.67 & 0.88 & 0.43 & 0.72 \\
Task-Dependent & Embedding Diversity & 0.76 & 0.43 & 0.75 & 0.43 & 0.81 & 0.76 & 0.84 & 0.84 \\
Task-Dependent & Compression Diversity & 0.26 & 0.06 & 0.38 & 0.25 & 0.74 & 0.71 & 0.05 & 0.24 \\
Task-Dependent & \textsc{Novelty-Bench} Functional Diversity & 0.59 & 0.37 & 0.13 & 0.45 & 0.69 & 0.59 & 0.62 & 0.73 \\
\midrule
\multicolumn{10}{c}{\uline{Task-Dependent Functional Diversity (Taxonomy-Guided LLM-Judges)}} \\
\addlinespace[0.1em]
General & GPT-4o & 0.85 & 0.62 & 0.91 & 0.62 & 0.80 & 0.89 & 0.77 & 0.84 \\
General &  Gemini-2.5-Flash & 0.83 & 0.62 & 0.91 & 0.62 & 0.72 & 0.87 & 0.41 & 0.84 \\
General &  Claude-4-Sonnet & 0.81 & 0.62 & 0.91 & 0.62 & 0.64 & 0.64 & 0.81 & 0.84 \\
\midrule
Task-Dependent & GPT-4o & 0.84 & 1.00 & 0.87 & 0.65 & 0.72 & 0.82 & 0.76 & 0.83 \\
Task-Dependent &  Gemini-2.5-Flash & 0.84 & 1.00 & 0.87 & 0.65 & 0.73 & 0.82 & 0.67 & 0.83 \\
Task-Dependent &  Claude-4-Sonnet & 0.85 & 1.00 & 0.87 & 0.65 & 0.81 & 0.69 & 0.83 & 0.83 \\
\bottomrule
\end{tabular}
\label{tab:annotation_res_using_avg}
\end{table}

\clearpage

\subsection{Task-Dependent Diversity Evaluation}\label{apx:diversity_results}

\clearpage

\begin{table}[h!]
\centering
\caption{\centering \# of Functionally Diverse Responses by Model, Sampling Strategy \& Task Category.}
\label{tab:fun_div}
\tiny
\begin{tabular}{cccccccccc}
\toprule
\makecell{Model} & \makecell{Sampling Strategy} & \makecell{A} & \makecell{B} & \makecell{C} & \makecell{D} & \makecell{E} & \makecell{F} & \makecell{G} & \makecell{H} \\
\midrule
gpt-4o & \makecell{Temperature \\ (t=0.0)} & \makecell{1.57 \\ (0.11)} & \makecell{1.19 \\ (0.10)} & \makecell{1.86 \\ (0.31)} & \makecell{1.33 \\ (0.09)} & \makecell{1.22 \\ (0.07)} & \makecell{1.13 \\ (0.10)} & \makecell{1.18 \\ (0.07)} & \makecell{1.08 \\ (0.03)} \\
\addlinespace[0.5em]
gpt-4o & \makecell{Temperature \\ (t=1.0)} & \makecell{2.11 \\ (0.19)} & \makecell{1.50 \\ (0.18)} & \makecell{3.14 \\ (0.46)} & \makecell{1.27 \\ (0.10)} & \makecell{1.30 \\ (0.10)} & \makecell{1.13 \\ (0.13)} & \makecell{1.40 \\ (0.13)} & \makecell{1.19 \\ (0.06)} \\
\addlinespace[0.5em]
gpt-4o & \makecell{Temperature \\ (t=2.0)} & \makecell{2.38 \\ (0.21)} & \makecell{1.69 \\ (0.25)} & \makecell{3.14 \\ (0.42)} & \makecell{1.31 \\ (0.10)} & \makecell{1.32 \\ (0.10)} & \makecell{1.09 \\ (0.09)} & \makecell{1.89 \\ (0.19)} & \makecell{1.24 \\ (0.06)} \\
\addlinespace[0.5em]
gpt-4o & \makecell{In-Context Regeneration \\ (General)} & \makecell{2.19 \\ (0.22)} & \makecell{4.94 \\ (0.06)} & \makecell{5.00 \\ (0.00)} & \makecell{1.24 \\ (0.10)} & \makecell{1.84 \\ (0.16)} & \makecell{1.35 \\ (0.24)} & \makecell{2.44 \\ (0.26)} & \makecell{2.84 \\ (0.20)} \\
\addlinespace[0.5em]
gpt-4o & \makecell{In-Context Regeneration \\ (Task-Dependent)} & \makecell{1.06 \\ (0.03)} & \makecell{5.00 \\ (0.00)} & \makecell{5.00 \\ (0.00)} & \makecell{1.31 \\ (0.07)} & \makecell{2.82 \\ (0.22)} & \makecell{1.74 \\ (0.28)} & \makecell{3.31 \\ (0.25)} & \makecell{3.20 \\ (0.20)} \\
\addlinespace[0.5em]
gpt-4o & \makecell{System Prompt \\ (General)} & \makecell{1.94 \\ (0.22)} & \makecell{5.00 \\ (0.00)} & \makecell{5.00 \\ (0.00)} & \makecell{1.10 \\ (0.06)} & \makecell{3.80 \\ (0.19)} & \makecell{2.30 \\ (0.39)} & \makecell{2.96 \\ (0.27)} & \makecell{3.65 \\ (0.19)} \\
\addlinespace[0.5em]
gpt-4o & \makecell{System Prompt \\ (Task-Dependent)} & \makecell{1.00 \\ (0.00)} & \makecell{5.00 \\ (0.00)} & \makecell{5.00 \\ (0.00)} & \makecell{1.69 \\ (0.13)} & \makecell{4.60 \\ (0.11)} & \makecell{2.95 \\ (0.37)} & \makecell{4.13 \\ (0.20)} & \makecell{4.20 \\ (0.15)} \\
\addlinespace[0.5em]
\midrule
\addlinespace[0.5em]
claude-4-sonnet & \makecell{Temperature \\ (t=0.0)} & \makecell{1.06 \\ (0.03)} & \makecell{1.00 \\ (0.00)} & \makecell{1.29 \\ (0.22)} & \makecell{1.02 \\ (0.02)} & \makecell{1.40 \\ (0.11)} & \makecell{1.04 \\ (0.04)} & \makecell{1.13 \\ (0.06)} & \makecell{1.10 \\ (0.04)} \\
\addlinespace[0.5em]
claude-4-sonnet & \makecell{Temperature \\ (t=0.5)} & \makecell{1.21 \\ (0.08)} & \makecell{1.06 \\ (0.06)} & \makecell{1.64 \\ (0.32)} & \makecell{1.00 \\ (0.00)} & \makecell{1.36 \\ (0.10)} & \makecell{1.17 \\ (0.10)} & \makecell{1.24 \\ (0.11)} & \makecell{1.14 \\ (0.05)} \\
\addlinespace[0.5em]
claude-4-sonnet & \makecell{Temperature \\ (t=1.0)} & \makecell{1.28 \\ (0.08)} & \makecell{1.19 \\ (0.19)} & \makecell{2.14 \\ (0.35)} & \makecell{1.04 \\ (0.03)} & \makecell{1.40 \\ (0.10)} & \makecell{1.13 \\ (0.07)} & \makecell{1.47 \\ (0.12)} & \makecell{1.25 \\ (0.07)} \\
\addlinespace[0.5em]
claude-4-sonnet & \makecell{In-Context Regeneration \\ (General)} & \makecell{1.60 \\ (0.13)} & \makecell{5.00 \\ (0.00)} & \makecell{4.93 \\ (0.07)} & \makecell{1.38 \\ (0.11)} & \makecell{4.02 \\ (0.19)} & \makecell{1.57 \\ (0.29)} & \makecell{2.98 \\ (0.27)} & \makecell{2.98 \\ (0.20)} \\
\addlinespace[0.5em]
claude-4-sonnet & \makecell{In-Context Regeneration \\ (Task-Dependent)} & \makecell{1.08 \\ (0.08)} & \makecell{5.00 \\ (0.00)} & \makecell{5.00 \\ (0.00)} & \makecell{2.49 \\ (0.18)} & \makecell{4.94 \\ (0.03)} & \makecell{3.13 \\ (0.30)} & \makecell{4.62 \\ (0.15)} & \makecell{4.11 \\ (0.14)} \\
\addlinespace[0.5em]
claude-4-sonnet & \makecell{System Prompt \\ (General)} & \makecell{1.17 \\ (0.10)} & \makecell{5.00 \\ (0.00)} & \makecell{5.00 \\ (0.00)} & \makecell{1.31 \\ (0.11)} & \makecell{4.40 \\ (0.14)} & \makecell{2.00 \\ (0.32)} & \makecell{3.05 \\ (0.27)} & \makecell{3.50 \\ (0.19)} \\
\addlinespace[0.5em]
claude-4-sonnet & \makecell{System Prompt \\ (Task-Dependent)} & \makecell{1.08 \\ (0.08)} & \makecell{5.00 \\ (0.00)} & \makecell{5.00 \\ (0.00)} & \makecell{1.85 \\ (0.16)} & \makecell{4.86 \\ (0.06)} & \makecell{2.61 \\ (0.33)} & \makecell{4.20 \\ (0.19)} & \makecell{4.32 \\ (0.13)} \\
\addlinespace[0.5em]
\midrule
\addlinespace[0.5em]
gemini-2.5-flash & \makecell{Temperature \\ (t=0.0)} & \makecell{1.25 \\ (0.07)} & \makecell{1.12 \\ (0.09)} & \makecell{1.57 \\ (0.17)} & \makecell{1.02 \\ (0.02)} & \makecell{1.26 \\ (0.06)} & \makecell{1.09 \\ (0.06)} & \makecell{1.13 \\ (0.05)} & \makecell{1.07 \\ (0.03)} \\
\addlinespace[0.5em]
gemini-2.5-flash & \makecell{Temperature \\ (t=1.0)} & \makecell{2.55 \\ (0.21)} & \makecell{1.81 \\ (0.26)} & \makecell{3.21 \\ (0.41)} & \makecell{1.13 \\ (0.06)} & \makecell{2.04 \\ (0.17)} & \makecell{1.22 \\ (0.11)} & \makecell{1.87 \\ (0.21)} & \makecell{1.40 \\ (0.10)} \\
\addlinespace[0.5em]
gemini-2.5-flash & \makecell{Temperature \\ (t=2.0)} & \makecell{2.68 \\ (0.20)} & \makecell{1.88 \\ (0.20)} & \makecell{3.29 \\ (0.41)} & \makecell{1.05 \\ (0.04)} & \makecell{1.84 \\ (0.14)} & \makecell{1.39 \\ (0.14)} & \makecell{2.22 \\ (0.23)} & \makecell{1.61 \\ (0.12)} \\
\addlinespace[0.5em]
gemini-2.5-flash & \makecell{In-Context Regeneration \\ (General)} & \makecell{1.30 \\ (0.12)} & \makecell{4.94 \\ (0.06)} & \makecell{5.00 \\ (0.00)} & \makecell{1.35 \\ (0.11)} & \makecell{1.54 \\ (0.13)} & \makecell{1.22 \\ (0.18)} & \makecell{2.31 \\ (0.25)} & \makecell{2.85 \\ (0.20)} \\
\addlinespace[0.5em]
gemini-2.5-flash & \makecell{In-Context Regeneration \\ (Task-Dependent)} & \makecell{1.06 \\ (0.06)} & \makecell{5.00 \\ (0.00)} & \makecell{5.00 \\ (0.00)} & \makecell{1.49 \\ (0.10)} & \makecell{3.04 \\ (0.19)} & \makecell{1.61 \\ (0.25)} & \makecell{3.64 \\ (0.23)} & \makecell{3.25 \\ (0.19)} \\
\addlinespace[0.5em]
gemini-2.5-flash & \makecell{System Prompt \\ (General)} & \makecell{1.15 \\ (0.11)} & \makecell{4.94 \\ (0.06)} & \makecell{5.00 \\ (0.00)} & \makecell{1.20 \\ (0.08)} & \makecell{2.68 \\ (0.21)} & \makecell{1.91 \\ (0.32)} & \makecell{2.67 \\ (0.27)} & \makecell{3.50 \\ (0.19)} \\
\addlinespace[0.5em]
gemini-2.5-flash & \makecell{System Prompt \\ (Task-Dependent)} & \makecell{1.06 \\ (0.06)} & \makecell{4.94 \\ (0.06)} & \makecell{5.00 \\ (0.00)} & \makecell{1.62 \\ (0.15)} & \makecell{4.08 \\ (0.17)} & \makecell{2.96 \\ (0.39)} & \makecell{4.22 \\ (0.17)} & \makecell{4.00 \\ (0.16)} \\
\addlinespace[0.5em]
\midrule
\addlinespace[0.5em]

Llama-3.1-8B-Instruct & \makecell{Temperature \\ (t=0.0)} & \makecell{1.09 \\ (0.05)} & \makecell{1.25 \\ (0.14)} & \makecell{1.71 \\ (0.27)} & \makecell{2.40 \\ (0.21)} & \makecell{2.58 \\ (0.20)} & \makecell{1.48 \\ (0.22)} & \makecell{1.47 \\ (0.15)} & \makecell{1.11 \\ (0.04)} \\
\addlinespace[0.5em]
Llama-3.1-8B-Instruct & \makecell{Temperature \\ (t=0.5)} & \makecell{1.19 \\ (0.09)} & \makecell{1.81 \\ (0.26)} & \makecell{2.79 \\ (0.47)} & \makecell{2.75 \\ (0.20)} & \makecell{3.46 \\ (0.20)} & \makecell{1.70 \\ (0.29)} & \makecell{1.89 \\ (0.21)} & \makecell{1.49 \\ (0.10)} \\
\addlinespace[0.5em]
Llama-3.1-8B-Instruct & \makecell{Temperature \\ (t=1.0)} & \makecell{1.45 \\ (0.13)} & \makecell{2.38 \\ (0.30)} & \makecell{3.71 \\ (0.37)} & \makecell{2.89 \\ (0.21)} & \makecell{4.20 \\ (0.17)} & \makecell{2.17 \\ (0.31)} & \makecell{2.22 \\ (0.22)} & \makecell{1.75 \\ (0.12)} \\
\addlinespace[0.5em]
Llama-3.1-8B-Instruct & \makecell{In-Context Regeneration \\ (General)} & \makecell{2.02 \\ (0.18)} & \makecell{4.94 \\ (0.06)} & \makecell{5.00 \\ (0.00)} & \makecell{2.22 \\ (0.19)} & \makecell{3.00 \\ (0.21)} & \makecell{2.22 \\ (0.35)} & \makecell{2.80 \\ (0.25)} & \makecell{2.91 \\ (0.20)} \\
\addlinespace[0.5em]
Llama-3.1-8B-Instruct & \makecell{In-Context Regeneration \\ (Task-Dependent)} & \makecell{1.55 \\ (0.16)} & \makecell{3.94 \\ (0.44)} & \makecell{4.71 \\ (0.29)} & \makecell{3.11 \\ (0.21)} & \makecell{4.56 \\ (0.11)} & \makecell{3.43 \\ (0.36)} & \makecell{4.16 \\ (0.21)} & \makecell{3.72 \\ (0.17)} \\
\addlinespace[0.5em]
Llama-3.1-8B-Instruct & \makecell{System Prompt \\ (General)} & \makecell{1.59 \\ (0.17)} & \makecell{4.94 \\ (0.06)} & \makecell{5.00 \\ (0.00)} & \makecell{1.92 \\ (0.16)} & \makecell{4.56 \\ (0.12)} & \makecell{2.91 \\ (0.43)} & \makecell{3.11 \\ (0.25)} & \makecell{3.98 \\ (0.15)} \\
\addlinespace[0.5em]
Llama-3.1-8B-Instruct & \makecell{System Prompt \\ (Task-Dependent)} & \makecell{1.20 \\ (0.10)} & \makecell{4.25 \\ (0.40)} & \makecell{5.00 \\ (0.00)} & \makecell{2.75 \\ (0.19)} & \makecell{4.70 \\ (0.10)} & \makecell{4.13 \\ (0.30)} & \makecell{3.79 \\ (0.21)} & \makecell{4.35 \\ (0.12)} \\
\addlinespace[0.5em]
\midrule
\addlinespace[0.5em]
Mistral-7B-Instruct-v0.3 & \makecell{Temperature \\ (t=0.0)} & \makecell{1.81 \\ (0.16)} & \makecell{1.31 \\ (0.20)} & \makecell{1.71 \\ (0.22)} & \makecell{3.15 \\ (0.21)} & \makecell{2.80 \\ (0.20)} & \makecell{1.57 \\ (0.27)} & \makecell{1.53 \\ (0.14)} & \makecell{1.20 \\ (0.06)} \\
\addlinespace[0.5em]
Mistral-7B-Instruct-v0.3 & \makecell{Temperature \\ (t=0.5)} & \makecell{2.83 \\ (0.20)} & \makecell{2.00 \\ (0.27)} & \makecell{3.00 \\ (0.39)} & \makecell{4.05 \\ (0.18)} & \makecell{3.42 \\ (0.21)} & \makecell{2.04 \\ (0.32)} & \makecell{2.00 \\ (0.22)} & \makecell{1.28 \\ (0.07)} \\
\addlinespace[0.5em]
Mistral-7B-Instruct-v0.3 & \makecell{Temperature \\ (t=1.0)} & \makecell{3.68 \\ (0.20)} & \makecell{2.69 \\ (0.31)} & \makecell{3.43 \\ (0.31)} & \makecell{4.44 \\ (0.15)} & \makecell{3.98 \\ (0.17)} & \makecell{2.13 \\ (0.33)} & \makecell{2.51 \\ (0.23)} & \makecell{1.52 \\ (0.11)} \\
\addlinespace[0.5em]
Mistral-7B-Instruct-v0.3 & \makecell{In-Context Regeneration \\ (General)} & \makecell{1.15 \\ (0.08)} & \makecell{4.88 \\ (0.12)} & \makecell{4.79 \\ (0.21)} & \makecell{1.49 \\ (0.10)} & \makecell{1.30 \\ (0.09)} & \makecell{1.48 \\ (0.25)} & \makecell{1.71 \\ (0.19)} & \makecell{2.62 \\ (0.20)} \\
\addlinespace[0.5em]
Mistral-7B-Instruct-v0.3 & \makecell{In-Context Regeneration \\ (Task-Dependent)} & \makecell{1.57 \\ (0.17)} & \makecell{4.31 \\ (0.37)} & \makecell{4.64 \\ (0.23)} & \makecell{1.76 \\ (0.14)} & \makecell{3.00 \\ (0.20)} & \makecell{2.30 \\ (0.30)} & \makecell{2.04 \\ (0.19)} & \makecell{2.68 \\ (0.19)} \\
\addlinespace[0.5em]
Mistral-7B-Instruct-v0.3 & \makecell{System Prompt \\ (General)} & \makecell{3.83 \\ (0.24)} & \makecell{4.88 \\ (0.12)} & \makecell{4.64 \\ (0.29)} & \makecell{2.82 \\ (0.22)} & \makecell{4.52 \\ (0.12)} & \makecell{3.09 \\ (0.40)} & \makecell{2.96 \\ (0.28)} & \makecell{3.92 \\ (0.18)} \\
\addlinespace[0.5em]
Mistral-7B-Instruct-v0.3 & \makecell{System Prompt \\ (Task-Dependent)} & \makecell{3.08 \\ (0.26)} & \makecell{4.94 \\ (0.06)} & \makecell{5.00 \\ (0.00)} & \makecell{3.00 \\ (0.22)} & \makecell{4.88 \\ (0.07)} & \makecell{4.00 \\ (0.35)} & \makecell{3.69 \\ (0.23)} & \makecell{4.42 \\ (0.12)} \\

\bottomrule
\end{tabular}
\end{table}

\clearpage

\begin{table}[h!]
\centering
\caption{\centering Vocabulary Diversity by Model, Sampling Strategy \& Task Category.}
\label{tab:vocab_div}
\tiny
\begin{tabular}{cccccccccc}
\toprule
\makecell{Model} & \makecell{Sampling Strategy} & \makecell{A} & \makecell{B} & \makecell{C} & \makecell{D} & \makecell{E} & \makecell{F} & \makecell{G} & \makecell{H} \\
\midrule
gpt-4o & \makecell{Temperature \\ (t=0.0)} & \makecell{0.29 \\ (0.02)} & \makecell{0.31 \\ (0.05)} & \makecell{0.26 \\ (0.07)} & \makecell{0.41 \\ (0.02)} & \makecell{0.53 \\ (0.01)} & \makecell{0.53 \\ (0.03)} & \makecell{0.51 \\ (0.04)} & \makecell{0.48 \\ (0.02)} \\
\addlinespace[0.5em]
gpt-4o & \makecell{Temperature \\ (t=1.0)} & \makecell{0.49 \\ (0.03)} & \makecell{0.52 \\ (0.06)} & \makecell{0.68 \\ (0.07)} & \makecell{0.63 \\ (0.01)} & \makecell{0.72 \\ (0.00)} & \makecell{0.72 \\ (0.01)} & \makecell{0.75 \\ (0.02)} & \makecell{0.71 \\ (0.01)} \\
\addlinespace[0.5em]
gpt-4o & \makecell{Temperature \\ (t=2.0)} & \makecell{0.54 \\ (0.02)} & \makecell{0.58 \\ (0.04)} & \makecell{0.71 \\ (0.05)} & \makecell{0.66 \\ (0.01)} & \makecell{0.75 \\ (0.00)} & \makecell{0.75 \\ (0.01)} & \makecell{0.79 \\ (0.02)} & \makecell{0.74 \\ (0.01)} \\
\addlinespace[0.5em]
gpt-4o & \makecell{In-Context Regeneration \\ (General)} & \makecell{0.61 \\ (0.03)} & \makecell{0.96 \\ (0.03)} & \makecell{0.95 \\ (0.03)} & \makecell{0.52 \\ (0.02)} & \makecell{0.72 \\ (0.01)} & \makecell{0.75 \\ (0.01)} & \makecell{0.84 \\ (0.02)} & \makecell{0.83 \\ (0.01)} \\
\addlinespace[0.5em]
gpt-4o & \makecell{In-Context Regeneration \\ (Task-Dependent)} & \makecell{0.53 \\ (0.02)} & \makecell{0.96 \\ (0.03)} & \makecell{0.96 \\ (0.03)} & \makecell{0.53 \\ (0.01)} & \makecell{0.72 \\ (0.01)} & \makecell{0.76 \\ (0.02)} & \makecell{0.88 \\ (0.01)} & \makecell{0.86 \\ (0.01)} \\
\addlinespace[0.5em]
gpt-4o & \makecell{System Prompt \\ (General)} & \makecell{0.68 \\ (0.02)} & \makecell{0.78 \\ (0.04)} & \makecell{0.77 \\ (0.08)} & \makecell{0.55 \\ (0.01)} & \makecell{0.78 \\ (0.01)} & \makecell{0.82 \\ (0.01)} & \makecell{0.86 \\ (0.01)} & \makecell{0.85 \\ (0.00)} \\
\addlinespace[0.5em]
gpt-4o & \makecell{System Prompt \\ (Task-Dependent)} & \makecell{0.51 \\ (0.02)} & \makecell{0.81 \\ (0.05)} & \makecell{0.75 \\ (0.08)} & \makecell{0.65 \\ (0.02)} & \makecell{0.79 \\ (0.01)} & \makecell{0.80 \\ (0.03)} & \makecell{0.87 \\ (0.01)} & \makecell{0.86 \\ (0.01)} \\
\addlinespace[0.5em]
\midrule
\addlinespace[0.5em]
claude-4-sonnet & \makecell{Temperature \\ (t=0.0)} & \makecell{0.20 \\ (0.03)} & \makecell{0.07 \\ (0.04)} & \makecell{0.16 \\ (0.06)} & \makecell{0.28 \\ (0.02)} & \makecell{0.53 \\ (0.02)} & \makecell{0.43 \\ (0.04)} & \makecell{0.48 \\ (0.04)} & \makecell{0.42 \\ (0.02)} \\
\addlinespace[0.5em]
claude-4-sonnet & \makecell{Temperature \\ (t=0.5)} & \makecell{0.35 \\ (0.02)} & \makecell{0.17 \\ (0.05)} & \makecell{0.29 \\ (0.07)} & \makecell{0.35 \\ (0.03)} & \makecell{0.63 \\ (0.01)} & \makecell{0.57 \\ (0.04)} & \makecell{0.62 \\ (0.03)} & \makecell{0.56 \\ (0.02)} \\
\addlinespace[0.5em]
claude-4-sonnet & \makecell{Temperature \\ (t=1.0)} & \makecell{0.45 \\ (0.02)} & \makecell{0.23 \\ (0.05)} & \makecell{0.46 \\ (0.07)} & \makecell{0.42 \\ (0.02)} & \makecell{0.68 \\ (0.01)} & \makecell{0.63 \\ (0.04)} & \makecell{0.68 \\ (0.03)} & \makecell{0.63 \\ (0.02)} \\
\addlinespace[0.5em]
claude-4-sonnet & \makecell{In-Context Regeneration \\ (General)} & \makecell{0.74 \\ (0.01)} & \makecell{0.78 \\ (0.05)} & \makecell{0.92 \\ (0.02)} & \makecell{0.73 \\ (0.01)} & \makecell{0.78 \\ (0.00)} & \makecell{0.83 \\ (0.01)} & \makecell{0.87 \\ (0.01)} & \makecell{0.86 \\ (0.01)} \\
\addlinespace[0.5em]
claude-4-sonnet & \makecell{In-Context Regeneration \\ (Task-Dependent)} & \makecell{0.62 \\ (0.01)} & \makecell{0.75 \\ (0.06)} & \makecell{0.87 \\ (0.06)} & \makecell{0.76 \\ (0.01)} & \makecell{0.79 \\ (0.01)} & \makecell{0.82 \\ (0.01)} & \makecell{0.92 \\ (0.01)} & \makecell{0.87 \\ (0.01)} \\
\addlinespace[0.5em]
claude-4-sonnet & \makecell{System Prompt \\ (General)} & \makecell{0.71 \\ (0.01)} & \makecell{0.81 \\ (0.01)} & \makecell{0.81 \\ (0.04)} & \makecell{0.70 \\ (0.01)} & \makecell{0.78 \\ (0.01)} & \makecell{0.80 \\ (0.01)} & \makecell{0.86 \\ (0.01)} & \makecell{0.85 \\ (0.00)} \\
\addlinespace[0.5em]
claude-4-sonnet & \makecell{System Prompt \\ (Task-Dependent)} & \makecell{0.68 \\ (0.01)} & \makecell{0.81 \\ (0.01)} & \makecell{0.81 \\ (0.06)} & \makecell{0.75 \\ (0.01)} & \makecell{0.81 \\ (0.00)} & \makecell{0.82 \\ (0.01)} & \makecell{0.90 \\ (0.01)} & \makecell{0.86 \\ (0.00)} \\
\addlinespace[0.5em]
\midrule
\addlinespace[0.5em]
gemini-2.5-flash & \makecell{Temperature \\ (t=0.0)} & \makecell{0.07 \\ (0.02)} & \makecell{0.12 \\ (0.05)} & \makecell{0.25 \\ (0.07)} & \makecell{0.20 \\ (0.02)} & \makecell{0.35 \\ (0.02)} & \makecell{0.34 \\ (0.03)} & \makecell{0.31 \\ (0.03)} & \makecell{0.29 \\ (0.02)} \\
\addlinespace[0.5em]
gemini-2.5-flash & \makecell{Temperature \\ (t=1.0)} & \makecell{0.38 \\ (0.04)} & \makecell{0.44 \\ (0.08)} & \makecell{0.70 \\ (0.07)} & \makecell{0.52 \\ (0.02)} & \makecell{0.76 \\ (0.00)} & \makecell{0.74 \\ (0.02)} & \makecell{0.76 \\ (0.03)} & \makecell{0.71 \\ (0.02)} \\
\addlinespace[0.5em]
gemini-2.5-flash & \makecell{Temperature \\ (t=2.0)} & \makecell{0.44 \\ (0.04)} & \makecell{0.44 \\ (0.07)} & \makecell{0.77 \\ (0.04)} & \makecell{0.56 \\ (0.02)} & \makecell{0.77 \\ (0.00)} & \makecell{0.78 \\ (0.01)} & \makecell{0.79 \\ (0.03)} & \makecell{0.77 \\ (0.01)} \\
\addlinespace[0.5em]
gemini-2.5-flash & \makecell{In-Context Regeneration \\ (General)} & \makecell{0.76 \\ (0.02)} & \makecell{0.98 \\ (0.02)} & \makecell{0.89 \\ (0.04)} & \makecell{0.63 \\ (0.01)} & \makecell{0.69 \\ (0.01)} & \makecell{0.74 \\ (0.02)} & \makecell{0.80 \\ (0.03)} & \makecell{0.83 \\ (0.01)} \\
\addlinespace[0.5em]
gemini-2.5-flash & \makecell{In-Context Regeneration \\ (Task-Dependent)} & \makecell{0.60 \\ (0.02)} & \makecell{0.95 \\ (0.03)} & \makecell{0.91 \\ (0.04)} & \makecell{0.69 \\ (0.01)} & \makecell{0.71 \\ (0.01)} & \makecell{0.77 \\ (0.02)} & \makecell{0.89 \\ (0.01)} & \makecell{0.87 \\ (0.01)} \\
\addlinespace[0.5em]
gemini-2.5-flash & \makecell{System Prompt \\ (General)} & \makecell{0.67 \\ (0.01)} & \makecell{0.85 \\ (0.05)} & \makecell{0.86 \\ (0.03)} & \makecell{0.66 \\ (0.01)} & \makecell{0.74 \\ (0.01)} & \makecell{0.81 \\ (0.01)} & \makecell{0.85 \\ (0.01)} & \makecell{0.85 \\ (0.00)} \\
\addlinespace[0.5em]
gemini-2.5-flash & \makecell{System Prompt \\ (Task-Dependent)} & \makecell{0.57 \\ (0.02)} & \makecell{0.92 \\ (0.05)} & \makecell{0.73 \\ (0.07)} & \makecell{0.71 \\ (0.01)} & \makecell{0.77 \\ (0.01)} & \makecell{0.83 \\ (0.01)} & \makecell{0.89 \\ (0.01)} & \makecell{0.87 \\ (0.00)} \\
\addlinespace[0.5em]
\midrule
\addlinespace[0.5em]
Llama-3.1-8B-Instruct & \makecell{Temperature \\ (t=0.0)} & \makecell{0.15 \\ (0.03)} & \makecell{0.07 \\ (0.02)} & \makecell{0.24 \\ (0.06)} & \makecell{0.45 \\ (0.03)} & \makecell{0.54 \\ (0.01)} & \makecell{0.47 \\ (0.04)} & \makecell{0.46 \\ (0.04)} & \makecell{0.47 \\ (0.02)} \\
\addlinespace[0.5em]
Llama-3.1-8B-Instruct & \makecell{Temperature \\ (t=0.5)} & \makecell{0.43 \\ (0.03)} & \makecell{0.32 \\ (0.05)} & \makecell{0.48 \\ (0.09)} & \makecell{0.64 \\ (0.02)} & \makecell{0.66 \\ (0.01)} & \makecell{0.63 \\ (0.02)} & \makecell{0.65 \\ (0.03)} & \makecell{0.67 \\ (0.01)} \\
\addlinespace[0.5em]
Llama-3.1-8B-Instruct & \makecell{Temperature \\ (t=1.0)} & \makecell{0.64 \\ (0.02)} & \makecell{0.57 \\ (0.04)} & \makecell{0.74 \\ (0.05)} & \makecell{0.72 \\ (0.01)} & \makecell{0.74 \\ (0.00)} & \makecell{0.74 \\ (0.02)} & \makecell{0.77 \\ (0.02)} & \makecell{0.75 \\ (0.01)} \\
\addlinespace[0.5em]
Llama-3.1-8B-Instruct & \makecell{In-Context Regeneration \\ (General)} & \makecell{0.58 \\ (0.02)} & \makecell{0.86 \\ (0.05)} & \makecell{0.94 \\ (0.03)} & \makecell{0.53 \\ (0.02)} & \makecell{0.52 \\ (0.02)} & \makecell{0.64 \\ (0.04)} & \makecell{0.73 \\ (0.03)} & \makecell{0.68 \\ (0.02)} \\
\addlinespace[0.5em]
Llama-3.1-8B-Instruct & \makecell{In-Context Regeneration \\ (Task-Dependent)} & \makecell{0.52 \\ (0.02)} & \makecell{0.89 \\ (0.04)} & \makecell{0.87 \\ (0.07)} & \makecell{0.60 \\ (0.01)} & \makecell{0.66 \\ (0.01)} & \makecell{0.73 \\ (0.03)} & \makecell{0.84 \\ (0.02)} & \makecell{0.80 \\ (0.01)} \\
\addlinespace[0.5em]
Llama-3.1-8B-Instruct & \makecell{System Prompt \\ (General)} & \makecell{0.65 \\ (0.02)} & \makecell{0.73 \\ (0.03)} & \makecell{0.72 \\ (0.04)} & \makecell{0.42 \\ (0.02)} & \makecell{0.64 \\ (0.02)} & \makecell{0.66 \\ (0.04)} & \makecell{0.71 \\ (0.03)} & \makecell{0.76 \\ (0.01)} \\
\addlinespace[0.5em]
Llama-3.1-8B-Instruct & \makecell{System Prompt \\ (Task-Dependent)} & \makecell{0.58 \\ (0.02)} & \makecell{0.66 \\ (0.03)} & \makecell{0.49 \\ (0.06)} & \makecell{0.57 \\ (0.03)} & \makecell{0.68 \\ (0.01)} & \makecell{0.72 \\ (0.03)} & \makecell{0.76 \\ (0.02)} & \makecell{0.76 \\ (0.01)} \\
\addlinespace[0.5em]
\midrule
\addlinespace[0.5em]
Mistral-7B-Instruct-v0.3 & \makecell{Temperature \\ (t=0.0)} & \makecell{0.33 \\ (0.03)} & \makecell{0.26 \\ (0.06)} & \makecell{0.38 \\ (0.06)} & \makecell{0.49 \\ (0.03)} & \makecell{0.55 \\ (0.01)} & \makecell{0.50 \\ (0.04)} & \makecell{0.53 \\ (0.04)} & \makecell{0.46 \\ (0.02)} \\
\addlinespace[0.5em]
Mistral-7B-Instruct-v0.3 & \makecell{Temperature \\ (t=0.5)} & \makecell{0.60 \\ (0.02)} & \makecell{0.60 \\ (0.03)} & \makecell{0.74 \\ (0.02)} & \makecell{0.70 \\ (0.01)} & \makecell{0.67 \\ (0.01)} & \makecell{0.67 \\ (0.02)} & \makecell{0.72 \\ (0.02)} & \makecell{0.67 \\ (0.01)} \\
\addlinespace[0.5em]
Mistral-7B-Instruct-v0.3 & \makecell{Temperature \\ (t=1.0)} & \makecell{0.72 \\ (0.01)} & \makecell{0.75 \\ (0.01)} & \makecell{0.82 \\ (0.02)} & \makecell{0.76 \\ (0.01)} & \makecell{0.73 \\ (0.01)} & \makecell{0.75 \\ (0.01)} & \makecell{0.80 \\ (0.01)} & \makecell{0.75 \\ (0.01)} \\
\addlinespace[0.5em]
Mistral-7B-Instruct-v0.3 & \makecell{In-Context Regeneration \\ (General)} & \makecell{0.46 \\ (0.02)} & \makecell{0.73 \\ (0.02)} & \makecell{0.63 \\ (0.06)} & \makecell{0.27 \\ (0.03)} & \makecell{0.28 \\ (0.02)} & \makecell{0.43 \\ (0.04)} & \makecell{0.51 \\ (0.03)} & \makecell{0.53 \\ (0.02)} \\
\addlinespace[0.5em]
Mistral-7B-Instruct-v0.3 & \makecell{In-Context Regeneration \\ (Task-Dependent)} & \makecell{0.42 \\ (0.02)} & \makecell{0.59 \\ (0.04)} & \makecell{0.47 \\ (0.08)} & \makecell{0.31 \\ (0.03)} & \makecell{0.38 \\ (0.02)} & \makecell{0.51 \\ (0.04)} & \makecell{0.63 \\ (0.03)} & \makecell{0.59 \\ (0.02)} \\
\addlinespace[0.5em]
Mistral-7B-Instruct-v0.3 & \makecell{System Prompt \\ (General)} & \makecell{0.64 \\ (0.02)} & \makecell{0.80 \\ (0.02)} & \makecell{0.81 \\ (0.03)} & \makecell{0.45 \\ (0.03)} & \makecell{0.72 \\ (0.02)} & \makecell{0.74 \\ (0.02)} & \makecell{0.72 \\ (0.03)} & \makecell{0.80 \\ (0.01)} \\
\addlinespace[0.5em]
Mistral-7B-Instruct-v0.3 & \makecell{System Prompt \\ (Task-Dependent)} & \makecell{0.63 \\ (0.02)} & \makecell{0.77 \\ (0.02)} & \makecell{0.79 \\ (0.02)} & \makecell{0.52 \\ (0.03)} & \makecell{0.76 \\ (0.02)} & \makecell{0.81 \\ (0.01)} & \makecell{0.78 \\ (0.02)} & \makecell{0.84 \\ (0.01)} \\
\bottomrule
\end{tabular}
\end{table}

\clearpage

\begin{table}[h!]
\centering
\caption{\centering Embedding Diversity by Model, Sampling Strategy \& Task Category.}
\label{tab:embed_div}
\tiny
\begin{tabular}{cccccccccc}
\toprule
\makecell{Model} & \makecell{Sampling Strategy} & \makecell{A} & \makecell{B} & \makecell{C} & \makecell{D} & \makecell{E} & \makecell{F} & \makecell{G} & \makecell{H} \\
\midrule
gpt-4o & \makecell{Temperature \\ (t=0.0)} & \makecell{0.02 \\ (0.00)} & \makecell{0.03 \\ (0.01)} & \makecell{0.02 \\ (0.01)} & \makecell{0.01 \\ (0.00)} & \makecell{0.01 \\ (0.00)} & \makecell{0.02 \\ (0.00)} & \makecell{0.03 \\ (0.00)} & \makecell{0.02 \\ (0.00)} \\
\addlinespace[0.5em]
gpt-4o & \makecell{Temperature \\ (t=1.0)} & \makecell{0.04 \\ (0.00)} & \makecell{0.04 \\ (0.01)} & \makecell{0.07 \\ (0.01)} & \makecell{0.03 \\ (0.00)} & \makecell{0.03 \\ (0.00)} & \makecell{0.04 \\ (0.00)} & \makecell{0.06 \\ (0.01)} & \makecell{0.04 \\ (0.00)} \\
\addlinespace[0.5em]
gpt-4o & \makecell{Temperature \\ (t=2.0)} & \makecell{0.05 \\ (0.00)} & \makecell{0.06 \\ (0.01)} & \makecell{0.07 \\ (0.02)} & \makecell{0.03 \\ (0.00)} & \makecell{0.03 \\ (0.00)} & \makecell{0.04 \\ (0.00)} & \makecell{0.07 \\ (0.01)} & \makecell{0.04 \\ (0.00)} \\
\addlinespace[0.5em]
gpt-4o & \makecell{In-Context Regeneration \\ (General)} & \makecell{0.07 \\ (0.01)} & \makecell{0.15 \\ (0.01)} & \makecell{0.16 \\ (0.01)} & \makecell{0.04 \\ (0.00)} & \makecell{0.04 \\ (0.00)} & \makecell{0.05 \\ (0.01)} & \makecell{0.10 \\ (0.01)} & \makecell{0.09 \\ (0.01)} \\
\addlinespace[0.5em]
gpt-4o & \makecell{In-Context Regeneration \\ (Task-Dependent)} & \makecell{0.03 \\ (0.00)} & \makecell{0.15 \\ (0.01)} & \makecell{0.16 \\ (0.01)} & \makecell{0.03 \\ (0.00)} & \makecell{0.05 \\ (0.00)} & \makecell{0.06 \\ (0.01)} & \makecell{0.13 \\ (0.01)} & \makecell{0.11 \\ (0.01)} \\
\addlinespace[0.5em]
gpt-4o & \makecell{System Prompt \\ (General)} & \makecell{0.05 \\ (0.00)} & \makecell{0.14 \\ (0.01)} & \makecell{0.13 \\ (0.01)} & \makecell{0.03 \\ (0.00)} & \makecell{0.06 \\ (0.00)} & \makecell{0.08 \\ (0.01)} & \makecell{0.12 \\ (0.01)} & \makecell{0.11 \\ (0.00)} \\
\addlinespace[0.5em]
gpt-4o & \makecell{System Prompt \\ (Task-Dependent)} & \makecell{0.02 \\ (0.00)} & \makecell{0.13 \\ (0.01)} & \makecell{0.12 \\ (0.01)} & \makecell{0.06 \\ (0.00)} & \makecell{0.08 \\ (0.00)} & \makecell{0.10 \\ (0.01)} & \makecell{0.14 \\ (0.01)} & \makecell{0.12 \\ (0.00)} \\
\addlinespace[0.5em]
\midrule
\addlinespace[0.5em]
claude-4-sonnet & \makecell{Temperature \\ (t=0.0)} & \makecell{0.01 \\ (0.00)} & \makecell{0.01 \\ (0.00)} & \makecell{0.01 \\ (0.00)} & \makecell{0.01 \\ (0.00)} & \makecell{0.01 \\ (0.00)} & \makecell{0.01 \\ (0.00)} & \makecell{0.02 \\ (0.00)} & \makecell{0.01 \\ (0.00)} \\
\addlinespace[0.5em]
claude-4-sonnet & \makecell{Temperature \\ (t=0.5)} & \makecell{0.02 \\ (0.00)} & \makecell{0.01 \\ (0.00)} & \makecell{0.02 \\ (0.00)} & \makecell{0.01 \\ (0.00)} & \makecell{0.02 \\ (0.00)} & \makecell{0.02 \\ (0.00)} & \makecell{0.04 \\ (0.00)} & \makecell{0.02 \\ (0.00)} \\
\addlinespace[0.5em]
claude-4-sonnet & \makecell{Temperature \\ (t=1.0)} & \makecell{0.03 \\ (0.00)} & \makecell{0.02 \\ (0.00)} & \makecell{0.05 \\ (0.01)} & \makecell{0.01 \\ (0.00)} & \makecell{0.02 \\ (0.00)} & \makecell{0.02 \\ (0.00)} & \makecell{0.05 \\ (0.00)} & \makecell{0.03 \\ (0.00)} \\
\addlinespace[0.5em]
claude-4-sonnet & \makecell{In-Context Regeneration \\ (General)} & \makecell{0.07 \\ (0.00)} & \makecell{0.13 \\ (0.01)} & \makecell{0.14 \\ (0.01)} & \makecell{0.04 \\ (0.00)} & \makecell{0.06 \\ (0.00)} & \makecell{0.07 \\ (0.01)} & \makecell{0.11 \\ (0.01)} & \makecell{0.10 \\ (0.00)} \\
\addlinespace[0.5em]
claude-4-sonnet & \makecell{In-Context Regeneration \\ (Task-Dependent)} & \makecell{0.03 \\ (0.00)} & \makecell{0.12 \\ (0.01)} & \makecell{0.14 \\ (0.01)} & \makecell{0.06 \\ (0.00)} & \makecell{0.08 \\ (0.00)} & \makecell{0.09 \\ (0.01)} & \makecell{0.16 \\ (0.01)} & \makecell{0.13 \\ (0.00)} \\
\addlinespace[0.5em]
claude-4-sonnet & \makecell{System Prompt \\ (General)} & \makecell{0.05 \\ (0.00)} & \makecell{0.16 \\ (0.01)} & \makecell{0.12 \\ (0.01)} & \makecell{0.04 \\ (0.00)} & \makecell{0.07 \\ (0.00)} & \makecell{0.07 \\ (0.01)} & \makecell{0.12 \\ (0.01)} & \makecell{0.10 \\ (0.00)} \\
\addlinespace[0.5em]
claude-4-sonnet & \makecell{System Prompt \\ (Task-Dependent)} & \makecell{0.04 \\ (0.00)} & \makecell{0.16 \\ (0.01)} & \makecell{0.11 \\ (0.01)} & \makecell{0.06 \\ (0.00)} & \makecell{0.08 \\ (0.00)} & \makecell{0.08 \\ (0.01)} & \makecell{0.14 \\ (0.01)} & \makecell{0.12 \\ (0.00)} \\
\addlinespace[0.5em]
\midrule
\addlinespace[0.5em]
gemini-2.5-flash & \makecell{Temperature \\ (t=0.0)} & \makecell{0.01 \\ (0.00)} & \makecell{0.01 \\ (0.01)} & \makecell{0.02 \\ (0.01)} & \makecell{0.01 \\ (0.00)} & \makecell{0.01 \\ (0.00)} & \makecell{0.02 \\ (0.00)} & \makecell{0.02 \\ (0.00)} & \makecell{0.01 \\ (0.00)} \\
\addlinespace[0.5em]
gemini-2.5-flash & \makecell{Temperature \\ (t=1.0)} & \makecell{0.04 \\ (0.01)} & \makecell{0.04 \\ (0.01)} & \makecell{0.09 \\ (0.02)} & \makecell{0.02 \\ (0.00)} & \makecell{0.03 \\ (0.00)} & \makecell{0.04 \\ (0.00)} & \makecell{0.07 \\ (0.01)} & \makecell{0.04 \\ (0.00)} \\
\addlinespace[0.5em]
gemini-2.5-flash & \makecell{Temperature \\ (t=2.0)} & \makecell{0.05 \\ (0.01)} & \makecell{0.05 \\ (0.01)} & \makecell{0.10 \\ (0.01)} & \makecell{0.02 \\ (0.00)} & \makecell{0.04 \\ (0.00)} & \makecell{0.05 \\ (0.01)} & \makecell{0.07 \\ (0.01)} & \makecell{0.05 \\ (0.00)} \\
\addlinespace[0.5em]
gemini-2.5-flash & \makecell{In-Context Regeneration \\ (General)} & \makecell{0.08 \\ (0.00)} & \makecell{0.14 \\ (0.01)} & \makecell{0.15 \\ (0.01)} & \makecell{0.03 \\ (0.00)} & \makecell{0.04 \\ (0.00)} & \makecell{0.05 \\ (0.01)} & \makecell{0.10 \\ (0.01)} & \makecell{0.09 \\ (0.01)} \\
\addlinespace[0.5em]
gemini-2.5-flash & \makecell{In-Context Regeneration \\ (Task-Dependent)} & \makecell{0.04 \\ (0.00)} & \makecell{0.14 \\ (0.01)} & \makecell{0.15 \\ (0.01)} & \makecell{0.04 \\ (0.00)} & \makecell{0.04 \\ (0.00)} & \makecell{0.06 \\ (0.01)} & \makecell{0.13 \\ (0.01)} & \makecell{0.10 \\ (0.00)} \\
\addlinespace[0.5em]
gemini-2.5-flash & \makecell{System Prompt \\ (General)} & \makecell{0.04 \\ (0.00)} & \makecell{0.13 \\ (0.01)} & \makecell{0.13 \\ (0.01)} & \makecell{0.03 \\ (0.00)} & \makecell{0.05 \\ (0.00)} & \makecell{0.07 \\ (0.01)} & \makecell{0.11 \\ (0.01)} & \makecell{0.10 \\ (0.00)} \\
\addlinespace[0.5em]
gemini-2.5-flash & \makecell{System Prompt \\ (Task-Dependent)} & \makecell{0.02 \\ (0.00)} & \makecell{0.13 \\ (0.01)} & \makecell{0.12 \\ (0.01)} & \makecell{0.05 \\ (0.00)} & \makecell{0.07 \\ (0.00)} & \makecell{0.10 \\ (0.01)} & \makecell{0.14 \\ (0.01)} & \makecell{0.11 \\ (0.00)} \\
\addlinespace[0.5em]
\midrule
\addlinespace[0.5em]
Llama-3.1-8B-Instruct & \makecell{Temperature \\ (t=0.0)} & \makecell{0.01 \\ (0.00)} & \makecell{0.01 \\ (0.00)} & \makecell{0.03 \\ (0.01)} & \makecell{0.02 \\ (0.00)} & \makecell{0.02 \\ (0.00)} & \makecell{0.02 \\ (0.00)} & \makecell{0.03 \\ (0.00)} & \makecell{0.02 \\ (0.00)} \\
\addlinespace[0.5em]
Llama-3.1-8B-Instruct & \makecell{Temperature \\ (t=0.5)} & \makecell{0.03 \\ (0.00)} & \makecell{0.04 \\ (0.01)} & \makecell{0.05 \\ (0.01)} & \makecell{0.04 \\ (0.00)} & \makecell{0.03 \\ (0.00)} & \makecell{0.03 \\ (0.00)} & \makecell{0.06 \\ (0.01)} & \makecell{0.04 \\ (0.00)} \\
\addlinespace[0.5em]
Llama-3.1-8B-Instruct & \makecell{Temperature \\ (t=1.0)} & \makecell{0.06 \\ (0.00)} & \makecell{0.08 \\ (0.01)} & \makecell{0.08 \\ (0.01)} & \makecell{0.04 \\ (0.00)} & \makecell{0.04 \\ (0.00)} & \makecell{0.05 \\ (0.01)} & \makecell{0.08 \\ (0.01)} & \makecell{0.06 \\ (0.00)} \\
\addlinespace[0.5em]
Llama-3.1-8B-Instruct & \makecell{In-Context Regeneration \\ (General)} & \makecell{0.07 \\ (0.00)} & \makecell{0.14 \\ (0.01)} & \makecell{0.16 \\ (0.01)} & \makecell{0.04 \\ (0.00)} & \makecell{0.04 \\ (0.00)} & \makecell{0.07 \\ (0.01)} & \makecell{0.11 \\ (0.01)} & \makecell{0.09 \\ (0.01)} \\
\addlinespace[0.5em]
Llama-3.1-8B-Instruct & \makecell{In-Context Regeneration \\ (Task-Dependent)} & \makecell{0.05 \\ (0.01)} & \makecell{0.12 \\ (0.01)} & \makecell{0.15 \\ (0.01)} & \makecell{0.05 \\ (0.00)} & \makecell{0.06 \\ (0.00)} & \makecell{0.09 \\ (0.01)} & \makecell{0.14 \\ (0.01)} & \makecell{0.11 \\ (0.00)} \\
\addlinespace[0.5em]
Llama-3.1-8B-Instruct & \makecell{System Prompt \\ (General)} & \makecell{0.07 \\ (0.00)} & \makecell{0.14 \\ (0.01)} & \makecell{0.11 \\ (0.01)} & \makecell{0.03 \\ (0.00)} & \makecell{0.07 \\ (0.00)} & \makecell{0.08 \\ (0.01)} & \makecell{0.10 \\ (0.01)} & \makecell{0.10 \\ (0.00)} \\
\addlinespace[0.5em]
Llama-3.1-8B-Instruct & \makecell{System Prompt \\ (Task-Dependent)} & \makecell{0.04 \\ (0.00)} & \makecell{0.12 \\ (0.01)} & \makecell{0.10 \\ (0.01)} & \makecell{0.06 \\ (0.00)} & \makecell{0.08 \\ (0.00)} & \makecell{0.11 \\ (0.01)} & \makecell{0.12 \\ (0.01)} & \makecell{0.11 \\ (0.00)} \\
\addlinespace[0.5em]
\midrule
\addlinespace[0.5em]
Mistral-7B-Instruct-v0.3 & \makecell{Temperature \\ (t=0.0)} & \makecell{0.03 \\ (0.00)} & \makecell{0.03 \\ (0.01)} & \makecell{0.03 \\ (0.01)} & \makecell{0.03 \\ (0.00)} & \makecell{0.02 \\ (0.00)} & \makecell{0.02 \\ (0.00)} & \makecell{0.03 \\ (0.00)} & \makecell{0.02 \\ (0.00)} \\
\addlinespace[0.5em]
Mistral-7B-Instruct-v0.3 & \makecell{Temperature \\ (t=0.5)} & \makecell{0.06 \\ (0.00)} & \makecell{0.07 \\ (0.01)} & \makecell{0.08 \\ (0.01)} & \makecell{0.05 \\ (0.00)} & \makecell{0.04 \\ (0.00)} & \makecell{0.04 \\ (0.00)} & \makecell{0.06 \\ (0.01)} & \makecell{0.04 \\ (0.00)} \\
\addlinespace[0.5em]
Mistral-7B-Instruct-v0.3 & \makecell{Temperature \\ (t=1.0)} & \makecell{0.08 \\ (0.00)} & \makecell{0.10 \\ (0.01)} & \makecell{0.10 \\ (0.01)} & \makecell{0.06 \\ (0.00)} & \makecell{0.04 \\ (0.00)} & \makecell{0.05 \\ (0.00)} & \makecell{0.08 \\ (0.01)} & \makecell{0.05 \\ (0.00)} \\
\addlinespace[0.5em]
Mistral-7B-Instruct-v0.3 & \makecell{In-Context Regeneration \\ (General)} & \makecell{0.04 \\ (0.00)} & \makecell{0.14 \\ (0.01)} & \makecell{0.11 \\ (0.01)} & \makecell{0.02 \\ (0.00)} & \makecell{0.02 \\ (0.00)} & \makecell{0.04 \\ (0.01)} & \makecell{0.06 \\ (0.01)} & \makecell{0.06 \\ (0.00)} \\
\addlinespace[0.5em]
Mistral-7B-Instruct-v0.3 & \makecell{In-Context Regeneration \\ (Task-Dependent)} & \makecell{0.03 \\ (0.00)} & \makecell{0.11 \\ (0.01)} & \makecell{0.08 \\ (0.01)} & \makecell{0.02 \\ (0.00)} & \makecell{0.03 \\ (0.00)} & \makecell{0.06 \\ (0.01)} & \makecell{0.09 \\ (0.01)} & \makecell{0.07 \\ (0.00)} \\
\addlinespace[0.5em]
Mistral-7B-Instruct-v0.3 & \makecell{System Prompt \\ (General)} & \makecell{0.07 \\ (0.00)} & \makecell{0.16 \\ (0.01)} & \makecell{0.13 \\ (0.01)} & \makecell{0.04 \\ (0.00)} & \makecell{0.08 \\ (0.00)} & \makecell{0.09 \\ (0.01)} & \makecell{0.11 \\ (0.01)} & \makecell{0.11 \\ (0.00)} \\
\addlinespace[0.5em]
Mistral-7B-Instruct-v0.3 & \makecell{System Prompt \\ (Task-Dependent)} & \makecell{0.06 \\ (0.01)} & \makecell{0.16 \\ (0.01)} & \makecell{0.14 \\ (0.01)} & \makecell{0.04 \\ (0.00)} & \makecell{0.10 \\ (0.00)} & \makecell{0.12 \\ (0.01)} & \makecell{0.12 \\ (0.01)} & \makecell{0.14 \\ (0.00)} \\
\bottomrule
\end{tabular}
\end{table}

\clearpage

\begin{table}[h!]
\centering
\caption{\centering Compression Diversity by Model, Sampling Strategy \& Task Category.}
\label{tab:compression_div}
\tiny
\begin{tabular}{cccccccccc}
\toprule
\makecell{Model} & \makecell{Sampling Strategy} & \makecell{A} & \makecell{B} & \makecell{C} & \makecell{D} & \makecell{E} & \makecell{F} & \makecell{G} & \makecell{H} \\
\midrule
gpt-4o & \makecell{Temperature \\ (t=0.0)} & \makecell{0.52 \\ (0.01)} & \makecell{0.61 \\ (0.05)} & \makecell{0.90 \\ (0.19)} & \makecell{0.29 \\ (0.01)} & \makecell{0.36 \\ (0.00)} & \makecell{0.38 \\ (0.01)} & \makecell{0.49 \\ (0.02)} & \makecell{0.43 \\ (0.02)} \\
\addlinespace[0.5em]
gpt-4o & \makecell{Temperature \\ (t=1.0)} & \makecell{0.57 \\ (0.01)} & \makecell{0.69 \\ (0.07)} & \makecell{0.94 \\ (0.14)} & \makecell{0.34 \\ (0.01)} & \makecell{0.42 \\ (0.00)} & \makecell{0.44 \\ (0.01)} & \makecell{0.55 \\ (0.02)} & \makecell{0.49 \\ (0.02)} \\
\addlinespace[0.5em]
gpt-4o & \makecell{Temperature \\ (t=2.0)} & \makecell{0.58 \\ (0.01)} & \makecell{0.74 \\ (0.07)} & \makecell{0.91 \\ (0.12)} & \makecell{0.36 \\ (0.01)} & \makecell{0.43 \\ (0.00)} & \makecell{0.45 \\ (0.01)} & \makecell{0.57 \\ (0.03)} & \makecell{0.50 \\ (0.02)} \\
\addlinespace[0.5em]
gpt-4o & \makecell{In-Context Regeneration \\ (General)} & \makecell{0.96 \\ (0.07)} & \makecell{1.51 \\ (0.12)} & \makecell{2.34 \\ (0.30)} & \makecell{0.42 \\ (0.03)} & \makecell{0.45 \\ (0.00)} & \makecell{0.49 \\ (0.02)} & \makecell{0.69 \\ (0.06)} & \makecell{0.73 \\ (0.04)} \\
\addlinespace[0.5em]
gpt-4o & \makecell{In-Context Regeneration \\ (Task-Dependent)} & \makecell{0.70 \\ (0.02)} & \makecell{1.55 \\ (0.12)} & \makecell{2.29 \\ (0.31)} & \makecell{0.37 \\ (0.02)} & \makecell{0.44 \\ (0.01)} & \makecell{0.54 \\ (0.06)} & \makecell{0.70 \\ (0.05)} & \makecell{0.73 \\ (0.04)} \\
\addlinespace[0.5em]
gpt-4o & \makecell{System Prompt \\ (General)} & \makecell{0.66 \\ (0.01)} & \makecell{0.97 \\ (0.12)} & \makecell{1.89 \\ (0.52)} & \makecell{0.41 \\ (0.01)} & \makecell{0.51 \\ (0.00)} & \makecell{0.55 \\ (0.01)} & \makecell{0.67 \\ (0.04)} & \makecell{0.63 \\ (0.02)} \\
\addlinespace[0.5em]
gpt-4o & \makecell{System Prompt \\ (Task-Dependent)} & \makecell{0.72 \\ (0.02)} & \makecell{1.37 \\ (0.18)} & \makecell{2.09 \\ (0.34)} & \makecell{0.45 \\ (0.02)} & \makecell{0.51 \\ (0.00)} & \makecell{0.58 \\ (0.01)} & \makecell{0.61 \\ (0.02)} & \makecell{0.63 \\ (0.02)} \\
\addlinespace[0.5em]
\midrule
\addlinespace[0.5em]
claude-4-sonnet & \makecell{Temperature \\ (t=0.0)} & \makecell{0.38 \\ (0.01)} & \makecell{0.83 \\ (0.15)} & \makecell{1.29 \\ (0.53)} & \makecell{0.28 \\ (0.01)} & \makecell{0.38 \\ (0.00)} & \makecell{0.36 \\ (0.01)} & \makecell{0.44 \\ (0.01)} & \makecell{0.43 \\ (0.02)} \\
\addlinespace[0.5em]
claude-4-sonnet & \makecell{Temperature \\ (t=0.5)} & \makecell{0.42 \\ (0.01)} & \makecell{0.81 \\ (0.13)} & \makecell{1.32 \\ (0.52)} & \makecell{0.29 \\ (0.01)} & \makecell{0.41 \\ (0.00)} & \makecell{0.40 \\ (0.01)} & \makecell{0.48 \\ (0.01)} & \makecell{0.46 \\ (0.02)} \\
\addlinespace[0.5em]
claude-4-sonnet & \makecell{Temperature \\ (t=1.0)} & \makecell{0.44 \\ (0.01)} & \makecell{0.81 \\ (0.13)} & \makecell{1.36 \\ (0.52)} & \makecell{0.30 \\ (0.01)} & \makecell{0.42 \\ (0.00)} & \makecell{0.41 \\ (0.01)} & \makecell{0.50 \\ (0.02)} & \makecell{0.48 \\ (0.02)} \\
\addlinespace[0.5em]
claude-4-sonnet & \makecell{In-Context Regeneration \\ (General)} & \makecell{0.55 \\ (0.01)} & \makecell{1.08 \\ (0.15)} & \makecell{1.72 \\ (0.51)} & \makecell{0.39 \\ (0.01)} & \makecell{0.45 \\ (0.00)} & \makecell{0.48 \\ (0.01)} & \makecell{0.57 \\ (0.02)} & \makecell{0.58 \\ (0.03)} \\
\addlinespace[0.5em]
claude-4-sonnet & \makecell{In-Context Regeneration \\ (Task-Dependent)} & \makecell{0.51 \\ (0.01)} & \makecell{1.11 \\ (0.14)} & \makecell{1.93 \\ (0.50)} & \makecell{0.38 \\ (0.01)} & \makecell{0.45 \\ (0.00)} & \makecell{0.46 \\ (0.01)} & \makecell{0.58 \\ (0.02)} & \makecell{0.59 \\ (0.03)} \\
\addlinespace[0.5em]
claude-4-sonnet & \makecell{System Prompt \\ (General)} & \makecell{0.51 \\ (0.00)} & \makecell{0.60 \\ (0.02)} & \makecell{1.14 \\ (0.50)} & \makecell{0.42 \\ (0.01)} & \makecell{0.48 \\ (0.00)} & \makecell{0.50 \\ (0.01)} & \makecell{0.59 \\ (0.02)} & \makecell{0.55 \\ (0.01)} \\
\addlinespace[0.5em]
claude-4-sonnet & \makecell{System Prompt \\ (Task-Dependent)} & \makecell{0.55 \\ (0.01)} & \makecell{0.59 \\ (0.01)} & \makecell{1.51 \\ (0.53)} & \makecell{0.42 \\ (0.01)} & \makecell{0.49 \\ (0.00)} & \makecell{0.51 \\ (0.01)} & \makecell{0.60 \\ (0.02)} & \makecell{0.55 \\ (0.01)} \\
\addlinespace[0.5em]
\midrule
\addlinespace[0.5em]
gemini-2.5-flash & \makecell{Temperature \\ (t=0.0)} & \makecell{0.57 \\ (0.02)} & \makecell{1.49 \\ (0.22)} & \makecell{1.41 \\ (0.51)} & \makecell{0.25 \\ (0.00)} & \makecell{0.30 \\ (0.00)} & \makecell{0.34 \\ (0.01)} & \makecell{0.41 \\ (0.02)} & \makecell{0.54 \\ (0.06)} \\
\addlinespace[0.5em]
gemini-2.5-flash & \makecell{Temperature \\ (t=1.0)} & \makecell{0.62 \\ (0.02)} & \makecell{1.55 \\ (0.20)} & \makecell{1.45 \\ (0.50)} & \makecell{0.31 \\ (0.01)} & \makecell{0.39 \\ (0.00)} & \makecell{0.42 \\ (0.01)} & \makecell{0.53 \\ (0.03)} & \makecell{0.62 \\ (0.05)} \\
\addlinespace[0.5em]
gemini-2.5-flash & \makecell{Temperature \\ (t=2.0)} & \makecell{0.63 \\ (0.01)} & \makecell{1.62 \\ (0.21)} & \makecell{1.42 \\ (0.50)} & \makecell{0.32 \\ (0.01)} & \makecell{0.40 \\ (0.00)} & \makecell{0.43 \\ (0.01)} & \makecell{0.55 \\ (0.03)} & \makecell{0.64 \\ (0.05)} \\
\addlinespace[0.5em]
gemini-2.5-flash & \makecell{In-Context Regeneration \\ (General)} & \makecell{0.92 \\ (0.03)} & \makecell{1.89 \\ (0.14)} & \makecell{2.25 \\ (0.49)} & \makecell{0.34 \\ (0.01)} & \makecell{0.39 \\ (0.00)} & \makecell{0.45 \\ (0.02)} & \makecell{0.59 \\ (0.04)} & \makecell{0.71 \\ (0.06)} \\
\addlinespace[0.5em]
gemini-2.5-flash & \makecell{In-Context Regeneration \\ (Task-Dependent)} & \makecell{0.77 \\ (0.02)} & \makecell{1.90 \\ (0.17)} & \makecell{1.99 \\ (0.50)} & \makecell{0.34 \\ (0.01)} & \makecell{0.38 \\ (0.00)} & \makecell{0.46 \\ (0.02)} & \makecell{0.55 \\ (0.02)} & \makecell{0.69 \\ (0.05)} \\
\addlinespace[0.5em]
gemini-2.5-flash & \makecell{System Prompt \\ (General)} & \makecell{0.63 \\ (0.01)} & \makecell{1.21 \\ (0.13)} & \makecell{1.39 \\ (0.52)} & \makecell{0.37 \\ (0.01)} & \makecell{0.45 \\ (0.00)} & \makecell{0.51 \\ (0.01)} & \makecell{0.61 \\ (0.02)} & \makecell{0.56 \\ (0.01)} \\
\addlinespace[0.5em]
gemini-2.5-flash & \makecell{System Prompt \\ (Task-Dependent)} & \makecell{0.70 \\ (0.01)} & \makecell{1.80 \\ (0.16)} & \makecell{0.93 \\ (0.16)} & \makecell{0.37 \\ (0.01)} & \makecell{0.44 \\ (0.00)} & \makecell{0.54 \\ (0.01)} & \makecell{0.58 \\ (0.02)} & \makecell{0.59 \\ (0.03)} \\
\addlinespace[0.5em]
\midrule
\addlinespace[0.5em]
Llama-3.1-8B-Instruct & \makecell{Temperature \\ (t=0.0)} & \makecell{0.57 \\ (0.02)} & \makecell{0.63 \\ (0.05)} & \makecell{0.82 \\ (0.18)} & \makecell{0.27 \\ (0.01)} & \makecell{0.34 \\ (0.00)} & \makecell{0.32 \\ (0.02)} & \makecell{0.43 \\ (0.03)} & \makecell{0.36 \\ (0.01)} \\
\addlinespace[0.5em]
Llama-3.1-8B-Instruct & \makecell{Temperature \\ (t=0.5)} & \makecell{0.64 \\ (0.01)} & \makecell{0.67 \\ (0.05)} & \makecell{0.90 \\ (0.18)} & \makecell{0.30 \\ (0.01)} & \makecell{0.37 \\ (0.00)} & \makecell{0.37 \\ (0.01)} & \makecell{0.49 \\ (0.03)} & \makecell{0.41 \\ (0.01)} \\
\addlinespace[0.5em]
Llama-3.1-8B-Instruct & \makecell{Temperature \\ (t=1.0)} & \makecell{0.68 \\ (0.01)} & \makecell{0.74 \\ (0.05)} & \makecell{1.01 \\ (0.19)} & \makecell{0.36 \\ (0.01)} & \makecell{0.40 \\ (0.00)} & \makecell{0.42 \\ (0.02)} & \makecell{0.51 \\ (0.02)} & \makecell{0.44 \\ (0.01)} \\
\addlinespace[0.5em]
Llama-3.1-8B-Instruct & \makecell{In-Context Regeneration \\ (General)} & \makecell{0.69 \\ (0.02)} & \makecell{1.31 \\ (0.14)} & \makecell{2.38 \\ (0.30)} & \makecell{0.30 \\ (0.01)} & \makecell{0.35 \\ (0.01)} & \makecell{0.41 \\ (0.03)} & \makecell{0.58 \\ (0.05)} & \makecell{0.49 \\ (0.03)} \\
\addlinespace[0.5em]
Llama-3.1-8B-Instruct & \makecell{In-Context Regeneration \\ (Task-Dependent)} & \makecell{0.66 \\ (0.02)} & \makecell{1.41 \\ (0.14)} & \makecell{2.38 \\ (0.36)} & \makecell{0.27 \\ (0.01)} & \makecell{0.41 \\ (0.01)} & \makecell{0.43 \\ (0.03)} & \makecell{0.57 \\ (0.05)} & \makecell{0.52 \\ (0.03)} \\
\addlinespace[0.5em]
Llama-3.1-8B-Instruct & \makecell{System Prompt \\ (General)} & \makecell{0.52 \\ (0.01)} & \makecell{0.56 \\ (0.02)} & \makecell{0.67 \\ (0.07)} & \makecell{0.33 \\ (0.01)} & \makecell{0.44 \\ (0.01)} & \makecell{0.45 \\ (0.02)} & \makecell{0.55 \\ (0.02)} & \makecell{0.51 \\ (0.01)} \\
\addlinespace[0.5em]
Llama-3.1-8B-Instruct & \makecell{System Prompt \\ (Task-Dependent)} & \makecell{0.62 \\ (0.02)} & \makecell{0.59 \\ (0.03)} & \makecell{0.77 \\ (0.06)} & \makecell{0.34 \\ (0.02)} & \makecell{0.45 \\ (0.01)} & \makecell{0.48 \\ (0.01)} & \makecell{0.54 \\ (0.02)} & \makecell{0.50 \\ (0.01)} \\
\addlinespace[0.5em]
\midrule
\addlinespace[0.5em]
Mistral-7B-Instruct-v0.3 & \makecell{Temperature \\ (t=0.0)} & \makecell{0.44 \\ (0.01)} & \makecell{0.41 \\ (0.01)} & \makecell{0.47 \\ (0.02)} & \makecell{0.32 \\ (0.01)} & \makecell{0.37 \\ (0.00)} & \makecell{0.36 \\ (0.01)} & \makecell{0.44 \\ (0.01)} & \makecell{0.38 \\ (0.01)} \\
\addlinespace[0.5em]
Mistral-7B-Instruct-v0.3 & \makecell{Temperature \\ (t=0.5)} & \makecell{0.51 \\ (0.01)} & \makecell{0.48 \\ (0.02)} & \makecell{0.54 \\ (0.03)} & \makecell{0.36 \\ (0.01)} & \makecell{0.41 \\ (0.00)} & \makecell{0.41 \\ (0.01)} & \makecell{0.50 \\ (0.01)} & \makecell{0.44 \\ (0.01)} \\
\addlinespace[0.5em]
Mistral-7B-Instruct-v0.3 & \makecell{Temperature \\ (t=1.0)} & \makecell{0.54 \\ (0.01)} & \makecell{0.52 \\ (0.02)} & \makecell{0.59 \\ (0.03)} & \makecell{0.39 \\ (0.01)} & \makecell{0.43 \\ (0.00)} & \makecell{0.44 \\ (0.01)} & \makecell{0.53 \\ (0.02)} & \makecell{0.46 \\ (0.01)} \\
\addlinespace[0.5em]
Mistral-7B-Instruct-v0.3 & \makecell{In-Context Regeneration \\ (General)} & \makecell{0.50 \\ (0.01)} & \makecell{0.52 \\ (0.02)} & \makecell{0.90 \\ (0.31)} & \makecell{0.31 \\ (0.02)} & \makecell{0.32 \\ (0.01)} & \makecell{0.37 \\ (0.02)} & \makecell{0.45 \\ (0.02)} & \makecell{0.42 \\ (0.01)} \\
\addlinespace[0.5em]
Mistral-7B-Instruct-v0.3 & \makecell{In-Context Regeneration \\ (Task-Dependent)} & \makecell{0.48 \\ (0.01)} & \makecell{0.49 \\ (0.02)} & \makecell{0.56 \\ (0.06)} & \makecell{0.33 \\ (0.02)} & \makecell{0.33 \\ (0.01)} & \makecell{0.35 \\ (0.02)} & \makecell{0.42 \\ (0.01)} & \makecell{0.40 \\ (0.01)} \\
\addlinespace[0.5em]
Mistral-7B-Instruct-v0.3 & \makecell{System Prompt \\ (General)} & \makecell{0.57 \\ (0.01)} & \makecell{0.62 \\ (0.01)} & \makecell{0.62 \\ (0.02)} & \makecell{0.37 \\ (0.01)} & \makecell{0.49 \\ (0.01)} & \makecell{0.54 \\ (0.01)} & \makecell{0.59 \\ (0.02)} & \makecell{0.57 \\ (0.01)} \\
\addlinespace[0.5em]
Mistral-7B-Instruct-v0.3 & \makecell{System Prompt \\ (Task-Dependent)} & \makecell{0.60 \\ (0.01)} & \makecell{0.62 \\ (0.02)} & \makecell{0.66 \\ (0.02)} & \makecell{0.40 \\ (0.01)} & \makecell{0.50 \\ (0.01)} & \makecell{0.56 \\ (0.01)} & \makecell{0.54 \\ (0.01)} & \makecell{0.58 \\ (0.01)} \\
\bottomrule
\end{tabular}
\end{table}

\clearpage

\subsection{Diversity-Quality Tradeoff Evaluation}\label{apx:dq_tradeoff_results}

\clearpage

\clearpage

\clearpage

\begin{table}[h!]
\centering
\caption{\centering Checklist-Based Quality by Model, Sampling Strategy \& Task Category.}
\label{tab:checklist_quality}
\tiny
\begin{tabular}{cccccccccc}
\toprule
\makecell{Model} & \makecell{Sampling strategy} & \makecell{A} & \makecell{B} & \makecell{C} & \makecell{D} & \makecell{E} & \makecell{F} & \makecell{G} & \makecell{H} \\
\midrule
gpt-4o & \makecell{Temperature \\ (t=0.0)} & \makecell{3.76 \\ (0.11)} & \makecell{4.61 \\ (0.15)} & \makecell{4.61 \\ (0.14)} & \makecell{4.00 \\ (0.15)} & \makecell{4.62 \\ (0.06)} & \makecell{4.40 \\ (0.19)} & \makecell{4.76 \\ (0.05)} & \makecell{4.78 \\ (0.04)} \\
\addlinespace[0.5em]
gpt-4o & \makecell{Temperature \\ (t=1.0)} & \makecell{3.74 \\ (0.12)} & \makecell{4.65 \\ (0.16)} & \makecell{4.58 \\ (0.17)} & \makecell{3.98 \\ (0.15)} & \makecell{4.66 \\ (0.06)} & \makecell{4.41 \\ (0.19)} & \makecell{4.78 \\ (0.05)} & \makecell{4.78 \\ (0.03)} \\
\addlinespace[0.5em]
gpt-4o & \makecell{Temperature \\ (t=2.0)} & \makecell{3.66 \\ (0.12)} & \makecell{4.73 \\ (0.11)} & \makecell{4.52 \\ (0.18)} & \makecell{3.99 \\ (0.16)} & \makecell{4.63 \\ (0.06)} & \makecell{4.37 \\ (0.20)} & \makecell{4.74 \\ (0.06)} & \makecell{4.77 \\ (0.04)} \\
\addlinespace[0.5em]
gpt-4o & \makecell{In-Context Regeneration \\ (General)} & \makecell{3.37 \\ (0.13)} & \makecell{4.71 \\ (0.14)} & \makecell{4.30 \\ (0.24)} & \makecell{3.92 \\ (0.14)} & \makecell{4.42 \\ (0.07)} & \makecell{4.11 \\ (0.19)} & \makecell{4.66 \\ (0.07)} & \makecell{4.14 \\ (0.09)} \\
\addlinespace[0.5em]
gpt-4o & \makecell{In-Context Regeneration \\ (Task-Dependent)} & \makecell{3.54 \\ (0.11)} & \makecell{4.81 \\ (0.09)} & \makecell{4.33 \\ (0.23)} & \makecell{3.97 \\ (0.13)} & \makecell{4.26 \\ (0.09)} & \makecell{4.00 \\ (0.20)} & \makecell{4.63 \\ (0.06)} & \makecell{4.13 \\ (0.08)} \\
\addlinespace[0.5em]
gpt-4o & \makecell{System Prompt \\ (General)} & \makecell{3.52 \\ (0.13)} & \makecell{4.56 \\ (0.13)} & \makecell{4.30 \\ (0.25)} & \makecell{3.83 \\ (0.16)} & \makecell{4.01 \\ (0.08)} & \makecell{3.82 \\ (0.18)} & \makecell{4.70 \\ (0.05)} & \makecell{4.34 \\ (0.07)} \\
\addlinespace[0.5em]
gpt-4o & \makecell{System Prompt \\ (Task-Dependent)} & \makecell{3.48 \\ (0.12)} & \makecell{4.72 \\ (0.14)} & \makecell{4.32 \\ (0.26)} & \makecell{3.47 \\ (0.17)} & \makecell{3.66 \\ (0.08)} & \makecell{3.39 \\ (0.22)} & \makecell{4.45 \\ (0.09)} & \makecell{4.12 \\ (0.08)} \\
\addlinespace[0.5em]
\addlinespace[0.5em]
\midrule
\addlinespace[0.5em]
claude-4-sonnet & \makecell{Temperature \\ (t=0.0)} & \makecell{3.05 \\ (0.15)} & \makecell{4.70 \\ (0.11)} & \makecell{4.12 \\ (0.33)} & \makecell{4.29 \\ (0.13)} & \makecell{4.79 \\ (0.04)} & \makecell{4.43 \\ (0.17)} & \makecell{4.65 \\ (0.11)} & \makecell{4.85 \\ (0.03)} \\
\addlinespace[0.5em]
claude-4-sonnet & \makecell{Temperature \\ (t=0.5)} & \makecell{3.09 \\ (0.14)} & \makecell{4.68 \\ (0.11)} & \makecell{4.19 \\ (0.30)} & \makecell{4.24 \\ (0.13)} & \makecell{4.75 \\ (0.04)} & \makecell{4.45 \\ (0.17)} & \makecell{4.69 \\ (0.11)} & \makecell{4.86 \\ (0.03)} \\
\addlinespace[0.5em]
claude-4-sonnet & \makecell{Temperature \\ (t=1.0)} & \makecell{3.09 \\ (0.14)} & \makecell{4.67 \\ (0.12)} & \makecell{4.23 \\ (0.28)} & \makecell{4.33 \\ (0.12)} & \makecell{4.76 \\ (0.04)} & \makecell{4.45 \\ (0.17)} & \makecell{4.73 \\ (0.09)} & \makecell{4.85 \\ (0.03)} \\
\addlinespace[0.5em]
claude-4-sonnet & \makecell{In-Context Regeneration \\ (General)} & \makecell{3.19 \\ (0.12)} & \makecell{4.63 \\ (0.13)} & \makecell{4.33 \\ (0.27)} & \makecell{4.30 \\ (0.12)} & \makecell{4.59 \\ (0.07)} & \makecell{4.36 \\ (0.16)} & \makecell{4.72 \\ (0.11)} & \makecell{4.70 \\ (0.04)} \\
\addlinespace[0.5em]
claude-4-sonnet & \makecell{In-Context Regeneration \\ (Task-Dependent)} & \makecell{3.14 \\ (0.13)} & \makecell{4.62 \\ (0.15)} & \makecell{4.52 \\ (0.16)} & \makecell{3.94 \\ (0.11)} & \makecell{4.22 \\ (0.07)} & \makecell{4.16 \\ (0.16)} & \makecell{4.53 \\ (0.10)} & \makecell{4.38 \\ (0.06)} \\
\addlinespace[0.5em]
claude-4-sonnet & \makecell{System Prompt \\ (General)} & \makecell{3.14 \\ (0.14)} & \makecell{4.43 \\ (0.12)} & \makecell{4.43 \\ (0.18)} & \makecell{4.30 \\ (0.11)} & \makecell{4.22 \\ (0.07)} & \makecell{4.17 \\ (0.20)} & \makecell{4.60 \\ (0.11)} & \makecell{4.53 \\ (0.06)} \\
\addlinespace[0.5em]
claude-4-sonnet & \makecell{System Prompt \\ (Task-Dependent)} & \makecell{3.26 \\ (0.14)} & \makecell{4.35 \\ (0.12)} & \makecell{4.37 \\ (0.21)} & \makecell{4.17 \\ (0.11)} & \makecell{4.03 \\ (0.08)} & \makecell{3.79 \\ (0.21)} & \makecell{4.60 \\ (0.07)} & \makecell{4.24 \\ (0.07)} \\
\addlinespace[0.5em]
\midrule
\addlinespace[0.5em]
gemini-2.5-flash & \makecell{Temperature \\ (t=0.0)} & \makecell{3.45 \\ (0.12)} & \makecell{4.70 \\ (0.16)} & \makecell{4.35 \\ (0.24)} & \makecell{4.13 \\ (0.14)} & \makecell{4.80 \\ (0.04)} & \makecell{4.31 \\ (0.19)} & \makecell{4.81 \\ (0.05)} & \makecell{4.56 \\ (0.08)} \\
\addlinespace[0.5em]
gemini-2.5-flash & \makecell{Temperature \\ (t=1.0)} & \makecell{3.37 \\ (0.11)} & \makecell{4.81 \\ (0.10)} & \makecell{4.41 \\ (0.15)} & \makecell{4.07 \\ (0.14)} & \makecell{4.81 \\ (0.04)} & \makecell{4.32 \\ (0.18)} & \makecell{4.73 \\ (0.07)} & \makecell{4.57 \\ (0.07)} \\
\addlinespace[0.5em]
gemini-2.5-flash & \makecell{Temperature \\ (t=2.0)} & \makecell{3.35 \\ (0.12)} & \makecell{4.85 \\ (0.09)} & \makecell{4.45 \\ (0.14)} & \makecell{4.05 \\ (0.14)} & \makecell{4.79 \\ (0.04)} & \makecell{4.23 \\ (0.17)} & \makecell{4.73 \\ (0.06)} & \makecell{4.54 \\ (0.08)} \\
\addlinespace[0.5em]
gemini-2.5-flash & \makecell{In-Context Regeneration \\ (General)} & \makecell{3.13 \\ (0.13)} & \makecell{4.82 \\ (0.09)} & \makecell{4.26 \\ (0.20)} & \makecell{4.13 \\ (0.13)} & \makecell{4.79 \\ (0.03)} & \makecell{4.21 \\ (0.17)} & \makecell{4.69 \\ (0.10)} & \makecell{4.32 \\ (0.08)} \\
\addlinespace[0.5em]
gemini-2.5-flash & \makecell{In-Context Regeneration \\ (Task-Dependent)} & \makecell{3.31 \\ (0.11)} & \makecell{4.88 \\ (0.10)} & \makecell{4.43 \\ (0.16)} & \makecell{3.93 \\ (0.13)} & \makecell{4.70 \\ (0.05)} & \makecell{4.12 \\ (0.17)} & \makecell{4.52 \\ (0.08)} & \makecell{4.33 \\ (0.08)} \\
\addlinespace[0.5em]
gemini-2.5-flash & \makecell{System Prompt \\ (General)} & \makecell{3.45 \\ (0.12)} & \makecell{4.82 \\ (0.06)} & \makecell{4.21 \\ (0.18)} & \makecell{4.42 \\ (0.09)} & \makecell{4.42 \\ (0.07)} & \makecell{3.95 \\ (0.21)} & \makecell{4.68 \\ (0.07)} & \makecell{4.44 \\ (0.06)} \\
\addlinespace[0.5em]
gemini-2.5-flash & \makecell{System Prompt \\ (Task-Dependent)} & \makecell{3.40 \\ (0.12)} & \makecell{4.89 \\ (0.07)} & \makecell{4.33 \\ (0.20)} & \makecell{4.26 \\ (0.11)} & \makecell{4.41 \\ (0.06)} & \makecell{3.49 \\ (0.22)} & \makecell{4.41 \\ (0.10)} & \makecell{4.26 \\ (0.07)} \\
\addlinespace[0.5em]
\midrule
\addlinespace[0.5em]
Llama-3.1-8B-Instruct & \makecell{Temperature \\ (t=0.0)} & \makecell{1.99 \\ (0.09)} & \makecell{4.54 \\ (0.20)} & \makecell{4.23 \\ (0.27)} & \makecell{3.09 \\ (0.18)} & \makecell{3.01 \\ (0.13)} & \makecell{3.57 \\ (0.26)} & \makecell{4.31 \\ (0.13)} & \makecell{4.52 \\ (0.07)} \\
\addlinespace[0.5em]
Llama-3.1-8B-Instruct & \makecell{Temperature \\ (t=0.5)} & \makecell{1.98 \\ (0.09)} & \makecell{4.59 \\ (0.15)} & \makecell{4.18 \\ (0.25)} & \makecell{3.02 \\ (0.17)} & \makecell{3.10 \\ (0.13)} & \makecell{3.70 \\ (0.24)} & \makecell{4.23 \\ (0.13)} & \makecell{4.53 \\ (0.06)} \\
\addlinespace[0.5em]
Llama-3.1-8B-Instruct & \makecell{Temperature \\ (t=1.0)} & \makecell{1.97 \\ (0.08)} & \makecell{4.48 \\ (0.18)} & \makecell{4.22 \\ (0.23)} & \makecell{2.93 \\ (0.18)} & \makecell{3.00 \\ (0.12)} & \makecell{3.61 \\ (0.24)} & \makecell{4.40 \\ (0.09)} & \makecell{4.50 \\ (0.06)} \\
\addlinespace[0.5em]
Llama-3.1-8B-Instruct & \makecell{In-Context Regeneration \\ (General)} & \makecell{2.21 \\ (0.09)} & \makecell{4.68 \\ (0.13)} & \makecell{4.12 \\ (0.29)} & \makecell{3.04 \\ (0.17)} & \makecell{2.85 \\ (0.13)} & \makecell{3.44 \\ (0.25)} & \makecell{4.37 \\ (0.10)} & \makecell{4.28 \\ (0.08)} \\
\addlinespace[0.5em]
Llama-3.1-8B-Instruct & \makecell{In-Context Regeneration \\ (Task-Dependent)} & \makecell{2.07 \\ (0.09)} & \makecell{4.59 \\ (0.18)} & \makecell{4.13 \\ (0.23)} & \makecell{2.62 \\ (0.16)} & \makecell{2.49 \\ (0.10)} & \makecell{2.93 \\ (0.22)} & \makecell{3.78 \\ (0.14)} & \makecell{3.97 \\ (0.08)} \\
\addlinespace[0.5em]
Llama-3.1-8B-Instruct & \makecell{System Prompt \\ (General)} & \makecell{2.22 \\ (0.12)} & \makecell{4.19 \\ (0.17)} & \makecell{4.05 \\ (0.29)} & \makecell{2.79 \\ (0.19)} & \makecell{2.38 \\ (0.09)} & \makecell{2.98 \\ (0.19)} & \makecell{3.95 \\ (0.12)} & \makecell{3.74 \\ (0.09)} \\
\addlinespace[0.5em]
Llama-3.1-8B-Instruct & \makecell{System Prompt \\ (Task-Dependent)} & \makecell{2.72 \\ (0.15)} & \makecell{4.42 \\ (0.13)} & \makecell{3.88 \\ (0.27)} & \makecell{2.51 \\ (0.17)} & \makecell{2.23 \\ (0.08)} & \makecell{2.48 \\ (0.18)} & \makecell{3.77 \\ (0.15)} & \makecell{3.61 \\ (0.10)} \\
\addlinespace[0.5em]
\midrule
\addlinespace[0.5em]
Mistral-7B-Instruct-v0.3 & \makecell{Temperature \\ (t=0.0)} & \makecell{2.58 \\ (0.13)} & \makecell{4.05 \\ (0.25)} & \makecell{3.37 \\ (0.23)} & \makecell{1.90 \\ (0.15)} & \makecell{3.22 \\ (0.13)} & \makecell{3.63 \\ (0.24)} & \makecell{3.75 \\ (0.15)} & \makecell{4.37 \\ (0.07)} \\
\addlinespace[0.5em]
Mistral-7B-Instruct-v0.3 & \makecell{Temperature \\ (t=0.5)} & \makecell{2.54 \\ (0.12)} & \makecell{3.97 \\ (0.24)} & \makecell{3.20 \\ (0.27)} & \makecell{1.83 \\ (0.13)} & \makecell{3.15 \\ (0.13)} & \makecell{3.62 \\ (0.23)} & \makecell{3.82 \\ (0.14)} & \makecell{4.39 \\ (0.07)} \\
\addlinespace[0.5em]
Mistral-7B-Instruct-v0.3 & \makecell{Temperature \\ (t=1.0)} & \makecell{2.41 \\ (0.11)} & \makecell{4.08 \\ (0.21)} & \makecell{3.29 \\ (0.25)} & \makecell{1.71 \\ (0.13)} & \makecell{3.12 \\ (0.13)} & \makecell{3.61 \\ (0.22)} & \makecell{3.86 \\ (0.14)} & \makecell{4.35 \\ (0.07)} \\
\addlinespace[0.5em]
Mistral-7B-Instruct-v0.3 & \makecell{In-Context Regeneration \\ (General)} & \makecell{2.64 \\ (0.13)} & \makecell{4.15 \\ (0.13)} & \makecell{3.44 \\ (0.23)} & \makecell{1.89 \\ (0.14)} & \makecell{2.95 \\ (0.14)} & \makecell{3.36 \\ (0.23)} & \makecell{3.75 \\ (0.15)} & \makecell{4.17 \\ (0.07)} \\
\addlinespace[0.5em]
Mistral-7B-Instruct-v0.3 & \makecell{In-Context Regeneration \\ (Task-Dependent)} & \makecell{2.57 \\ (0.14)} & \makecell{4.00 \\ (0.25)} & \makecell{3.60 \\ (0.32)} & \makecell{1.84 \\ (0.15)} & \makecell{2.70 \\ (0.13)} & \makecell{3.06 \\ (0.22)} & \makecell{3.45 \\ (0.16)} & \makecell{3.94 \\ (0.08)} \\
\addlinespace[0.5em]
Mistral-7B-Instruct-v0.3 & \makecell{System Prompt \\ (General)} & \makecell{2.48 \\ (0.12)} & \makecell{3.99 \\ (0.21)} & \makecell{3.07 \\ (0.37)} & \makecell{1.50 \\ (0.10)} & \makecell{2.24 \\ (0.08)} & \makecell{2.54 \\ (0.18)} & \makecell{3.45 \\ (0.17)} & \makecell{3.59 \\ (0.09)} \\
\addlinespace[0.5em]
Mistral-7B-Instruct-v0.3 & \makecell{System Prompt \\ (Task-Dependent)} & \makecell{2.55 \\ (0.12)} & \makecell{4.21 \\ (0.17)} & \makecell{3.13 \\ (0.34)} & \makecell{1.66 \\ (0.13)} & \makecell{2.08 \\ (0.08)} & \makecell{2.22 \\ (0.17)} & \makecell{3.12 \\ (0.16)} & \makecell{3.25 \\ (0.11)} \\
\bottomrule
\end{tabular}
\end{table}

\clearpage

\begin{table}[h!]
\centering
\caption{\centering Athene-RM-8B Reward by Model, Sampling Strategy \& Task Category.}
\label{tab:reward_quality}
\tiny
\begin{tabular}{cccccccccc}
\toprule
\makecell{Model} & \makecell{Sampling Strategy} & \makecell{A} & \makecell{B} & \makecell{C} & \makecell{D} &  \makecell{E} & \makecell{F} & \makecell{G} & \makecell{H} \\
\midrule
gpt-4o & \makecell{Temperature \\ (t=0.0)} & \makecell{0.37 \\ (0.09)} & \makecell{0.46 \\ (0.13)} & \makecell{0.78 \\ (0.14)} & \makecell{0.50 \\ (0.12)} & \makecell{0.28 \\ (0.06)} & \makecell{0.94 \\ (0.21)} & \makecell{0.81 \\ (0.09)} & \makecell{0.96 \\ (0.08)} \\
\addlinespace[0.5em]
gpt-4o & \makecell{Temperature \\ (t=1.0)} & \makecell{0.35 \\ (0.08)} & \makecell{0.43 \\ (0.14)} & \makecell{0.78 \\ (0.12)} & \makecell{0.47 \\ (0.13)} & \makecell{0.49 \\ (0.06)} & \makecell{0.97 \\ (0.20)} & \makecell{0.81 \\ (0.10)} & \makecell{0.98 \\ (0.07)} \\
\addlinespace[0.5em]
gpt-4o & \makecell{Temperature \\ (t=2.0)} & \makecell{0.34 \\ (0.08)} & \makecell{0.41 \\ (0.12)} & \makecell{0.80 \\ (0.12)} & \makecell{0.45 \\ (0.14)} & \makecell{0.56 \\ (0.08)} & \makecell{1.04 \\ (0.21)} & \makecell{0.85 \\ (0.10)} & \makecell{1.07 \\ (0.07)} \\
\addlinespace[0.5em]
gpt-4o & \makecell{In-Context Regeneration \\ (General)} & \makecell{-0.38 \\ (0.08)} & \makecell{-0.36 \\ (0.11)} & \makecell{0.04 \\ (0.21)} & \makecell{0.45 \\ (0.11)} & \makecell{-0.40 \\ (0.07)} & \makecell{0.30 \\ (0.19)} & \makecell{0.65 \\ (0.15)} & \makecell{-0.38 \\ (0.10)} \\
\addlinespace[0.5em]
gpt-4o & \makecell{In-Context Regeneration \\ (Task-Dependent)} & \makecell{-0.08 \\ (0.08)} & \makecell{-0.36 \\ (0.10)} & \makecell{0.05 \\ (0.21)} & \makecell{0.54 \\ (0.12)} & \makecell{-0.24 \\ (0.08)} & \makecell{0.29 \\ (0.24)} & \makecell{0.61 \\ (0.16)} & \makecell{-0.24 \\ (0.10)} \\
\addlinespace[0.5em]
gpt-4o & \makecell{System Prompt \\ (General)} & \makecell{-0.05 \\ (0.08)} & \makecell{-0.05 \\ (0.16)} & \makecell{0.15 \\ (0.22)} & \makecell{0.39 \\ (0.13)} & \makecell{-1.12 \\ (0.08)} & \makecell{-0.64 \\ (0.18)} & \makecell{0.22 \\ (0.13)} & \makecell{-0.70 \\ (0.10)} \\
\addlinespace[0.5em]
gpt-4o & \makecell{System Prompt \\ (Task-Dependent)} & \makecell{-0.19 \\ (0.09)} & \makecell{-0.25 \\ (0.19)} & \makecell{-0.13 \\ (0.24)} & \makecell{-0.02 \\ (0.12)} & \makecell{-1.18 \\ (0.07)} & \makecell{-1.25 \\ (0.20)} & \makecell{0.16 \\ (0.15)} & \makecell{-0.91 \\ (0.11)} \\
\addlinespace[0.5em]
\midrule
\addlinespace[0.5em]
claude-4-sonnet & \makecell{Temperature \\ (t=0.0)} & \makecell{0.42 \\ (0.08)} & \makecell{0.12 \\ (0.20)} & \makecell{0.69 \\ (0.20)} & \makecell{0.08 \\ (0.15)} & \makecell{1.10 \\ (0.07)} & \makecell{1.07 \\ (0.16)} & \makecell{0.96 \\ (0.12)} & \makecell{1.13 \\ (0.07)} \\
\addlinespace[0.5em]
claude-4-sonnet & \makecell{Temperature \\ (t=0.5)} & \makecell{0.43 \\ (0.08)} & \makecell{0.11 \\ (0.21)} & \makecell{0.68 \\ (0.20)} & \makecell{0.03 \\ (0.15)} & \makecell{1.08 \\ (0.07)} & \makecell{1.07 \\ (0.17)} & \makecell{0.97 \\ (0.12)} & \makecell{1.17 \\ (0.07)} \\
\addlinespace[0.5em]
claude-4-sonnet & \makecell{Temperature \\ (t=1.0)} & \makecell{0.41 \\ (0.08)} & \makecell{0.11 \\ (0.21)} & \makecell{0.67 \\ (0.20)} & \makecell{0.14 \\ (0.14)} & \makecell{1.09 \\ (0.07)} & \makecell{1.13 \\ (0.15)} & \makecell{0.96 \\ (0.11)} & \makecell{1.16 \\ (0.07)} \\
\addlinespace[0.5em]
claude-4-sonnet & \makecell{In-Context Regeneration \\ (General)} & \makecell{0.33 \\ (0.08)} & \makecell{-0.06 \\ (0.16)} & \makecell{0.46 \\ (0.16)} & \makecell{0.25 \\ (0.13)} & \makecell{0.88 \\ (0.07)} & \makecell{0.89 \\ (0.15)} & \makecell{0.70 \\ (0.11)} & \makecell{0.78 \\ (0.07)} \\
\addlinespace[0.5em]
claude-4-sonnet & \makecell{In-Context Regeneration \\ (Task-Dependent)} & \makecell{0.26 \\ (0.08)} & \makecell{-0.25 \\ (0.19)} & \makecell{0.42 \\ (0.16)} & \makecell{-0.15 \\ (0.11)} & \makecell{0.76 \\ (0.06)} & \makecell{0.74 \\ (0.15)} & \makecell{0.48 \\ (0.12)} & \makecell{0.71 \\ (0.07)} \\
\addlinespace[0.5em]
claude-4-sonnet & \makecell{System Prompt \\ (General)} & \makecell{0.29 \\ (0.08)} & \makecell{0.20 \\ (0.12)} & \makecell{0.26 \\ (0.21)} & \makecell{0.39 \\ (0.11)} & \makecell{-0.25 \\ (0.08)} & \makecell{0.28 \\ (0.18)} & \makecell{0.53 \\ (0.12)} & \makecell{0.29 \\ (0.08)} \\
\addlinespace[0.5em]
claude-4-sonnet & \makecell{System Prompt \\ (Task-Dependent)} & \makecell{0.28 \\ (0.08)} & \makecell{0.12 \\ (0.15)} & \makecell{0.29 \\ (0.19)} & \makecell{0.18 \\ (0.11)} & \makecell{-0.35 \\ (0.08)} & \makecell{-0.16 \\ (0.16)} & \makecell{0.50 \\ (0.13)} & \makecell{-0.02 \\ (0.10)} \\
\addlinespace[0.5em]
\midrule
\addlinespace[0.5em]
gemini-2.5-flash & \makecell{Temperature \\ (t=0.0)} & \makecell{0.05 \\ (0.10)} & \makecell{-0.30 \\ (0.09)} & \makecell{0.22 \\ (0.17)} & \makecell{-0.20 \\ (0.14)} & \makecell{1.25 \\ (0.15)} & \makecell{0.65 \\ (0.23)} & \makecell{0.64 \\ (0.14)} & \makecell{0.56 \\ (0.14)} \\
\addlinespace[0.5em]
gemini-2.5-flash & \makecell{Temperature \\ (t=1.0)} & \makecell{0.12 \\ (0.09)} & \makecell{-0.21 \\ (0.10)} & \makecell{0.37 \\ (0.15)} & \makecell{-0.24 \\ (0.15)} & \makecell{1.27 \\ (0.12)} & \makecell{0.70 \\ (0.26)} & \makecell{0.67 \\ (0.13)} & \makecell{0.55 \\ (0.13)} \\
\addlinespace[0.5em]
gemini-2.5-flash & \makecell{Temperature \\ (t=2.0)} & \makecell{0.13 \\ (0.09)} & \makecell{-0.31 \\ (0.10)} & \makecell{0.35 \\ (0.15)} & \makecell{-0.26 \\ (0.15)} & \makecell{1.05 \\ (0.13)} & \makecell{0.69 \\ (0.22)} & \makecell{0.59 \\ (0.12)} & \makecell{0.54 \\ (0.13)} \\
\addlinespace[0.5em]
gemini-2.5-flash & \makecell{In-Context Regeneration \\ (General)} & \makecell{-0.51 \\ (0.09)} & \makecell{-0.64 \\ (0.11)} & \makecell{0.27 \\ (0.14)} & \makecell{0.04 \\ (0.13)} & \makecell{1.27 \\ (0.09)} & \makecell{0.71 \\ (0.22)} & \makecell{0.62 \\ (0.14)} & \makecell{0.45 \\ (0.15)} \\
\addlinespace[0.5em]
gemini-2.5-flash & \makecell{In-Context Regeneration \\ (Task-Dependent)} & \makecell{-0.16 \\ (0.09)} & \makecell{-0.62 \\ (0.11)} & \makecell{0.15 \\ (0.15)} & \makecell{-0.28 \\ (0.13)} & \makecell{1.27 \\ (0.09)} & \makecell{0.56 \\ (0.24)} & \makecell{0.46 \\ (0.15)} & \makecell{0.44 \\ (0.13)} \\
\addlinespace[0.5em]
gemini-2.5-flash & \makecell{System Prompt \\ (General)} & \makecell{0.10 \\ (0.08)} & \makecell{-0.28 \\ (0.12)} & \makecell{-0.04 \\ (0.23)} & \makecell{0.24 \\ (0.11)} & \makecell{-0.23 \\ (0.07)} & \makecell{0.16 \\ (0.19)} & \makecell{0.69 \\ (0.11)} & \makecell{0.26 \\ (0.09)} \\
\addlinespace[0.5em]
gemini-2.5-flash & \makecell{System Prompt \\ (Task-Dependent)} & \makecell{-0.22 \\ (0.08)} & \makecell{-0.56 \\ (0.09)} & \makecell{0.08 \\ (0.20)} & \makecell{0.12 \\ (0.12)} & \makecell{0.37 \\ (0.09)} & \makecell{-0.62 \\ (0.18)} & \makecell{0.32 \\ (0.16)} & \makecell{0.02 \\ (0.11)} \\
\addlinespace[0.5em]
\midrule
\addlinespace[0.5em]
Llama-3.1-8B-Instruct & \makecell{Temperature \\ (t=0.0)} & \makecell{-0.84 \\ (0.07)} & \makecell{0.02 \\ (0.10)} & \makecell{0.03 \\ (0.20)} & \makecell{-0.87 \\ (0.18)} & \makecell{-0.91 \\ (0.07)} & \makecell{-0.49 \\ (0.27)} & \makecell{-0.22 \\ (0.15)} & \makecell{0.04 \\ (0.08)} \\
\addlinespace[0.5em]
Llama-3.1-8B-Instruct & \makecell{Temperature \\ (t=0.5)} & \makecell{-0.81 \\ (0.07)} & \makecell{0.02 \\ (0.12)} & \makecell{0.03 \\ (0.14)} & \makecell{-0.94 \\ (0.17)} & \makecell{-0.90 \\ (0.07)} & \makecell{-0.30 \\ (0.18)} & \makecell{-0.16 \\ (0.14)} & \makecell{0.04 \\ (0.07)} \\
\addlinespace[0.5em]
Llama-3.1-8B-Instruct & \makecell{Temperature \\ (t=1.0)} & \makecell{-0.95 \\ (0.06)} & \makecell{-0.17 \\ (0.16)} & \makecell{0.14 \\ (0.13)} & \makecell{-0.86 \\ (0.15)} & \makecell{-0.76 \\ (0.07)} & \makecell{-0.29 \\ (0.17)} & \makecell{-0.06 \\ (0.12)} & \makecell{0.06 \\ (0.07)} \\
\addlinespace[0.5em]
Llama-3.1-8B-Instruct & \makecell{In-Context Regeneration \\ (General)} & \makecell{-0.62 \\ (0.06)} & \makecell{-0.34 \\ (0.15)} & \makecell{-0.21 \\ (0.23)} & \makecell{-0.92 \\ (0.17)} & \makecell{-1.34 \\ (0.08)} & \makecell{-0.87 \\ (0.14)} & \makecell{-0.29 \\ (0.14)} & \makecell{-0.51 \\ (0.08)} \\
\addlinespace[0.5em]
Llama-3.1-8B-Instruct & \makecell{In-Context Regeneration \\ (Task-Dependent)} & \makecell{-0.74 \\ (0.06)} & \makecell{-0.57 \\ (0.12)} & \makecell{-0.09 \\ (0.16)} & \makecell{-1.47 \\ (0.16)} & \makecell{-1.59 \\ (0.08)} & \makecell{-1.24 \\ (0.13)} & \makecell{-0.75 \\ (0.15)} & \makecell{-0.92 \\ (0.09)} \\
\addlinespace[0.5em]
Llama-3.1-8B-Instruct & \makecell{System Prompt \\ (General)} & \makecell{-0.24 \\ (0.08)} & \makecell{-0.36 \\ (0.22)} & \makecell{-0.02 \\ (0.19)} & \makecell{-0.83 \\ (0.16)} & \makecell{-2.08 \\ (0.09)} & \makecell{-1.53 \\ (0.17)} & \makecell{-0.80 \\ (0.17)} & \makecell{-1.35 \\ (0.11)} \\
\addlinespace[0.5em]
Llama-3.1-8B-Instruct & \makecell{System Prompt \\ (Task-Dependent)} & \makecell{-0.14 \\ (0.09)} & \makecell{-0.19 \\ (0.19)} & \makecell{0.00 \\ (0.22)} & \makecell{-1.05 \\ (0.15)} & \makecell{-2.24 \\ (0.09)} & \makecell{-2.04 \\ (0.19)} & \makecell{-0.86 \\ (0.18)} & \makecell{-1.48 \\ (0.12)} \\
\addlinespace[0.5em]
\midrule
\addlinespace[0.5em]
Mistral-7B-Instruct-v0.3 & \makecell{Temperature \\ (t=0.0)} & \makecell{-0.08 \\ (0.07)} & \makecell{-0.08 \\ (0.20)} & \makecell{-0.78 \\ (0.15)} & \makecell{-1.30 \\ (0.15)} & \makecell{-1.12 \\ (0.08)} & \makecell{-0.45 \\ (0.13)} & \makecell{-0.35 \\ (0.13)} & \makecell{-0.39 \\ (0.07)} \\
\addlinespace[0.5em]
Mistral-7B-Instruct-v0.3 & \makecell{Temperature \\ (t=0.5)} & \makecell{-0.06 \\ (0.06)} & \makecell{-0.07 \\ (0.16)} & \makecell{-0.84 \\ (0.15)} & \makecell{-1.37 \\ (0.13)} & \makecell{-1.06 \\ (0.08)} & \makecell{-0.39 \\ (0.11)} & \makecell{-0.29 \\ (0.11)} & \makecell{-0.32 \\ (0.06)} \\
\addlinespace[0.5em]
Mistral-7B-Instruct-v0.3 & \makecell{Temperature \\ (t=1.0)} & \makecell{-0.09 \\ (0.06)} & \makecell{-0.17 \\ (0.16)} & \makecell{-0.74 \\ (0.14)} & \makecell{-1.40 \\ (0.12)} & \makecell{-0.98 \\ (0.07)} & \makecell{-0.36 \\ (0.12)} & \makecell{-0.18 \\ (0.10)} & \makecell{-0.28 \\ (0.06)} \\
\addlinespace[0.5em]
Mistral-7B-Instruct-v0.3 & \makecell{In-Context Regeneration \\ (General)} & \makecell{-0.18 \\ (0.07)} & \makecell{-0.33 \\ (0.13)} & \makecell{-0.74 \\ (0.20)} & \makecell{-1.32 \\ (0.13)} & \makecell{-1.22 \\ (0.09)} & \makecell{-0.76 \\ (0.13)} & \makecell{-0.46 \\ (0.12)} & \makecell{-0.73 \\ (0.08)} \\
\addlinespace[0.5em]
Mistral-7B-Instruct-v0.3 & \makecell{In-Context Regeneration \\ (Task-Dependent)} & \makecell{-0.10 \\ (0.07)} & \makecell{-0.48 \\ (0.15)} & \makecell{-0.59 \\ (0.21)} & \makecell{-1.38 \\ (0.13)} & \makecell{-1.54 \\ (0.11)} & \makecell{-1.01 \\ (0.17)} & \makecell{-1.00 \\ (0.12)} & \makecell{-1.07 \\ (0.09)} \\
\addlinespace[0.5em]
Mistral-7B-Instruct-v0.3 & \makecell{System Prompt \\ (General)} & \makecell{-0.41 \\ (0.08)} & \makecell{-0.49 \\ (0.25)} & \makecell{-1.17 \\ (0.22)} & \makecell{-1.81 \\ (0.13)} & \makecell{-2.19 \\ (0.08)} & \makecell{-2.06 \\ (0.21)} & \makecell{-1.04 \\ (0.17)} & \makecell{-1.81 \\ (0.13)} \\
\addlinespace[0.5em]
Mistral-7B-Instruct-v0.3 & \makecell{System Prompt \\ (Task-Dependent)} & \makecell{-0.41 \\ (0.08)} & \makecell{-0.33 \\ (0.18)} & \makecell{-1.20 \\ (0.25)} & \makecell{-1.86 \\ (0.15)} & \makecell{-2.30 \\ (0.09)} & \makecell{-2.46 \\ (0.19)} & \makecell{-1.29 \\ (0.16)} & \makecell{-2.15 \\ (0.15)} \\
\bottomrule
\end{tabular}
\end{table}

\clearpage

\begin{table}[h!]
\centering
\tiny
\caption{\centering Accuracy by Model, Sampling Strategy, and Evaluation Dataset.}
\label{tab:accuracy}
\begin{tabular}{cccc}
\multicolumn{4}{c}{(For Tasks with Singular Verifiable Rewards)} \\
\addlinespace[0.5em]
\toprule
\makecell{Model} & \makecell{Sampling Strategy} & \makecell{Math-500} & \makecell{Simple-QA} \\
\midrule
gpt-4o & \makecell{Temperature \\ (t=0.0)} & \makecell{0.59 \\ (0.06)} & \makecell{0.37 \\ (0.06)} \\
\addlinespace[0.5em]
gpt-4o & \makecell{Temperature \\ (t=1.0)} & \makecell{0.59 \\ (0.06)} & \makecell{0.36 \\ (0.06)} \\
\addlinespace[0.5em]
gpt-4o & \makecell{Temperature \\ (t=2.0)} & \makecell{0.57 \\ (0.06)} & \makecell{0.38 \\ (0.06)} \\
\addlinespace[0.5em]
gpt-4o & \makecell{In-Context Regeneration \\ (General)} & \makecell{0.57 \\ (0.07)} & \makecell{0.30 \\ (0.06)} \\
\addlinespace[0.5em]
gpt-4o & \makecell{In-Context Regeneration \\ (Task-Dependent)} & \makecell{0.59 \\ (0.07)} & \makecell{0.35 \\ (0.07)} \\
\addlinespace[0.5em]
gpt-4o & \makecell{System Prompt \\ (General)} & \makecell{0.61 \\ (0.07)} & \makecell{0.37 \\ (0.07)} \\
\addlinespace[0.5em]
gpt-4o & \makecell{System Prompt \\ (Task-Dependent)} & \makecell{0.63 \\ (0.07)} & \makecell{0.28 \\ (0.06)} \\
\addlinespace[0.5em]
\midrule
\addlinespace[0.5em]
claude-4-sonnet & \makecell{Temperature \\ (t=0.0)} & \makecell{0.69 \\ (0.06)} & \makecell{0.17 \\ (0.05)} \\
\addlinespace[0.5em]
claude-4-sonnet & \makecell{Temperature \\ (t=0.5)} & \makecell{0.68 \\ (0.06)} & \makecell{0.17 \\ (0.05)} \\
\addlinespace[0.5em]
claude-4-sonnet & \makecell{Temperature \\ (t=1.0)} & \makecell{0.68 \\ (0.06)} & \makecell{0.18 \\ (0.05)} \\
\addlinespace[0.5em]
claude-4-sonnet & \makecell{In-Context Regeneration \\ (General)} & \makecell{0.71 \\ (0.06)} & \makecell{0.19 \\ (0.05)} \\
\addlinespace[0.5em]
claude-4-sonnet & \makecell{In-Context Regeneration \\ (Task-Dependent)} & \makecell{0.69 \\ (0.06)} & \makecell{0.17 \\ (0.05)} \\
\addlinespace[0.5em]
claude-4-sonnet & \makecell{System Prompt \\ (General)} & \makecell{0.70 \\ (0.06)} & \makecell{0.21 \\ (0.06)} \\
\addlinespace[0.5em]
claude-4-sonnet & \makecell{System Prompt \\ (Task-Dependent)} & \makecell{0.73 \\ (0.06)} & \makecell{0.21 \\ (0.06)} \\
\addlinespace[0.5em]
\midrule
\addlinespace[0.5em]
gemini-2.5-flash & \makecell{Temperature \\ (t=0.0)} & \makecell{0.63 \\ (0.07)} & \makecell{0.34 \\ (0.06)} \\
\addlinespace[0.5em]
gemini-2.5-flash & \makecell{Temperature \\ (t=1.0)} & \makecell{0.64 \\ (0.06)} & \makecell{0.27 \\ (0.05)} \\
\addlinespace[0.5em]
gemini-2.5-flash & \makecell{Temperature \\ (t=2.0)} & \makecell{0.63 \\ (0.06)} & \makecell{0.25 \\ (0.05)} \\
\addlinespace[0.5em]
gemini-2.5-flash & \makecell{In-Context Regeneration \\ (General)} & \makecell{0.65 \\ (0.06)} & \makecell{0.31 \\ (0.06)} \\
\addlinespace[0.5em]
gemini-2.5-flash & \makecell{In-Context Regeneration \\ (Task-Dependent)} & \makecell{0.62 \\ (0.06)} & \makecell{0.35 \\ (0.07)} \\
\addlinespace[0.5em]
gemini-2.5-flash & \makecell{System Prompt \\ (General)} & \makecell{0.75 \\ (0.06)} & \makecell{0.25 \\ (0.06)} \\
\addlinespace[0.5em]
gemini-2.5-flash & \makecell{System Prompt \\ (Task-Dependent)} & \makecell{0.73 \\ (0.06)} & \makecell{0.29 \\ (0.06)} \\
\addlinespace[0.5em]
\midrule
\addlinespace[0.5em]
Llama-3.1-8B-Instruct & \makecell{Temperature \\ (t=0.0)} & \makecell{0.40 \\ (0.06)} & \makecell{0.03 \\ (0.02)} \\
\addlinespace[0.5em]
Llama-3.1-8B-Instruct & \makecell{Temperature \\ (t=1.0)} & \makecell{0.36 \\ (0.05)} & \makecell{0.03 \\ (0.02)} \\
\addlinespace[0.5em]
Llama-3.1-8B-Instruct & \makecell{Temperature \\ (t=2.0)} & \makecell{0.36 \\ (0.05)} & \makecell{0.03 \\ (0.02)} \\
\addlinespace[0.5em]
Llama-3.1-8B-Instruct & \makecell{In-Context Regeneration \\ (General)} & \makecell{0.39 \\ (0.06)} & \makecell{0.03 \\ (0.02)} \\
\addlinespace[0.5em]
Llama-3.1-8B-Instruct & \makecell{In-Context Regeneration \\ (Task-Dependent)} & \makecell{0.33 \\ (0.06)} & \makecell{0.02 \\ (0.01)} \\
\addlinespace[0.5em]
Llama-3.1-8B-Instruct & \makecell{System Prompt \\ (General)} & \makecell{0.41 \\ (0.07)} & \makecell{0.05 \\ (0.03)} \\
\addlinespace[0.5em]
Llama-3.1-8B-Instruct & \makecell{System Prompt \\ (Task-Dependent)} & \makecell{0.41 \\ (0.06)} & \makecell{0.08 \\ (0.04)} \\
\addlinespace[0.5em]
\midrule
\addlinespace[0.5em]
Mistral-7B-Instruct-v0.3 & \makecell{Temperature \\ (t=0.0)} & \makecell{0.08 \\ (0.03)} & \makecell{0.04 \\ (0.02)} \\
\addlinespace[0.5em]
Mistral-7B-Instruct-v0.3 & \makecell{Temperature \\ (t=1.0)} & \makecell{0.08 \\ (0.02)} & \makecell{0.05 \\ (0.03)} \\
\addlinespace[0.5em]
Mistral-7B-Instruct-v0.3 & \makecell{Temperature \\ (t=2.0)} & \makecell{0.06 \\ (0.02)} & \makecell{0.06 \\ (0.03)} \\
\addlinespace[0.5em]
Mistral-7B-Instruct-v0.3 & \makecell{In-Context Regeneration \\ (General)} & \makecell{0.09 \\ (0.04)} & \makecell{0.02 \\ (0.02)} \\
\addlinespace[0.5em]
Mistral-7B-Instruct-v0.3 & \makecell{In-Context Regeneration \\ (Task-Dependent)} & \makecell{0.14 \\ (0.05)} & \makecell{0.06 \\ (0.03)} \\
\addlinespace[0.5em]
Mistral-7B-Instruct-v0.3 & \makecell{System Prompt \\ (General)} & \makecell{0.06 \\ (0.03)} & \makecell{0.06 \\ (0.03)} \\
\addlinespace[0.5em]
Mistral-7B-Instruct-v0.3 & \makecell{System Prompt \\ (Task-Dependent)} & \makecell{0.09 \\ (0.03)} & \makecell{0.05 \\ (0.02)} \\
\bottomrule
\end{tabular}
\end{table}

\clearpage

\subsection{Alignment Experiment Results}\label{apx:alignment_exp_results}

\clearpage

\begin{table}[h!]
\centering
\caption{\centering \# of Functionally Diverse Responses by Alignment Method (DPO vs GRPO)}
\label{tab:fun_div_alignment}
\tiny
\begin{tabular}{cccccccccc}
\multicolumn{10}{c}{(Using Only GPT-4o as the Functional Diversity Judge)} \\
\toprule
\makecell{Model} & \makecell{Sampling Strategy} & \makecell{A} & \makecell{B} & \makecell{C} & \makecell{D} & \makecell{E} & \makecell{F} & \makecell{G} & \makecell{H} \\
\midrule
Llama-3.1-8B-Instruct & \makecell{Temperature \\ (t=0.1)} & \makecell{1.08 \\ (0.05)} & \makecell{1.25 \\ (0.14)} & \makecell{1.71 \\ (0.27)} & \makecell{1.93 \\ (0.18)} & \makecell{1.56 \\ (0.14)} & \makecell{1.52 \\ (0.23)} & \makecell{1.47 \\ (0.15)} & \makecell{1.09 \\ (0.04)} \\
\addlinespace
Llama-3.1-8B-Instruct & \makecell{Temperature \\ (t=0.5)} & \makecell{1.19 \\ (0.09)} & \makecell{1.75 \\ (0.25)} & \makecell{2.79 \\ (0.47)} & \makecell{2.22 \\ (0.18)} & \makecell{2.02 \\ (0.17)} & \makecell{1.48 \\ (0.24)} & \makecell{1.91 \\ (0.21)} & \makecell{1.47 \\ (0.10)} \\
\addlinespace
Llama-3.1-8B-Instruct & \makecell{Temperature \\ (t=1.0)} & \makecell{1.45 \\ (0.13)} & \makecell{2.31 \\ (0.30)} & \makecell{3.71 \\ (0.37)} & \makecell{2.42 \\ (0.20)} & \makecell{2.68 \\ (0.21)} & \makecell{2.00 \\ (0.29)} & \makecell{2.51 \\ (0.24)} & \makecell{1.78 \\ (0.13)} \\
\addlinespace
\midrule
\addlinespace
\makecell{Online DPO \\ (Wildchat, $\beta=0.01$, Step 1000)} & \makecell{Temperature \\ (t=0.1)} & \makecell{1.72 \\ (0.16)} & \makecell{1.38 \\ (0.20)} & \makecell{2.00 \\ (0.33)} & \makecell{1.42 \\ (0.11)} & \makecell{1.66 \\ (0.12)} & \makecell{1.35 \\ (0.18)} & \makecell{1.47 \\ (0.11)} & \makecell{1.49 \\ (0.09)} \\
\addlinespace
\makecell{Online DPO \\ (Wildchat, $\beta=0.01$, Step 1000)} & \makecell{Temperature \\ (t=0.5)} & \makecell{3.09 \\ (0.21)} & \makecell{2.31 \\ (0.30)} & \makecell{4.00 \\ (0.42)} & \makecell{2.11 \\ (0.18)} & \makecell{3.26 \\ (0.18)} & \makecell{2.22 \\ (0.31)} & \makecell{2.53 \\ (0.20)} & \makecell{1.93 \\ (0.14)} \\
\addlinespace
\makecell{Online DPO \\ (Wildchat, $\beta=0.01$, Step 1000)} & \makecell{Temperature \\ (t=1.0)} & \makecell{3.43 \\ (0.21)} & \makecell{2.38 \\ (0.30)} & \makecell{4.71 \\ (0.16)} & \makecell{2.40 \\ (0.19)} & \makecell{3.60 \\ (0.20)} & \makecell{2.70 \\ (0.34)} & \makecell{3.31 \\ (0.22)} & \makecell{2.22 \\ (0.16)} \\
\addlinespace
\midrule
\addlinespace
\makecell{Online DPO \\ (Wildchat, $\beta=0.1$, Step 1000)} & \makecell{Temperature \\ (t=0.1)} & \makecell{1.19 \\ (0.08)} & \makecell{1.12 \\ (0.09)} & \makecell{1.64 \\ (0.25)} & \makecell{2.13 \\ (0.16)} & \makecell{1.46 \\ (0.09)} & \makecell{1.43 \\ (0.23)} & \makecell{2.11 \\ (0.22)} & \makecell{1.31 \\ (0.07)} \\
\addlinespace
\makecell{Online DPO \\ (Wildchat, $\beta=0.1$, Step 1000)} & \makecell{Temperature \\ (t=0.5)} & \makecell{1.43 \\ (0.12)} & \makecell{1.69 \\ (0.27)} & \makecell{2.43 \\ (0.42)} & \makecell{2.27 \\ (0.18)} & \makecell{2.38 \\ (0.15)} & \makecell{1.87 \\ (0.26)} & \makecell{2.71 \\ (0.23)} & \makecell{1.43 \\ (0.08)} \\
\addlinespace
\makecell{Online DPO \\ (Wildchat, $\beta=0.1$, Step 1000)} & \makecell{Temperature \\ (t=1.0)} & \makecell{1.64 \\ (0.15)} & \makecell{2.12 \\ (0.26)} & \makecell{2.86 \\ (0.38)} & \makecell{2.29 \\ (0.18)} & \makecell{2.52 \\ (0.17)} & \makecell{2.13 \\ (0.31)} & \makecell{2.89 \\ (0.23)} & \makecell{1.58 \\ (0.10)} \\
\addlinespace
\midrule
\addlinespace
\makecell{Online DPO \\ (Ultrafeedback, $\beta=0.01$, Step 1000)} & \makecell{Temperature \\ (t=0.1)} & \makecell{2.43 \\ (0.23)} & \makecell{1.31 \\ (0.20)} & \makecell{1.86 \\ (0.38)} & \makecell{1.05 \\ (0.03)} & \makecell{1.04 \\ (0.03)} & \makecell{1.48 \\ (0.24)} & \makecell{1.73 \\ (0.19)} & \makecell{1.25 \\ (0.09)} \\
\addlinespace
\makecell{Online DPO \\ (Ultrafeedback, $\beta=0.01$, Step 1000)} & \makecell{Temperature \\ (t=0.5)} & \makecell{2.13 \\ (0.20)} & \makecell{1.38 \\ (0.22)} & \makecell{2.14 \\ (0.43)} & \makecell{1.07 \\ (0.06)} & \makecell{1.00 \\ (0.00)} & \makecell{1.43 \\ (0.23)} & \makecell{1.58 \\ (0.18)} & \makecell{1.24 \\ (0.09)} \\
\addlinespace
\makecell{Online DPO \\ (Ultrafeedback, $\beta=0.01$, Step 1000)}& \makecell{Temperature \\ (t=1.0)} & \makecell{2.17 \\ (0.20)} & \makecell{1.62 \\ (0.31)} & \makecell{1.86 \\ (0.39)} & \makecell{1.07 \\ (0.06)} & \makecell{1.02 \\ (0.02)} & \makecell{1.52 \\ (0.23)} & \makecell{1.69 \\ (0.20)} & \makecell{1.23 \\ (0.08)} \\
\addlinespace
\midrule
\addlinespace
\makecell{GRPO \\ (Wildchat, $\beta=0.001$, Step 1000)} & \makecell{Temperature \\ (t=0.1)} & \makecell{2.79 \\ (0.21)} & \makecell{1.62 \\ (0.24)} & \makecell{2.71 \\ (0.40)} & \makecell{2.84 \\ (0.19)} & \makecell{1.12 \\ (0.05)} & \makecell{1.09 \\ (0.06)} & \makecell{1.22 \\ (0.08)} & \makecell{1.12 \\ (0.04)} \\
\addlinespace
\makecell{GRPO \\ (Wildchat, $\beta=0.001$, Step 1000)} & \makecell{Temperature \\ (t=0.5)} & \makecell{3.55 \\ (0.19)} & \makecell{2.31 \\ (0.28)} & \makecell{3.29 \\ (0.34)} & \makecell{3.13 \\ (0.20)} & \makecell{1.18 \\ (0.07)} & \makecell{1.17 \\ (0.08)} & \makecell{1.47 \\ (0.15)} & \makecell{1.27 \\ (0.07)} \\
\addlinespace
\makecell{GRPO \\ (Wildchat, $\beta=0.001$, Step 1000)} & \makecell{Temperature \\ (t=1.0)} & \makecell{4.00 \\ (0.16)} & \makecell{2.69 \\ (0.35)} & \makecell{3.93 \\ (0.27)} & \makecell{3.09 \\ (0.20)} & \makecell{1.16 \\ (0.06)} & \makecell{1.04 \\ (0.04)} & \makecell{1.67 \\ (0.17)} & \makecell{1.36 \\ (0.10)} \\
\addlinespace
\midrule
\addlinespace
\makecell{GRPO \\ (Wildchat, $\beta=0.01$, Step 1000)} & \makecell{Temperature \\ (t=0.1)} & \makecell{1.96 \\ (0.15)} & \makecell{1.88 \\ (0.24)} & \makecell{3.71 \\ (0.29)} & \makecell{2.25 \\ (0.16)} & \makecell{1.26 \\ (0.08)} & \makecell{1.09 \\ (0.06)} & \makecell{1.51 \\ (0.15)} & \makecell{1.26 \\ (0.09)} \\
\addlinespace
\makecell{GRPO \\ (Wildchat, $\beta=0.01$, Step 1000)} & \makecell{Temperature \\ (t=0.5)} & \makecell{2.19 \\ (0.19)} & \makecell{2.06 \\ (0.30)} & \makecell{3.86 \\ (0.40)} & \makecell{2.29 \\ (0.18)} & \makecell{1.62 \\ (0.12)} & \makecell{1.17 \\ (0.10)} & \makecell{1.71 \\ (0.18)} & \makecell{1.41 \\ (0.10)} \\
\addlinespace
\makecell{GRPO \\ (Wildchat, $\beta=0.01$, Step 1000)} & \makecell{Temperature \\ (t=1.0)} & \makecell{2.32 \\ (0.19)} & \makecell{2.56 \\ (0.34)} & \makecell{4.29 \\ (0.24)} & \makecell{2.38 \\ (0.18)} & \makecell{1.74 \\ (0.15)} & \makecell{1.35 \\ (0.16)} & \makecell{2.00 \\ (0.20)} & \makecell{1.42 \\ (0.10)} \\
\addlinespace
\midrule
\addlinespace
\makecell{GRPO \\ (Ultrafeedback, $\beta=0.001$, Step 1000)} & \makecell{Temperature \\ (t=0.1)} & \makecell{3.15 \\ (0.19)} & \makecell{1.69 \\ (0.24)} & \makecell{3.14 \\ (0.42)} & \makecell{2.25 \\ (0.17)} & \makecell{1.20 \\ (0.06)} & \makecell{1.52 \\ (0.21)} & \makecell{1.96 \\ (0.20)} & \makecell{1.24 \\ (0.06)} \\
\addlinespace
\makecell{GRPO \\ (Ultrafeedback, $\beta=0.001$, Step 1000)} & \makecell{Temperature \\ (t=0.5)} & \makecell{3.66 \\ (0.20)} & \makecell{2.56 \\ (0.30)} & \makecell{3.64 \\ (0.32)} & \makecell{2.73 \\ (0.21)} & \makecell{1.40 \\ (0.11)} & \makecell{1.87 \\ (0.30)} & \makecell{2.36 \\ (0.22)} & \makecell{1.45 \\ (0.09)} \\
\addlinespace
\makecell{GRPO \\ (Ultrafeedback, $\beta=0.001$, Step 1000)} & \makecell{Temperature \\ (t=1.0)} & \makecell{3.94 \\ (0.18)} & \makecell{2.75 \\ (0.30)} & \makecell{4.29 \\ (0.27)} & \makecell{2.67 \\ (0.20)} & \makecell{1.50 \\ (0.13)} & \makecell{1.83 \\ (0.29)} & \makecell{2.58 \\ (0.23)} & \makecell{1.51 \\ (0.10)} \\
\bottomrule
\end{tabular}
\end{table}

\clearpage

\begin{table}[h!]
\centering
\caption{\centering \# of Functionally Diverse Responses by Diversity-Promoting Alignment Strategy}
\label{tab:fun_div_darling}
\tiny
\resizebox{\textwidth}{!}{
\begin{tabular}{cccccccccc}
\multicolumn{10}{c}{(Using Only GPT-4o as the Functional Diversity Judge)} \\
\toprule
\makecell{Model} & \makecell{Sampling Strategy} & \makecell{A} & \makecell{B} & \makecell{C} & \makecell{D} & \makecell{E} & \makecell{F} & \makecell{G} & \makecell{H} \\
\midrule
Llama-3.1-8B-Instruct & \makecell{Temperature \\ (t=0.1)} & \makecell{1.08 \\ (0.05)} & \makecell{1.25 \\ (0.14)} & \makecell{1.71 \\ (0.27)} & \makecell{1.93 \\ (0.18)} & \makecell{1.56 \\ (0.14)} & \makecell{1.52 \\ (0.23)} & \makecell{1.47 \\ (0.15)} & \makecell{1.09 \\ (0.04)} \\
\addlinespace
Llama-3.1-8B-Instruct & \makecell{Temperature \\ (t=0.5)} & \makecell{1.19 \\ (0.09)} & \makecell{1.75 \\ (0.25)} & \makecell{2.79 \\ (0.47)} & \makecell{2.22 \\ (0.18)} & \makecell{2.02 \\ (0.17)} & \makecell{1.48 \\ (0.24)} & \makecell{1.91 \\ (0.21)} & \makecell{1.47 \\ (0.10)} \\
\addlinespace
Llama-3.1-8B-Instruct & \makecell{Temperature \\ (t=1.0)} & \makecell{1.45 \\ (0.13)} & \makecell{2.31 \\ (0.30)} & \makecell{3.71 \\ (0.37)} & \makecell{2.42 \\ (0.20)} & \makecell{2.68 \\ (0.21)} & \makecell{2.00 \\ (0.29)} & \makecell{2.51 \\ (0.24)} & \makecell{1.78 \\ (0.13)} \\
\addlinespace
\midrule
\addlinespace
\makecell{GRPO \\ (Wildchat, $\beta=0.001$, Step 1000)} & \makecell{Temperature \\ (t=0.1)} & \makecell{2.79 \\ (0.21)} & \makecell{1.62 \\ (0.24)} & \makecell{2.71 \\ (0.40)} & \makecell{2.84 \\ (0.19)} & \makecell{1.12 \\ (0.05)} & \makecell{1.09 \\ (0.06)} & \makecell{1.22 \\ (0.08)} & \makecell{1.12 \\ (0.04)} \\
\addlinespace
\makecell{GRPO \\ (Wildchat, $\beta=0.001$, Step 1000)} & \makecell{Temperature \\ (t=0.5)} & \makecell{3.55 \\ (0.19)} & \makecell{2.31 \\ (0.28)} & \makecell{3.29 \\ (0.34)} & \makecell{3.13 \\ (0.20)} & \makecell{1.18 \\ (0.07)} & \makecell{1.17 \\ (0.08)} & \makecell{1.47 \\ (0.15)} & \makecell{1.27 \\ (0.07)} \\
\addlinespace
\makecell{GRPO \\ (Wildchat, $\beta=0.001$, Step 1000)} & \makecell{Temperature \\ (t=1.0)} & \makecell{4.00 \\ (0.16)} & \makecell{2.69 \\ (0.35)} & \makecell{3.93 \\ (0.27)} & \makecell{3.09 \\ (0.20)} & \makecell{1.16 \\ (0.06)} & \makecell{1.04 \\ (0.04)} & \makecell{1.67 \\ (0.17)} & \makecell{1.36 \\ (0.10)} \\
\addlinespace
\midrule
\addlinespace
\makecell{GRPO \\ (Ultrafeedback, $\beta=0.001$, Step 1000)} & \makecell{Temperature \\ (t=0.1)} & \makecell{3.15 \\ (0.19)} & \makecell{1.69 \\ (0.24)} & \makecell{3.14 \\ (0.42)} & \makecell{2.25 \\ (0.17)} & \makecell{1.20 \\ (0.06)} & \makecell{1.52 \\ (0.21)} & \makecell{1.96 \\ (0.20)} & \makecell{1.24 \\ (0.06)} \\
\addlinespace
\makecell{GRPO \\ (Ultrafeedback, $\beta=0.001$, Step 1000)} & \makecell{Temperature \\ (t=0.5)} & \makecell{3.66 \\ (0.20)} & \makecell{2.56 \\ (0.30)} & \makecell{3.64 \\ (0.32)} & \makecell{2.73 \\ (0.21)} & \makecell{1.40 \\ (0.11)} & \makecell{1.87 \\ (0.30)} & \makecell{2.36 \\ (0.22)} & \makecell{1.45 \\ (0.09)} \\
\addlinespace
\makecell{GRPO \\ (Ultrafeedback, $\beta=0.001$, Step 1000)} & \makecell{Temperature \\ (t=1.0)} & \makecell{3.94 \\ (0.18)} & \makecell{2.75 \\ (0.30)} & \makecell{4.29 \\ (0.27)} & \makecell{2.67 \\ (0.20)} & \makecell{1.50 \\ (0.13)} & \makecell{1.83 \\ (0.29)} & \makecell{2.58 \\ (0.23)} & \makecell{1.51 \\ (0.10)} \\
\addlinespace
\midrule
\addlinespace
\makecell{GRPO w/DARLING \\ (Wildchat, $\beta=0.001$, Step 1000)} & \makecell{Temperature \\ (t=0.1)} & \makecell{3.47 \\ (0.19)} & \makecell{1.69 \\ (0.20)} & \makecell{3.14 \\ (0.44)} & \makecell{2.60 \\ (0.20)} & \makecell{1.70 \\ (0.15)} & \makecell{1.83 \\ (0.28)} & \makecell{2.11 \\ (0.23)} & \makecell{1.35 \\ (0.08)} \\
\addlinespace
\makecell{GRPO w/DARLING \\ (Wildchat, $\beta=0.001$, Step 1000)} & \makecell{Temperature \\ (t=0.5)} & \makecell{3.94 \\ (0.16)} & \makecell{2.25 \\ (0.25)} & \makecell{3.29 \\ (0.40)} & \makecell{2.75 \\ (0.19)} & \makecell{2.06 \\ (0.14)} & \makecell{2.22 \\ (0.32)} & \makecell{2.82 \\ (0.23)} & \makecell{1.64 \\ (0.12)} \\
\addlinespace
\makecell{GRPO w/DARLING \\ (Wildchat, $\beta=0.001$, Step 1000)} & \makecell{Temperature \\ (t=1.0)} & \makecell{4.26 \\ (0.14)} & \makecell{3.19 \\ (0.29)} & \makecell{3.64 \\ (0.41)} & \makecell{2.65 \\ (0.19)} & \makecell{2.52 \\ (0.19)} & \makecell{2.13 \\ (0.32)} & \makecell{3.53 \\ (0.22)} & \makecell{1.89 \\ (0.14)} \\
\addlinespace
\midrule
\addlinespace
\makecell{GRPO w/DARLING \\ (Ultrafeedback, $\beta=0.001$, Step 1000)} & \makecell{Temperature \\ (t=0.1)} & \makecell{2.43 \\ (0.18)} & \makecell{1.19 \\ (0.14)} & \makecell{2.29 \\ (0.42)} & \makecell{2.29 \\ (0.18)} & \makecell{1.50 \\ (0.12)} & \makecell{1.26 \\ (0.13)} & \makecell{1.62 \\ (0.16)} & \makecell{1.10 \\ (0.04)} \\
\addlinespace
\makecell{GRPO w/DARLING \\ (Ultrafeedback, $\beta=0.001$, Step 1000)} & \makecell{Temperature \\ (t=0.5)} & \makecell{3.42 \\ (0.20)} & \makecell{1.56 \\ (0.27)} & \makecell{2.86 \\ (0.47)} & \makecell{2.38 \\ (0.18)} & \makecell{1.58 \\ (0.13)} & \makecell{1.83 \\ (0.31)} & \makecell{2.13 \\ (0.23)} & \makecell{1.28 \\ (0.07)} \\
\addlinespace
\makecell{GRPO w/DARLING \\ (Ultrafeedback, $\beta=0.001$, Step 1000)} & \makecell{Temperature \\ (t=1.0)} & \makecell{3.53 \\ (0.20)} & \makecell{1.75 \\ (0.27)} & \makecell{3.14 \\ (0.40)} & \makecell{2.25 \\ (0.19)} & \makecell{1.52 \\ (0.12)} & \makecell{1.83 \\ (0.29)} & \makecell{2.16 \\ (0.22)} & \makecell{1.33 \\ (0.08)} \\
\addlinespace
\midrule
\addlinespace
\makecell{GRPO w/DARLING \& Task Diversity Judges \\ (Wildchat, $\beta=0.001$, Step 1000)} & \makecell{Temperature \\ (t=0.1)} & \makecell{2.58 \\ (0.19)} & \makecell{1.19 \\ (0.10)} & \makecell{2.21 \\ (0.45)} & \makecell{3.07 \\ (0.19)} & \makecell{3.10 \\ (0.19)} & \makecell{2.70 \\ (0.33)} & \makecell{3.44 \\ (0.23)} & \makecell{2.11 \\ (0.14)} \\
\addlinespace
\makecell{GRPO w/DARLING \& Task Diversity Judges \\ (Wildchat, $\beta=0.001$, Step 1000)} & \makecell{Temperature \\ (t=0.5)} & \makecell{3.42 \\ (0.19)} & \makecell{1.88 \\ (0.26)} & \makecell{2.43 \\ (0.45)} & \makecell{3.29 \\ (0.20)} & \makecell{3.98 \\ (0.17)} & \makecell{3.43 \\ (0.35)} & \makecell{3.93 \\ (0.22)} & \makecell{2.92 \\ (0.16)} \\
\addlinespace
\makecell{GRPO w/DARLING \& Task Diversity Judges \\ (Wildchat, $\beta=0.001$, Step 1000)} & \makecell{Temperature \\ (t=1.0)} & \makecell{3.89 \\ (0.18)} & \makecell{2.25 \\ (0.32)} & \makecell{3.21 \\ (0.38)} & \makecell{3.47 \\ (0.17)} & \makecell{4.46 \\ (0.13)} & \makecell{4.00 \\ (0.29)} & \makecell{4.36 \\ (0.19)} & \makecell{3.32 \\ (0.16)} \\
\addlinespace
\midrule
\addlinespace
\makecell{GRPO w/DARLING \& Task Diversity Judges \\ (Ultrachat, $\beta=0.001$, Step 1000)} & \makecell{Temperature \\ (t=0.1)} & \makecell{1.87 \\ (0.16)} & \makecell{1.25 \\ (0.11)} & \makecell{1.21 \\ (0.15)} & \makecell{1.95 \\ (0.17)} & \makecell{1.78 \\ (0.17)} & \makecell{2.00 \\ (0.31)} & \makecell{2.22 \\ (0.22)} & \makecell{1.33 \\ (0.08)} \\
\addlinespace
\makecell{GRPO w/DARLING \& Task Diversity Judges \\ (Ultrachat, $\beta=0.001$, Step 1000)} & \makecell{Temperature \\ (t=0.5)} & \makecell{3.23 \\ (0.21)} & \makecell{1.62 \\ (0.27)} & \makecell{2.07 \\ (0.37)} & \makecell{2.44 \\ (0.18)} & \makecell{2.66 \\ (0.18)} & \makecell{2.87 \\ (0.36)} & \makecell{2.89 \\ (0.24)} & \makecell{1.81 \\ (0.12)} \\
\addlinespace
\makecell{GRPO w/DARLING \& Task Diversity Judges \\ (Ultrachat, $\beta=0.001$, Step 1000)} & \makecell{Temperature \\ (t=1.0)} & \makecell{3.68 \\ (0.21)} & \makecell{2.31 \\ (0.28)} & \makecell{3.14 \\ (0.40)} & \makecell{2.65 \\ (0.19)} & \makecell{3.14 \\ (0.21)} & \makecell{3.17 \\ (0.35)} & \makecell{3.84 \\ (0.24)} & \makecell{2.43 \\ (0.16)} \\
\bottomrule
\end{tabular}
}
\end{table}

\clearpage

\subsection{Infinity-Chat Experiment Results}
\label{apx:infinity-chat-res}

\clearpage 

\subsection{Prompt-Level Illustrations of Functional Diversity}
\label{apx:prompt_illustrations}

\begin{figure}[h]
    \centering
    \includegraphics[width=\linewidth]{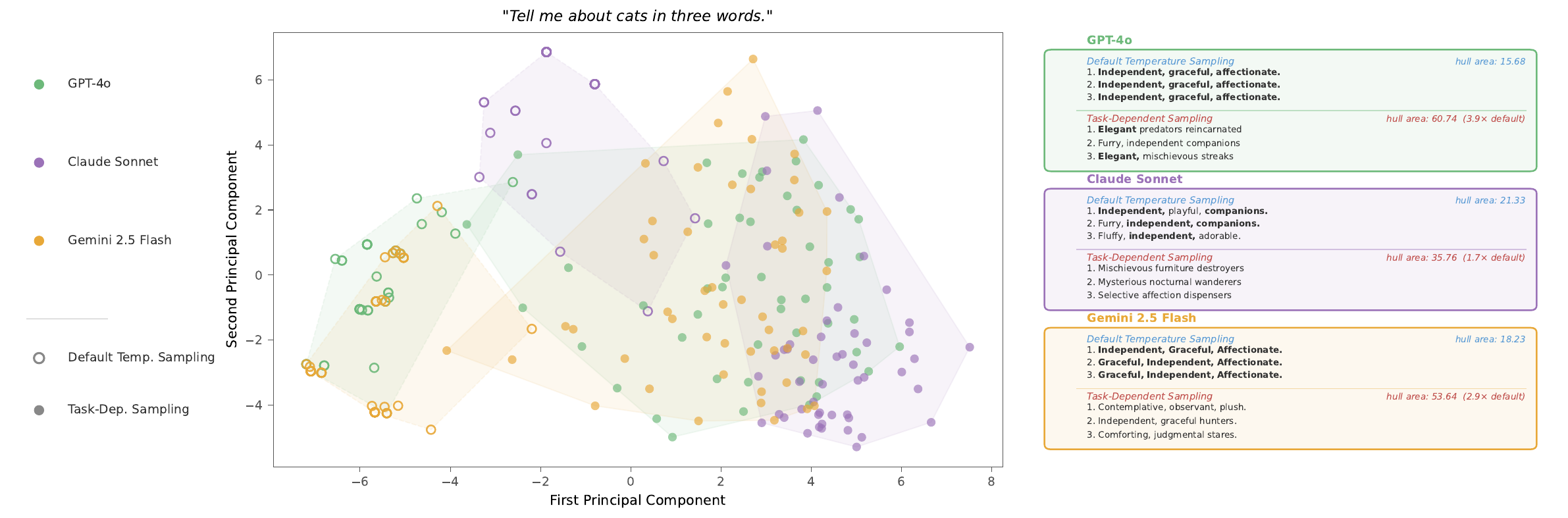}
    \caption{\textbf{Default sampling produces near-identical three-word descriptions of cats across all models; task-dependent sampling elicits genuinely distinct characterizations.}
    Prompt: \textit{``Tell me about cats in three words''} (Creative Writing, Category~G; \textsc{Infinity-Chat100}).
    Under default sampling, all three models converge on variations of ``Independent, graceful, affectionate,'' with GPT-4o and Gemini repeating the exact same triplet across multiple responses.
    Task-dependent sampling expands the hull area by 3.9$\times$ (GPT-4o), 1.7$\times$ (Claude-4-Sonnet), and 2.9$\times$ (Gemini 2.5 Flash), eliciting responses such as ``Elegant predators reincarnated,'' ``Mischievous furniture destroyers,'' and ``Comforting, judgmental stares'' --- conceptually distinct framings that go beyond surface-level lexical variation.}
    \label{fig:illus_cats}
\end{figure}

\begin{figure}[h]
    \centering
    \includegraphics[width=\linewidth]{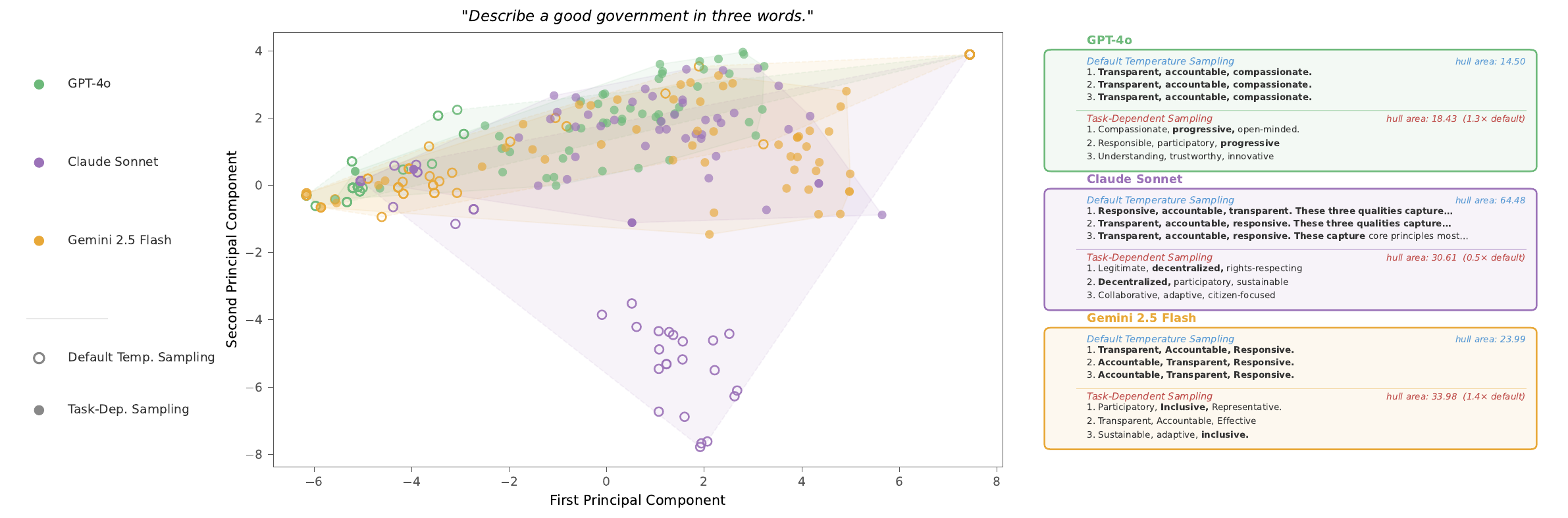}
    \caption{\textbf{For an opinion prompt about government, task-dependent sampling increases diversity for GPT-4o and Gemini but contracts it for Claude-4-Sonnet, illustrating that gains are not uniform across models.}
    Prompt: \textit{``Describe a good government in three words''} (Advice or Opinions, Category~H; \textsc{Infinity-Chat100}).
    GPT-4o and Gemini default to near-identical responses (``Transparent, accountable, compassionate'' and ``Accountable, Transparent, Responsive'' repeated verbatim), and task-dependent sampling increases their hull areas by 1.3$\times$ and 1.4$\times$ respectively.
    Claude-4-Sonnet, however, shows high default diversity (hull area 64.48) that contracts under task-dependent sampling (30.61, 0.5$\times$), suggesting that for already-diverse models on opinion tasks, task-dependent prompting may over-constrain the response space.
    This example motivates careful per-model evaluation of task-dependent sampling gains.}
    \label{fig:illus_gov_words}
\end{figure}

\begin{figure}[h]
    \centering
    \includegraphics[width=\linewidth]{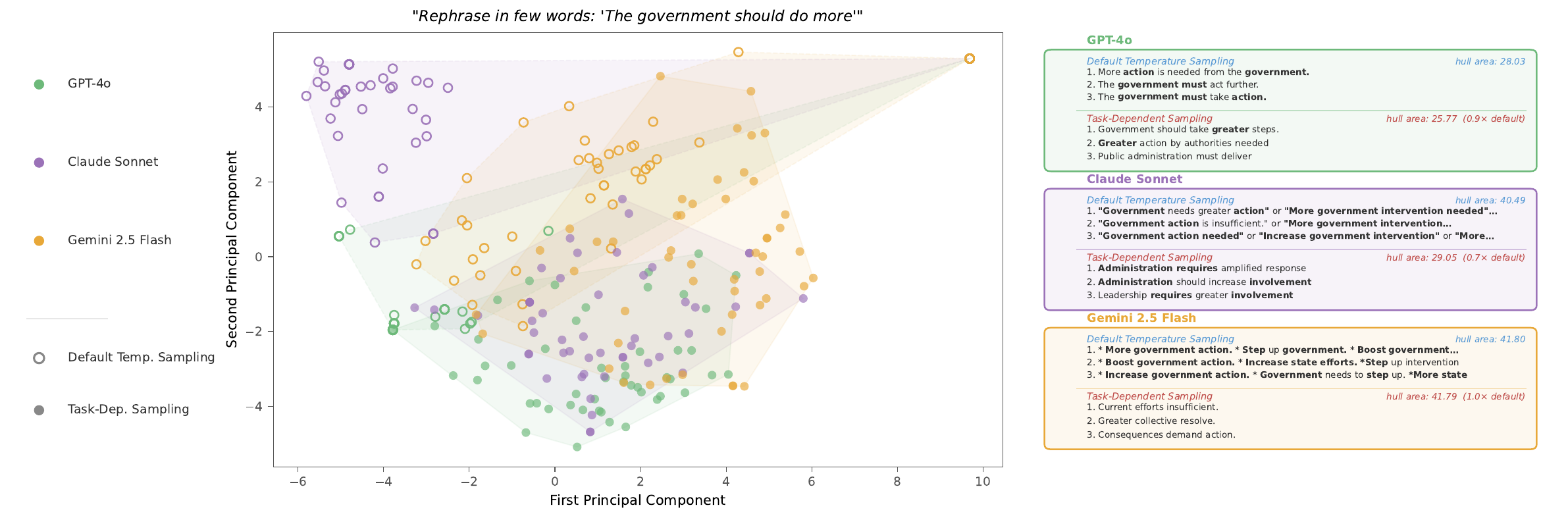}
    \caption{\textbf{A paraphrasing prompt with an implicit fidelity constraint illustrates that task-dependent sampling can be counterproductive when the task has a singular intended output.} 
    Prompt: \textit{``Rephrase in few words: `The government should do more'\,''} (hand-constructed). Although the prompt is open-ended in form, it carries an implicit constraint: a faithful, concise paraphrase of the source sentence. Hull areas under task-dependent sampling are 0.9$\times$ (GPT-4o), 0.7$\times$ (Claude-4-Sonnet), and 1.0$\times$ (Gemini 2.5 Flash) relative to default, confirming that task-dependent sampling does not artificially inflate diversity here. Default responses such as ``More action is needed from the government'' and task-dependent responses such as ``Government should take greater steps'' are semantically equivalent paraphrases, the correct behavior. This example motivates our taxonomy's distinction between prompts where diversity is meaningful and those where it is not, even when the prompt appears open-ended.}
    \label{fig:illus_govmore}
\end{figure}

\begin{figure}[h]
    \centering
    \includegraphics[width=\linewidth]{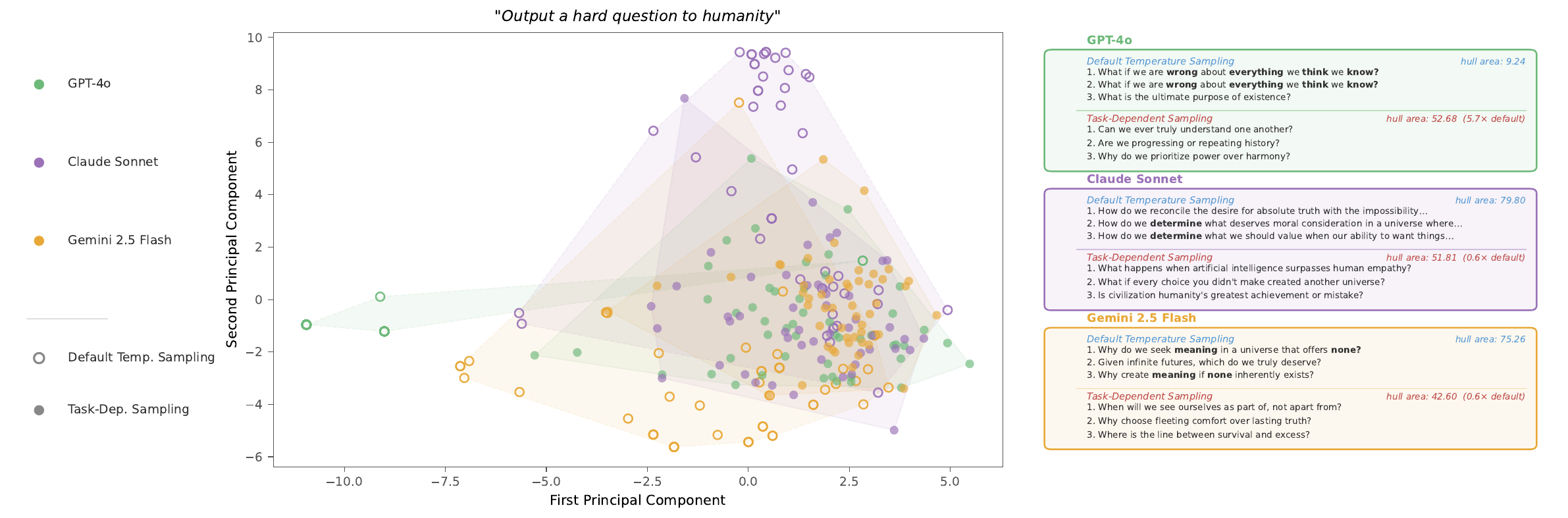}
    \caption{\textbf{A highly open-ended creative prompt reveals model-dependent behavior: task-dependent sampling dramatically expands diversity for GPT-4o but contracts it for Claude-4-Sonnet and Gemini 2.5 Flash, which already exhibit high baseline diversity.}
    Prompt: \textit{``Output a hard question to humanity''} (Creative Writing / Hypothetical, Category~G; \textsc{Infinity-Chat100}).
    GPT-4o defaults to repeating ``What if we are wrong about everything we think we know?'' and benefits greatly from task-dependent sampling (hull area $9.24 \to 52.68$, 5.7$\times$).
    Claude-4-Sonnet and Gemini 2.5 Flash already generate diverse philosophical questions by default (hull areas 79.80 and 75.26), and task-dependent sampling reduces their hull areas to 51.81 (0.6$\times$) and 42.60 (0.6$\times$) respectively.
    This pattern suggests that task-dependent sampling is most beneficial for models with low baseline diversity, and that aggregate results mask meaningful model-level heterogeneity.}
    \label{fig:illus_humanity}
\end{figure}

\begin{figure}[h]
    \centering
    \includegraphics[width=\linewidth]{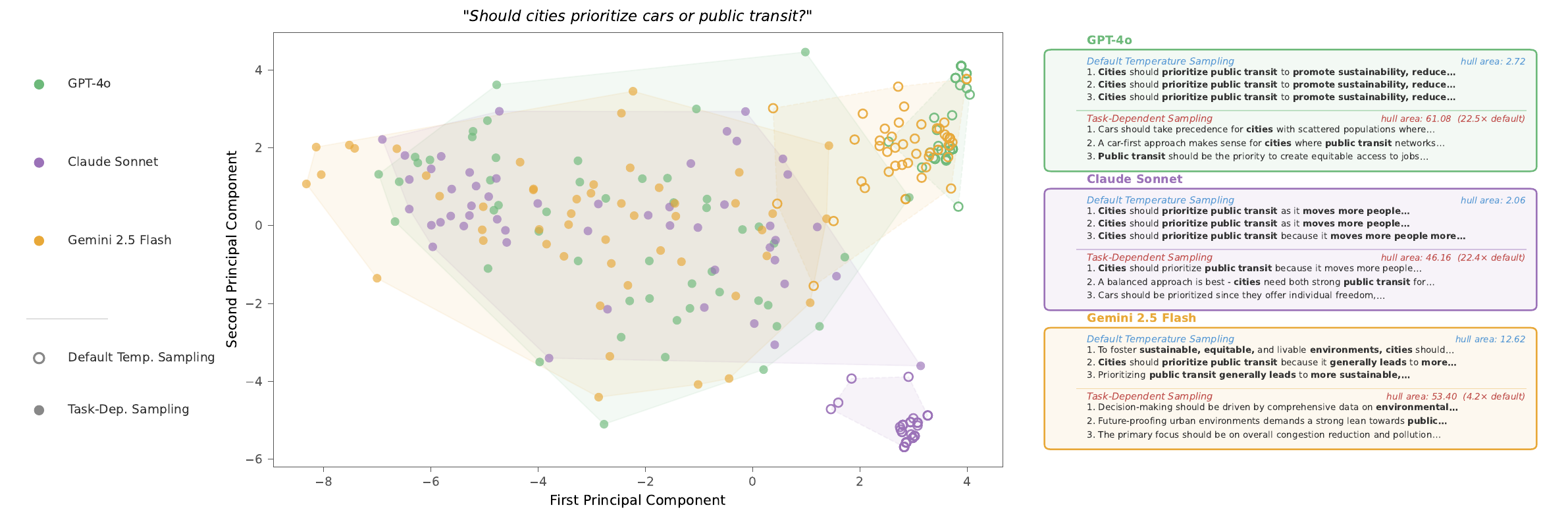}
    \caption{\textbf{For an opinion prompt about urban policy, default sampling produces homogenized pro-transit responses; task-dependent sampling elicits a broader range of perspectives including car-first and balanced viewpoints.}
    Prompt: \textit{``Should cities prioritize cars or public transit?''} (Advice or Opinions, Category~H; \textsc{Infinity-Chat100}).
    Under default sampling, all three models converge on pro-public-transit responses: GPT-4o repeats ``Cities should prioritize public transit to promote sustainability'' verbatim three times; Claude-4-Sonnet repeats ``Cities should prioritize public transit as it moves more people''; Gemini 2.5 Flash repeats ``Cities should prioritize public transit because it generally leads to more.''
    Task-dependent sampling elicits a wider range of viewpoints: GPT-4o produces responses acknowledging that ``Cars should take precedence for cities with scattered populations'' alongside pro-transit arguments; Claude-4-Sonnet introduces a ``balanced approach'' perspective; Gemini 2.5 Flash introduces environmental and congestion-reduction framings.
    This example illustrates the representation harm of homogenization on opinion tasks: without task-dependent sampling, users consistently receive a single perspective regardless of the model or sampling run.}
    \label{fig:illus_transit}
\end{figure}

The following figures illustrate functional diversity with examples of model responses to a few hand picked prompts. Prompts were selected to be short and yield concise responses where functional differences are visually apparent; these are illustrative examples rather than a representative sample. Each plot shows a PCA projection of response embeddings for three commercial models (GPT-4o, Claude-4-Sonnet, Gemini-2.5-Flash) under default temperature sampling (open circles) and task-dependent sampling (filled circles), with convex hulls indicating the spread of each model's outputs. Response excerpts are shown on the right. Figures~\ref{fig:illus_cats}--\ref{fig:illus_transit} include prompts from \textsc{Infinity-Chat100}~\citep{jiang2025artificial}; Figure~\ref{fig:illus_govmore} uses a hand-constructed prompt to illustrate a task where homogenization is desirable.

\end{document}